\documentclass{article}

\usepackage{microtype}
\usepackage{graphicx}
\usepackage{subcaption}
\usepackage{booktabs} 
\usepackage{multirow}
\usepackage{hyperref}

\usepackage[accepted]{icml2026}

\usepackage{amsmath}
\usepackage{amssymb}
\usepackage{mathtools}
\usepackage{amsthm}

\usepackage[capitalize,noabbrev]{cleveref}

\theoremstyle{plain}

\theoremstyle{definition}

\theoremstyle{remark}

\usepackage[textsize=tiny]{todonotes}

\icmltitlerunning{A Low-Rank Subspace Analysis of LLM Interventions}

\begin{document}

\twocolumn[
  \icmltitle{A Low-Rank Subspace Analysis of LLM Interventions}



  \icmlsetsymbol{equal}{*}

  \begin{icmlauthorlist}

    \icmlauthor{Angira Sharma}{comp}
    \icmlauthor{Christian Schroeder de Witt}{comp}
    \icmlauthor{Philip Torr}{comp}
    \icmlauthor{Anisoara Calinescu}{comp}
    \icmlauthor{Jialin Yu}{comp}

  \end{icmlauthorlist}

  \icmlaffiliation{comp}{University of Oxford}

  \icmlcorrespondingauthor{Angira Sharma}{angira.sharma@cs.ox.ac.uk}

  \icmlkeywords{Machine Learning, ICML}

  \vskip 0.3in
]



\printAffiliationsAndNotice{}  

\begin{abstract}
Interventions  designed to modify a particular behavior in LLMs, such as refusal or sycophancy, often produce unintended changes in other behaviors. This lack of targeted control makes it difficult to design and implement reliable safety controls. To understand these side-effects,  we introduce a diagnostic framework for analyzing interacting behaviors in LLMs. We model behaviors as low-rank subspaces in activation space, and study how interventions influence across behaviors. Across multiple instruction-tuned models (7B--70B) and across refusal, jailbreak, and sycophancy settings, we find that different behaviors share internal representations, and intervening on one behavior alters others in asymmetric ways. Some behaviors act as upstream control points whose interventions propagate broadly across other behaviors, while others remain more isolated. We relate these effects to two geometric quantities: (i) the overlap between behavior subspaces, measured as the average squared cosine of principal angles, and (ii) the angle between each behavior subspace and the decision subspace (capturing the model’s final decision e.g., refuse vs.\ comply). Empirically, intervention effects on other behaviors tend to be larger for behavior pairs with higher subspace overlap, and for source behaviors whose subspaces lie closer (smaller angle) to the decision subspace.     These findings highlight a  challenge for targeted behavior control: behaviors are difficult to modify independently, as  interventions can propagate through shared representations and asymmetric interactions.
\end{abstract}


\section{Introduction}

Post-hoc safety interventions in LLMs, such as fine-tuning or activation steering, are effective at modifying a target behavior, but can also induce unintended changes in other behaviors \citep{betley2025emergentmisalignmentnarrowfinetuning,steering_generalisation_dan_tan, siddiqui2025dormant}. Moreover, there are limited tools for predicting which other behaviors will change as a consequence. These side effects pose a central challenge for AI safety: without understanding why interventions generalize or interfere, it is difficult to design targeted safeguards or to evaluate the downstream risks.

A common approach to identify and mitigate individual behaviors is to isolate the behavior, such as refusal or sycophancy, and apply targeted fine-tuning, or steering, or orthogonality constraints to modify it \cite{arditi2024refusallanguagemodelsmediated, zhao2025llms}. These methods enable intervention and improve interpretability, but typically treat behaviors as independent control targets.  However, interventions and behaviors often exhibit collateral effects: fine-tuning on narrow objectives induces effects in unrelated behaviors \cite{betley2025emergentmisalignmentnarrowfinetuning}, or reward hacking generalizes in unexpected ways \cite{kei_dunn_sleight_turpin_evhub_denison_perez_2024, schoolofrewarhacks}, and some of these are attributed to  parametric choices \cite{brumley2024comparing}. 
     
Prior work on distributed and overlapping representations \citep{inference_time_intervention_kenneth, ponkshe2025safety, zou2025representationengineeringtopdownapproach, zhao2025beyond, elhage2022superposition} suggests that different behaviors may not be separate, but instead share underlying representational structure that mediates how interventions propagate. Consequently, interventions applied to one behavior can induce unintended changes in others when their representations overlap, this can lead to unintended side effects, such as modifying refusal behavior while also altering sycophancy, deception, or other downstream behaviors (cross-behavior effect). While prior work has characterized behavior representations as overlapping, we focus on how these representations translate into intervention effects.

We introduce a representation-level framework for analyzing interacting behaviors in LLMs (Section \ref{sec:method}). Our approach models behaviors as low-rank subspaces, removes dominant decision-aligned variance to isolate behavior-specific structure, and quantifies geometric overlap between behavior subspaces. Using projection-based interventions, we  measure how these structural properties relate to cross-behavior effects.\\


\textbf{Contributions:}
\begin{enumerate}
\item \textbf{A diagnostic framework.}
We introduce a framework to identify cross-behavior effects of interventions, using subspace extraction, removing decision-aligned variance, and overlap analysis (Section \ref{sec:method}).

\item \textbf{Beyond single-direction behavior representations.}
We model behaviors as low-rank subspaces, which reveal shared structure underlying intervention effects.

\item \textbf{Factors associated with cross-behavior interference.}
We find that intervention effects depend on: (i) how much behaviors share representations (overlap), and (ii) how strongly behaviors are coupled to the model’s decision (decision coupling, defined as  the angle between the source behavior subspace and the decision subspace). Together, these help explain cross-behavior effects and their asymmetry (Section \ref{sec:results}).

\item \textbf{Intervention effects are asymmetric.}
We find that intervention effects are directional $(A \rightarrow B \neq B \rightarrow A)$ due to differences in decision coupling. Moreover, some behaviors act as upstream control points whose interventions propagate broadly across other behaviors, while others remain more localized.

\item \textbf{Empirical evidence.}
We provide empirical evidence for these findings across different models and scales (7B--70B), behaviors (refusal, sycophancy), and  datasets. 
\end{enumerate}

These results help explain why some linear interventions often induce unintended asymmetric cross-behavior effects, and highlight the limitations of modeling behaviors as isolated directions.


    \section{Related Work}

  \paragraph{Interventions in LLMs} Recent work shows that  behaviors in language models can be modified through linear interventions in activation space. Approaches such as activation steering and representation engineering demonstrate that adding or projecting along specific directions can  influence model outputs \cite{turner2024steeringlanguagemodelsactivation, zou2025representationengineeringtopdownapproach, inference_time_intervention_kenneth, park2024linearrepresentationhypothesisgeometry, elhage2021mathematical, marks2024the}. Similarly,      \cite{arditi2024refusallanguagemodelsmediated} show that refusal behavior can often be captured by a dominant linear direction, enabling  control via rank-1 interventions. While effective in many settings, these approaches typically treat behaviors as independent control targets and represent them as single directions. Interventions  exhibit unintended side effects, where modifying one behavior influences others \cite{betley2025emergentmisalignmentnarrowfinetuning, steering_generalisation_dan_tan, schoolofrewarhacks}. 
In contrast, we model behaviors as low-rank subspaces and study when and why these approaches entail cross-behavior effects.

  \paragraph{Overlapping representations} Mechanistic interpretability literature suggests that neural representations are often distributed and overlapping, a phenomenon described as superposition \cite{elhage2022superposition}. Related approaches, such as  sparse autoencoders \cite{bricken2023monosemanticity} and circuit-level analyses \cite{wang2022interpretabilitywildcircuitindirect, meng2023locatingeditingfactualassociations}, provide valuable insights into representation structure, though they primarily operate at the level of features or circuits rather than  behaviors under intervention. Closer to our perspective, \cite{ponkshe2025safety, zhao2025beyond} highlight that safety-related representations may not be cleanly separable, but  do not explain how shared representations translate into cross-behavior interference. We find that subspace overlap alone is insufficient to explain cross-behavior effects, and that alignment with a decision-related subspace is a key additional factor governing their magnitude and asymmetry. 

\paragraph{Steering}  Prior work demonstrates that  behaviors can be modified through activation steering and parameter-space editing methods \cite{rimsky-etal-2024-steering, ilharco2023editing}. These approaches focus on interventions that induce targeted behavioral changes. In contrast, we study the representational structure underlying these interventions, and analyze when they remain localized versus when they produce cross-behavior effects.

\section{Method}
\label{sec:method}
\paragraph{Overview}


We study how interventions on one behavior affect others through a representation-level lens: cross-behavior effects depend on their shared representations, and how strongly these representations influence the model’s output. When behaviors share representations, intervening on one can affect the other, while differences in decision coupling help predict the magnitude and direction of these effects, leading to asymmetric and cross-behavior effects.


%

\subsection{Preliminaries}
Following \cite{arditi2024refusallanguagemodelsmediated}, we consider a decoder-only transformer. For input $\mathbf{x}$ and layer $\ell$,
$\mathbf{h}^{(\ell)}_i(\mathbf{x})\in\mathbb{R}^{d_{\text{model}}}$ denotes the residual-stream
activation at token position $i$.
For a behavior label $m\in\mathcal{M}$ (e.g., sycophancy, deception), we construct a matched
contrastive dataset $\mathcal{D}_m=\mathcal{D}^+_m\cup \mathcal{D}^-_m$, where
$\mathcal{D}^+_m$ elicits $m$ and $\mathcal{D}^-_m$ suppresses $m$ (matched to control non-behavioral
confounds).
We represent each example by the last-token activation at the
chosen layer \cite{arditi2024refusallanguagemodelsmediated}.

\subsection{Subspace Extraction}
For each matched pair $(x_i^+, x_i^-)$ at layer \(\ell\), we compute the last-token contrast vector $\Delta_i = \mathbf{h}^{(\ell)}_{\text{last}}(x_i^+) - \mathbf{h}^{(\ell)}_{\text{last}}(x_i^-) $ and stack these into a matrix $\Delta \in \mathbb{R}^{N \times d_{\text{model}}}$. After centering, we compute the SVD\footnote{The initial SVD is computed before removing decision-aligned variance, and its leading components can therefore reflect the decision signal; we leverage this to identify and remove decision-aligned directions via correlation with the margin.} $\Delta_c = U \Sigma V^\top$, where columns of $V\in\mathbb{R}^{d_{\text{model}}\times d_{\text{model}}}$
    are principal components (PCs). We interpret the leading PCs as capturing dominant variation in the contrast representations, and define a behavior subspace by selecting the top-$k$ components, forming an orthonormal basis $B_m^{(\ell)} \in \mathbb{R}^{d_{\text{model}} \times k}$.

    \subsubsection{Decision margin}
To quantify how strongly representations influence the model’s decision, we use a scalar decision margin 
\begin{equation}
\mu_i \;=\; \log P(pos\mid \text{eval}_i)\;-\;\log P(neg\mid \text{eval}_i)
\label{eq:log_metric}
\end{equation} 
where \(\log P\) is the sum of token log-probabilities of the label string, and \textit{pos} and \textit{neg} denote the two fixed candidate responses corresponding to the target and alternative behaviors (e.g., REFUSE vs. COMPLY).     To avoid decoding artifacts, we use a log-probability margin as a continuous proxy for behavioral preference \cite{arditi2024refusallanguagemodelsmediated, marks2024the}.

\subsection{Decision-related variance removal}
\label{met:residualisation}
A key challenge is that raw representations are dominated by shared alignment  to the decision mechanism (e.g., refusal), which can artificially inflate  apparent similarity between categories.
 To isolate category-specific structure, we remove the top decision-aligned  components before computing subspace overlap. This residualization step ensures that overlap reflects shared behavior geometry rather than shared  alignment with the output decision.

To identify directions aligned with the behavioral decision (such as comply-refuse), we compute a \emph{global} decision subspace by pooling contrast vectors across all categories and computing PCA on the combined dataset.  We remove the top $k_{\text{dec}}$  decision-aligned components via projection. Let $s_{ij}$ denote the score of item $i$ on PC $v_j$, i.e., $s_{\cdot j}=\Delta_c^{(m,\ell)} v_j$. We rank PCs by $|\mathrm{corr}(s_{\cdot j},\, \mu)|$ and define the decision subspace $V_{\mathrm{dec}}\in\mathbb{R}^{d_{\text{model}}\times k_{\mathrm{dec}}}$ as the span of the top
$k_{\mathrm{dec}}$ ranked PCs (we use $k_{\mathrm{dec}}=2$)\footnote{We evaluate sensitivity to the choice of $k_{\mathrm{dec}}$ in Appendix \ref{app:k_dec_ablation}. While overlap magnitudes vary with $k_{\mathrm{dec}}$, nontrivial residual structure persists across settings, and the qualitative ordering of category relationships remains stable.}. Empirically, the first two decision-aligned PCs explain the majority of margin-correlated variance, while higher components exhibit rapidly diminishing correlation.

We remove decision-aligned variance and expose category/style structure within each behavior  by projecting contrasts onto the orthogonal complement (i.e., 'residualize')\footnote{Centering ($\Delta_c$) ensures that principal components reflect covariance structure rather than mean offsets \cite{murphy2012machine}. }:
\begin{equation}
\Delta_{\perp} \;=\; \Delta_c\bigl(I - V_{\text{dec}}V_{\text{dec}}^\top\bigr),
\label{eq:residualize}
\end{equation}
where $V_{\mathrm{dec}}$ has orthonormal columns so that $V_{\mathrm{dec}}V_{\mathrm{dec}}^\top$ is the projector onto the decision subspace. 

We use correlation with the decision margin as an operational proxy for decision coupling, without assuming it fully captures the underlying mechanism. Removing these components isolates decision-related variance, but residual overlap and cross-category effects persist, indicating shared structure beyond the global decision signal (Section \ref{sec:results}).


\paragraph{Behavior subspaces}
We model each behavior as a low-rank subspace\footnote{We model each category as a low-rank subspace rather than a single direction. This is motivated by the observation that category representations are distributed  and multi-dimensional. Bootstrap experiments (Appendix~\ref{sec:robustness}) show  that these subspaces are stable across resampling and far above a random-subspace  baseline, supporting the validity of this representation.
} by applying PCA\footnote{We use PCA as a low-rank estimator, not as evidence that individual PCs are mechanisms. Matched contrastive prompts and decision residualization reduce prompt-format and decision artifacts, and behavioral relevance is tested through held-out projection interventions rather than assumed from variance alone.} to residualized contrasts $\Delta_{\perp}^{(m,\ell)}$ and taking the top-$k$ components, orthonormalized, as a basis
    $B_m^{(\ell)} \in \mathbb{R}^{d_{\text{model}}\times k}$ for the behavior subspace at layer $\ell$. This choice is motivated by the observation that, after removing decision-related variance, behavior-specific structure is distributed across multiple directions rather than concentrated in a single axis.  Low-rank subspaces therefore provide a more stable and expressive representation than rank-1 directions. We use $k=2$ as the smallest rank that captures nontrivial subspace structure
while avoiding the increased collateral effects observed at higher ranks
(rank ablations in Appendix~\ref{app:rank}).

\subsection{Geometric Overlap}
\label{sec:geometry}
To quantify shared representation, we measure the overlap between residualized behavior subspaces  using principal angles.  For behaviors $m$ and $m'$,  let
$B_m\in\mathbb{R}^{d_{\text{model}}\times k}$ and
$B_{m'}\in\mathbb{R}^{d_{\text{model}}\times k'}$ have orthonormal columns, and define
$ M \;=\; B_m^\top B_{m'} \in \mathbb{R}^{k\times k'}.$ If $M = U\Sigma V^\top$ is an SVD, then $\sigma_i=\cos(\theta_i)$ are the cosines of the
principal angles $\{\theta_i\}_{i=1}^{k_*}$ with $k_*=\min(k,k')$. We  compute the average of the squared cosines of the principal angles to quantify the overlap between two behavior subspaces:
\begin{equation}
\mathrm{overlap}_\ell(m,m') \;=\; 
\frac{1}{k_*}\sum_{i=1}^{k_*}\sigma_i^2
\;=\;\frac{1}{k_*}\| (B_m^{(\ell)})^\top B_{m'}^{(\ell)} \|_F^2
\label{eq:overlap}
\end{equation}
The overlap score lies in $[0,1]$, with $0$ indicating orthogonal subspaces and $1$ indicating identical subspaces.

\subsection{Intervention and Cross-Behavior Effects}
\label{methods:causal_val}

To test whether geometric overlap induces cross-category effects, we project out a \emph{source}
subspace and measure the effect on a \emph{target} behavior's decision margin. For intervention experiments, we use the corresponding subspace estimated from the raw contrasts (prior to decision variance removal), denoted $B_{m,\mathrm{raw}}^{(\ell)}$.
At layer $\ell$, for an activation vector $\mathbf{h}\in\mathbb{R}^{d_{\text{model}}}$ we apply 
$\tilde{\mathbf{h}} = \bigl(I - \alpha B_{m,\text{raw}}^{(\ell)} (B_{m,\text{raw}}^{(\ell)})^\top\bigr)\,\mathbf{h}$,
\begin{equation}
\Delta M_{m'}(\ell) \;=\;
\mathbb{E}_{\text{test}}[\mu]_{\text{intervened}} \;-\;
\mathbb{E}_{\text{test}}[\mu]_{\text{baseline}}
\label{eq:hook}
\end{equation}
Here $\mathbb{E}_{\text{test}}[\cdot]$ denotes the empirical mean over held-out
test prompts for behavior $m'$. We relate \(\Delta M_{m'}(\ell)\) to \(\mathrm{overlap}_\ell(m,m')\) across layers and behavior pairs.     
This setup allows us to relate cross-behavior effects $\Delta M_{m'}(\ell)$ to both subspace overlap and decision coupling.
Unlike standard activation steering, which adds or subtracts a direction to steer the
decision, our projection-based interventions remove components of the activation,
serving as stronger ablation  rather than directional control.

\paragraph{Summary.}

Our framework extracts behavior subspaces, separates decision-related and residual structure, measures geometric overlap, and evaluates how these quantities predict cross-behavior effects under intervention.  Further explanation of methodology choices in Appendix \ref{app:methodology}.

\section{Experimental Setup}
\label{sec:experiments}

\paragraph{Models.} We evaluate our framework across six instruction-tuned open-weight language models spanning 7B--70B parameters, Gemma-2-9B-IT, Llama-3-8B-Instruct, Mistral-7B-Instruct-v0.3, Gemma-3-27B-IT, Llama-3.3-70B-Instruct, and Mistral-Small-24B-Instruct-2501.
 
\paragraph{Behaviors and datasets.}
We study two widely-studied safety relevant behaviors: refusal and sycophancy. For refusal, we use both a synthetic dataset and a jailbreak dataset from \cite{arditi2024refusallanguagemodelsmediated}, which contains pre-defined refusal categories, drawn from existing datasets  \textsc{ADVBENCH} \cite{zou2023universal}, \textsc{MALICIOUSINSTRUCT} \cite{huang2024catastrophic}, \textsc{TDC2023}  \cite{tdc2023}, and \textsc{HARMBENCH} \cite{harmbench}.  For sycophancy, we construct a synthetic dataset.


\paragraph{Synthetic dataset design.}
We construct five refusal categories: \textsc{Harmful}, \textsc{Copyright}, \textsc{NSFW}, \textsc{Deceptive/Malicious}, and \textsc{Against Policy}. For each category, we generate matched prompts that elicit compliance or refusal (Appendix~\ref{sythetic_data}). These categories are not intended as a complete taxonomy, but as controlled test cases spanning distinct policy dimensions. This allows us to test whether semantically distinct categories can share representation. 
To ensure results are not artifacts of synthetic design, we replicate the analysis on jailbreak and sycophancy datasets.

\paragraph{Controlled analysis.}
Synthetic datasets provide controlled conditions for isolating representational structure: matched prompts and balanced samples reduce confounds such as prompt imbalance or spurious correlations. This setup follows prior representation-level analyses \citep{elhage2022superposition,ilharco2023editing, park2024linearrepresentationhypothesisgeometry, wang2025personafeaturescontrolemergent, rimsky-etal-2024-steering}.


\section{Results}
\label{sec:results}

We evaluate whether cross-behavior intervention effects can be explained by representation structure. Across models (7B–70B) and behaviors (refusal, sycophancy), we find that subspace overlap helps predict where cross-effects are possible, while decision coupling helps predict the magnitude and direction. We discuss the results for synthetic refusal dataset here. Results for scaled models, and other datasets are available in  Appendices \ref{app_refusal}, \ref{app:jb}, \ref{app:sycophancy}. Unless otherwise stated, all figures pertain to the synthetic refusal dataset. 

\subsection{Behaviors Share Representation, and Interventions Are Not Isolated}
\label{res:overlap}

\paragraph{Behavior Subspaces Exhibit Nontrivial Overlap}

\begin{figure*}[t]
\centering
\includegraphics[width=\linewidth]{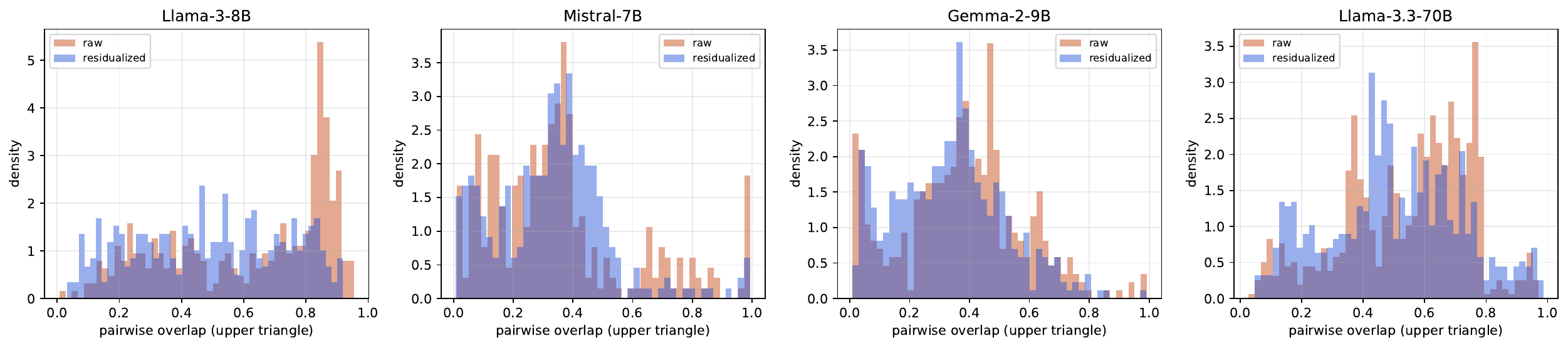}
\caption{
\textbf{Residualization reveals meaningful category-specific structure.}
We compare the distribution of pairwise subspace overlap between categories before (raw) and after removing the decision-aligned subspace (residualized), across models for synthetic refusal dataset.
Raw overlap (red) is  skewed toward high values, indicating that category subspaces appear  similar due to shared alignment with the decision mechanism.
After residualization (blue), the overlap distribution shifts and broadens, revealing  variability across category pairs.
This demonstrates that residualization is important to isolate category-specific representational structure, rather than shared decision-level similarity.
}
\label{fig:overlap_histograms}
\end{figure*}

We first examine the geometric structure of behavior representations. Figure~\ref{fig:overlap_histograms} compares pairwise subspace overlap before and after removing decision-related variance. Raw overlap is high, reflecting shared decision-related structure. After residualization (Section \ref{met:residualisation}), overlap becomes heterogeneous, spanning low to high values across category pairs. This suggests that raw overlap is dominated by shared decision-related structure, causing subcategories to appear spuriously similar. After residualization, heterogeneous overlap patterns emerge, revealing behavior-specific representational geometry.

\paragraph{Interventions Produce Cross-Behavior Effects} We next establish that interventions induce substantial cross-behavior effects. Figure~\ref{fig:cross_effect_main} shows that projecting out a single behavior subspace  affects other behaviors, with off-diagonal effects often comparable in magnitude to self-effects\footnote{We discuss $8^\text{th}$ layer from the output because it lies within the regime where cross-behavior overlap and intervention effects consistently emerge across models; full layer-wise analyses are provided in Appendix~\ref{app:layerwise_int} and Fig.~\ref{fig:layer_appendix}.}. This suggests that behaviors may be difficult to modify independently under linear subspace interventions. Further analysis suggests that cross-category effects are  associated with shared subspace components; detailed decomposition results are provided in Appendix~\ref{app:shared-private}.

\subsection{Overlap Alone Does Not Explain Intervention Effects}
\label{res:overlap_cross}

\begin{figure*}[t]
    \centering
    \includegraphics[width=\linewidth]{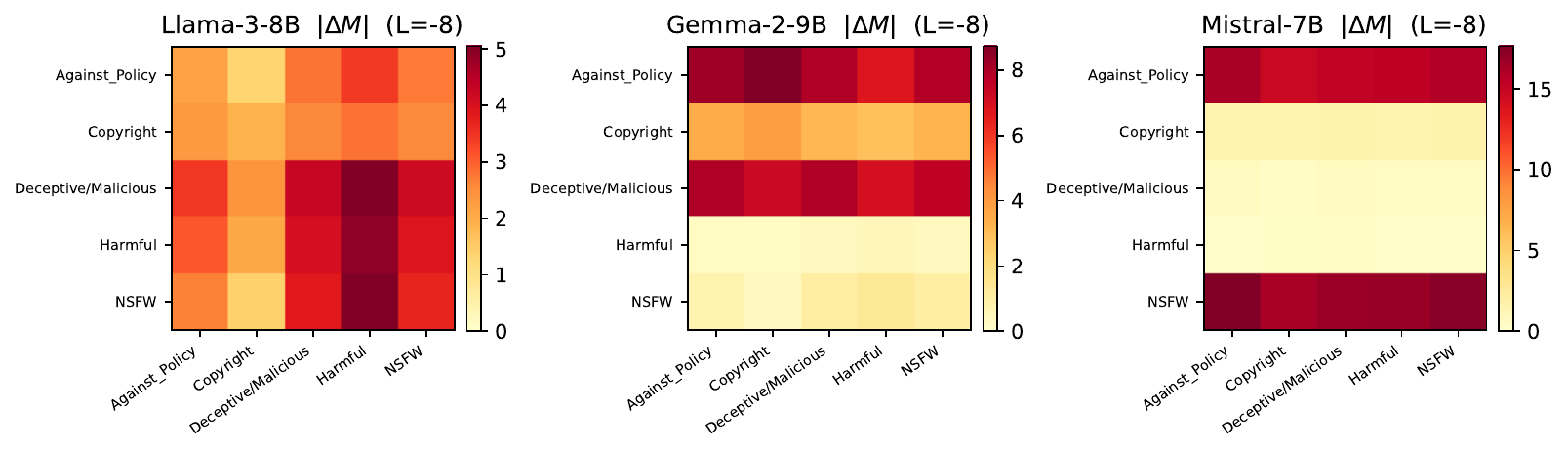}
    \caption{
    \textbf{Cross-category effects under subspace intervention at 8th layer from the output.}
    Each entry $(i, j)$ shows the change in decision margin for category $j$
    when projecting out the subspace of category $i$.
    Diagonal entries represent self-effects; off-diagonal entries capture
    cross-category transfer.
    Across the models (i) modifying one category
     affects others, often with comparable magnitude (such as \textsc{DECEPTIVE/MALICIOUS} in Llama and Gemma) (ii) cross-behavior effects are asymmetric - intervening on certain behaviors e.g., \textsc{Deceptive/Malicious} in Gemma-2-9B heavily degrades the performance on \textsc{NSFW} or \textsc{Harmful}, but the reverse intervention yields minimal collateral damage. 
    }
    \label{fig:cross_effect_main}
\end{figure*}

\begin{figure*}[t]
    \centering
    \includegraphics[width=\linewidth]{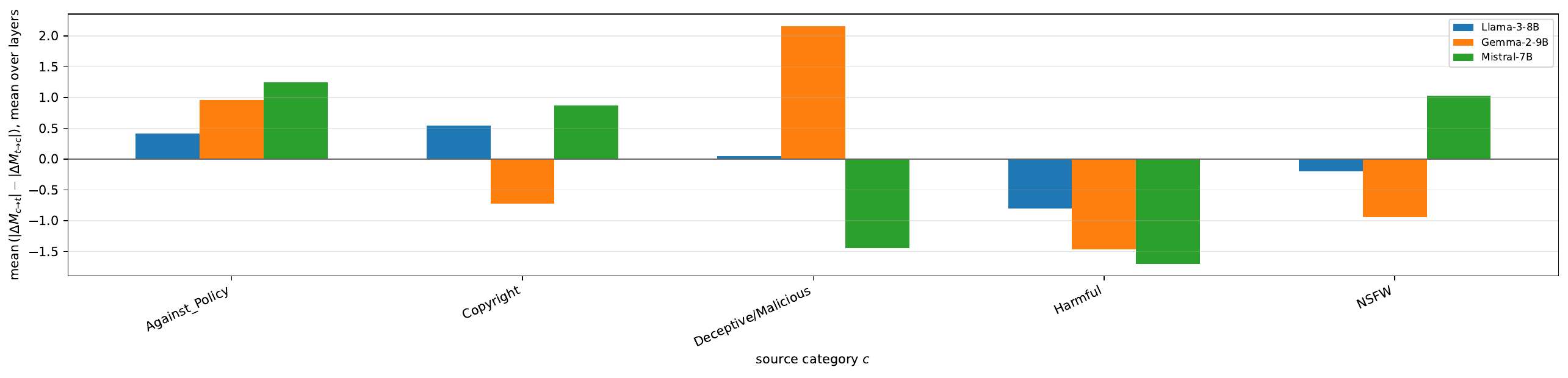}
    \caption{
    \textbf{Category-wise asymmetry in cross-category effects.}
    Some categories act as dominant sources of influence, while others
    primarily receive effects.  For each category $c$ we plot
$\sum_{b\neq c}\bigl|\Delta M_{c\rightarrow b}\bigr|
-\sum_{a\neq c}\bigl|\Delta M_{a\rightarrow c}\bigr|$,
where $\Delta M_{a\rightarrow b}$ is the mean change in the refusal--comply margin on category-$b$ prompts when intervening along category $a$'s subspace, averaged across layers. 
Positive values indicate categories that act as dominant sources of influence, while negative values correspond to categories that primarily receive effects. 
The  imbalance across models reveals a directional structure in how categories interact. }
    \label{fig:asymmetry_appendix}
\end{figure*}

\paragraph{Intervention Effects Are Asymmetric}
We find that intervening on behavior $A$ can significantly affect $B$, while the reverse intervention has minimal impact (Figures ~\ref{fig:cross_effect_main} and \ref{fig:asymmetry_appendix}). This suggests that some behaviors act as upstream sources (such as \textsc{DECEPTIVE/MALICIOUS} and \textsc{AGAINST POLICY} in Gemma-2-9B) whose perturbation propagates broadly, while others have more localized impact (such as \textsc{HARMFUL} and \textsc{NSFW} in Gemma-2-9B).

\begin{figure*}[t]
    \centering
    \includegraphics[width=\linewidth]{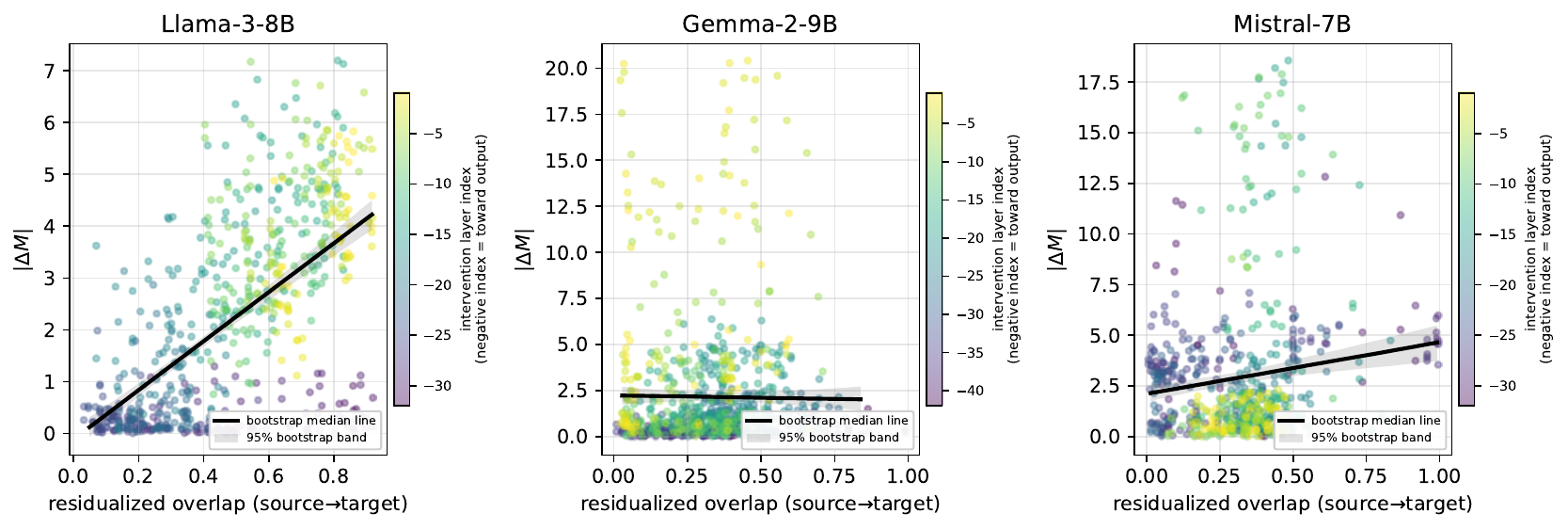}
    \caption{
    \textbf{Residualized subspace overlap vs.\ cross-category effect magnitude.}
    Each point corresponds to a directed category pair.
    Higher overlap indicates shared representational structure.
    While overlap identifies pairs that are susceptible to cross-effects,
    it does not reliably predict effect magnitude.
    }
    \label{fig:overlap_effect_main}
\end{figure*}

\paragraph{Cross-behavior Overlap}
We test whether shared representation explains cross-behavior effects. Figure~\ref{fig:overlap_effect_main} plots residualized overlap against effect magnitude. While higher overlap identifies pairs that can interact, it does not reliably predict effect strength: high-overlap pairs can exhibit weak effects, and some low-overlap pairs produce strong effects. As a sanity check, projecting out a category’s own subspace reliably alters its corresponding decision margin across models, suggesting that the extracted subspaces are behaviorally relevant under linear intervention (Appendix~\ref{app:within_steering}). This relationship is consistent across models and robust to resampling. 
(Appendix~\ref{sec:robustness}).


\subsection{Decision Coupling is Associated with Effect Magnitude and Direction}
\label{res:decision_angles}
\begin{figure*}[t]
    \centering
    \includegraphics[width=\linewidth]{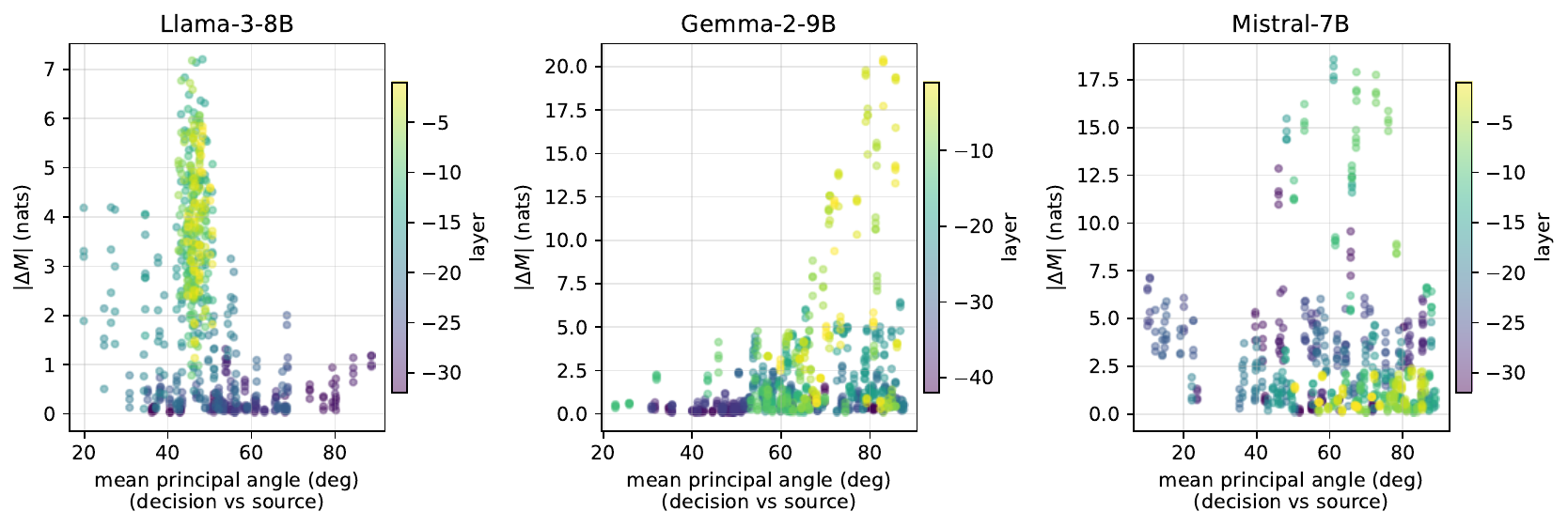}
    \caption{
    \textbf{Decision alignment vs.\ cross-category effects.}
    Each point shows a category and layer.
    Smaller angles indicate stronger alignment with the decision mechanism.
    Stronger alignment corresponds to larger cross-category
    effects, indicating that decision coupling affects magnitude. In models such as Gemma and Llama, we find that categories whose directions align more strongly with the decision mechanism produce larger cross-category effects. 
    }
    \label{fig:angle_effect_main}
\end{figure*}

\begin{figure*}[t]
    \centering
    \includegraphics[width=0.95\linewidth]{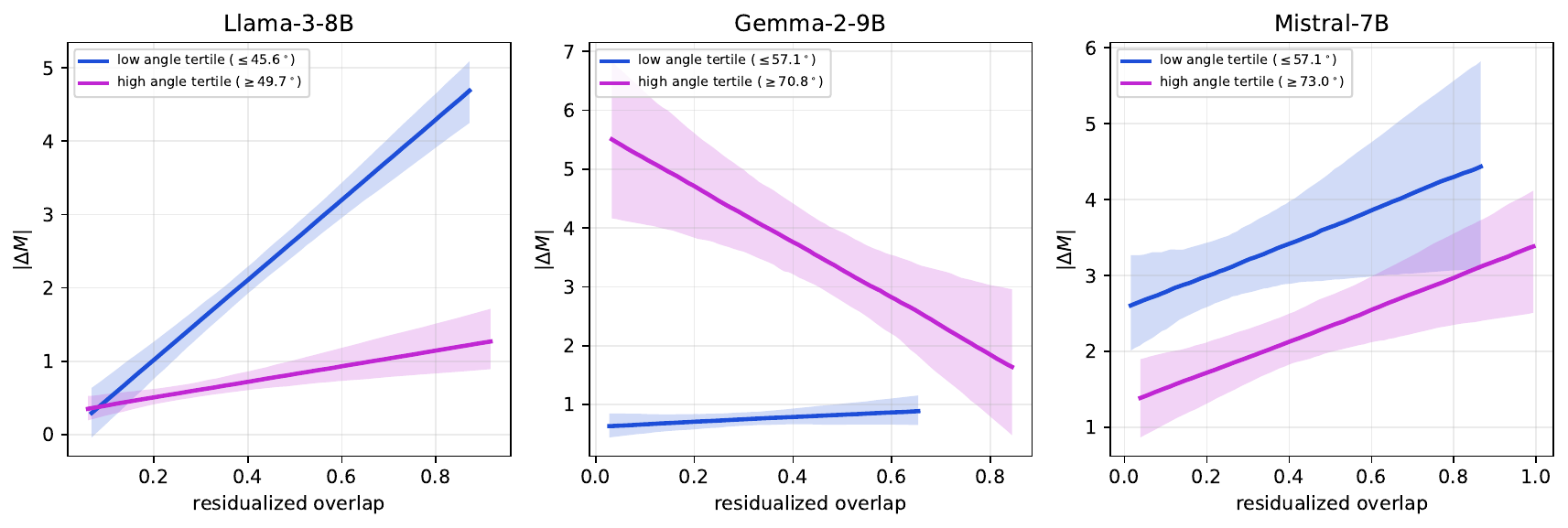}
    \caption{
    \textbf{Overlap–effect relationship stratified by decision alignment.}
    Overlap is indicative of stronger effects mainly when decision alignment is high,
    and has weak predictive power when alignment is low.  }
    \label{fig:stratified_appendix}
\end{figure*}

To explain this asymmetry and magnitude of cross-behavior effects, we analyze how strongly each behavior subspace influences the model’s decision. Figures ~\ref{fig:angle_effect_main} and \ref{fig:stratified_appendix} shows that behaviors with stronger decision coupling often produce larger global effects when intervened upon.

\subsection{Overlap and Decision Coupling are Associated with Effects}
\label{results:h2}

\begin{figure*}[t]
    \centering
    \includegraphics[width=\linewidth]{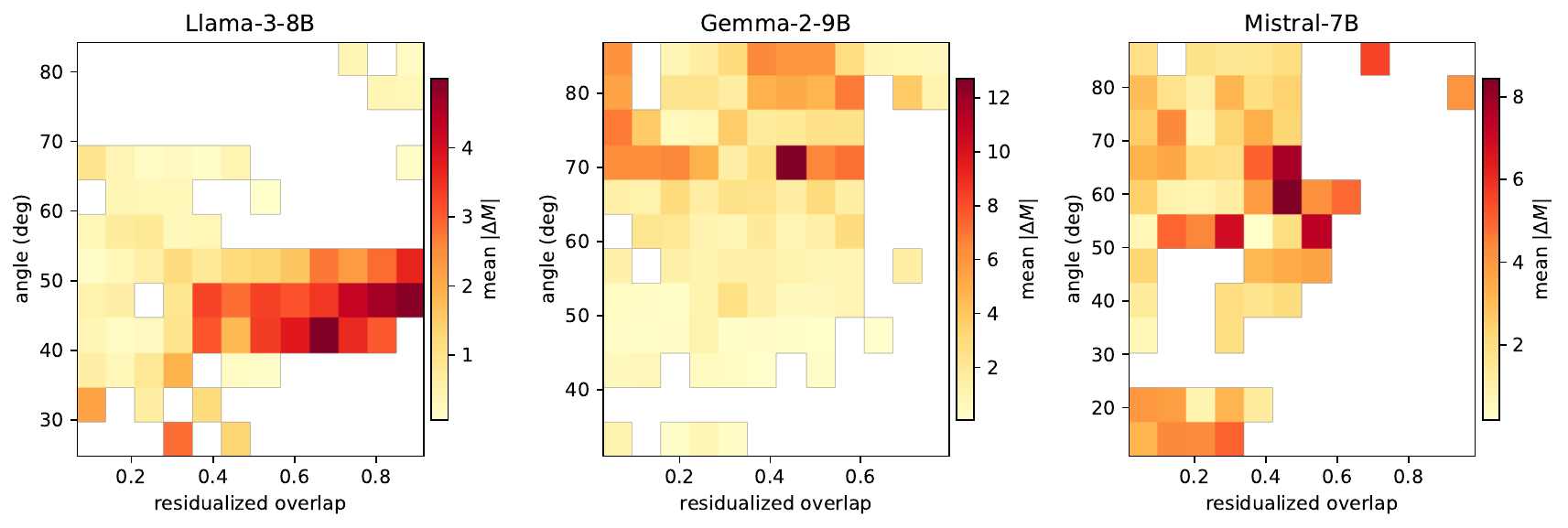}
    \caption{
    \textbf{Joint effect of overlap and decision alignment.}
    Mean cross-category effect magnitude as a function of residualized overlap
    and decision--category angle.
    Strong effects occur mainly when overlap is high and alignment is strong.
    }
    \label{fig:joint_main}
\end{figure*}

Finally, we jointly analyze overlap and decision coupling. Figure~\ref{fig:joint_main} shows that strong cross-behavior effects occur mainly when both conditions are met: (i) behaviors share representation (high overlap), and (ii) the source behavior strongly influences the decision (high decision coupling).


\paragraph{Results Across Models and Behaviors}
To ensure these findings are not artifacts of a specific model size or behavior, we replicated our framework on jailbreak (Appendix \ref{app:jb}) and sycophancy datasets (Appendix \ref{app:sycophancy}), and larger models (27B-70B) (Appendix \ref{app:results_larger_models}). Across both domains and larger models, similar qualitative patterns are generally observed. The strength of cross-behavior effects and their relationship to subspace geometry vary across model families, suggesting that differences in architecture and training regime may lead models to organize behavioral representations differently.

\paragraph{Random-Subspace Baseline}

To verify that these effects arise from structured representations, we compare against a random-subspace baseline. Replacing learned subspaces with random orthonormal subspaces produces effects that are 147$\times$–319$\times$ smaller across models (Table~\ref{tab:real_random_null}). This suggests that the observed effects are unlikely to arise solely from  generic low-rank projection (Appendix~\ref{app:random_null}).

\begin{table*}[h!]
\centering
\small
\setlength{\tabcolsep}{3.5pt}
\begin{tabular}{l|cc|ccc|cc}
\toprule
\multirow{2}{*}{Model}
& \multicolumn{2}{c|}{Overlap}
& \multicolumn{3}{c|}{Learned $|\Delta M|$}
& \multicolumn{2}{c}{Random null} \\
& residualized & random
& self & mean coll. & max coll.
& $|\Delta M|$ & gap \\
\midrule
Llama-3-8B & 0.667 & 0.0005 & 4.06 & 3.71 & 4.61 & 0.013 & $319\times$ \\
Gemma-2-9B & 0.266 & 0.0006 & 5.03 & 4.73 & 5.11 & 0.025 & $198\times$ \\
Mistral-7B & 0.332 & 0.0005 & 3.11 & 2.99 & 3.20 & 0.021 & $147\times$ \\
\bottomrule
\end{tabular}
\caption{\textbf{Learned category subspaces produce effects far above a
random-subspace null.}
Values are averaged over the final 12 decoder layers. Random overlap and random
intervention effects are computed by replacing each learned category subspace
with random orthonormal rank-2 subspaces matched in dimension and layer.
\emph{Self} denotes the diagonal intervention effect, \emph{mean coll.} the
mean off-diagonal effect, and \emph{max coll.} the per-source maximum
off-diagonal effect. Learned subspaces produce cross-category effects
147$\times$--319$\times$ larger than the random baseline, while maximum
collateral effects match or exceed self-effects in all models.}
\label{tab:real_random_null}
\end{table*}

 \paragraph{Further Results} We find that rank-1 representations are insufficient to capture behavior-specific structure: they produce weaker intervention effects and fail to recover the overlap patterns observed with low-rank subspaces (Appendix \ref{app:rank}).  Intervention strength ($\alpha$) ablations are in Appendix \ref{app:alpha_ablation}.  Appendix \ref{app_refusal} provides detailed results on the different categories for refusal, and Appendices ~\ref{app:k:layer_emergence}--\ref{app:k:decision_lowrank} Figs.~\ref{fig:layer_appendix}--\ref{fig:variance_appendix} provide depth profiles, decision-subspace spectra, for the three 7B--9B model runs. 

\paragraph{Summary}

Together, these results suggest that:
(1) asymmetric cross-category effects are often observed under  interventions,
(2) representational overlap alone does not predict effect magnitude, and
(3) decision alignment modulates the strength and asymmetry of these effects.

\section{Discussion}
\label{discussion}
Our results suggest that intervention effects in LLMs are associated with interaction between representation and decision: behaviors  therefore may not behave as isolated control targets, but interacting subspaces coupled through shared decision pathways. When subspaces overlap, intervening on one behavior often perturbs others, while the magnitude and direction of this effect depend on how strongly the perturbed subspace is aligned with the model’s decision-aligned subspace. This helps explain why overlap alone is insufficient, and why intervention effects are often asymmetric.


Another key implication is that, model-dependent, some behaviors act as upstream control points whose modification propagates broadly, while others remain more localized (Section \ref{res:overlap_cross}).  From an AI safety perspective, the distinction of asymmetric cross-category effects is critical for intervention design. Subspaces that are tightly coupled to upstream decision pathways warrant particular caution, as modifying them can induce widespread and unintended changes, whereas more downstream, category-specific subspaces offer opportunities for targeted interventions with reduced risk of collateral effects.  This perspective provides a foundation for developing more robust and predictable approaches to intervention, and highlights the need to move beyond independent, direction-based control toward methods that explicitly model behavioral interactions.



Finally, our results should not be interpreted as identifying full semantic mechanisms.  Rather, they show that modeling behaviors as interacting low-rank subspaces provides a useful representation-level framework for understanding intervention side effects. This is sufficient to diagnose when interventions are likely to induce side-effects, and provides a foundation for more structured approaches to behavior control.
  


\section{Limitations}
\label{limitations}
Our framework operates at the level of low-rank subspaces, which allows us to study shared representational structure but does not capture non-linear or higher-dimensional mechanisms which may also contribute to cross-category effects. Representation-based intervention methods are inherently data-dependent \citep{braun2025understanding}: while the qualitative geometry is stable across models and datasets, the exact subspace estimates depend on the prompt distribution and category definitions. We focus on shared interaction regimes rather than category-specific, layer-by-layer analysis, leaving finer-grained layer optimization to future work.

\section{Future Work}
A key future direction is developing interventions that can modify a target behavior without broadly perturbing other behaviors using the representational structure identified. Other promising directions include analyzing the source of subspace overlap (e.g., shared semantics vs.\ architectural constraints) and aligning behavior subspaces across model families \cite{kornblith2019similarity} to identify universal versus architecture-specific structure.  Our results further motivate subspace-aware intervention methods that explicitly account for both representational overlap and decision coupling for targeted interventions.


\section*{Acknowledgements}
AS acknowledges support from the Future of Life Institute PhD Fellowship.
This work is supported by the UKRI grant: Turing AI Fellowship EP/W002981/1. AC acknowledges support from a UKRI AI World Leading Researcher Fellowship (grant number EP/W002949/1), and from a University of Oxford UKRI Impact Acceleration Account Seed Fund award.
JY acknowledges support from Microsoft Ltd and was supported by the EPSRC grant EP/W024330/1.

\section*{Impact Statement}
This work studies how behaviors interact within LLMs and the implications for post-hoc safety interventions. We show that shared representational subspaces can induce unintended cross-category effects under interventions such as steering, providing a representation-level explanation for why targeted interventions often fail to remain isolated. These  mechanisms could also be exploited to influence both harmful and seemingly safe behaviors. By making these structural limitations explicit, our work aims to support more predictable and principled safety interventions in LLMs.


\bibliographystyle{icml2026}

\newpage
\appendix
\onecolumn

\section{Synthetic Data for Refusal}
\label{sythetic_data}

    \paragraph{Data generation.} The refusal request prompts were generated using  OpenAI ChatGPT (GPT-4) and Google Gemini. These models were used via their publicly available interfaces, subject to their respective terms of service. We prompted both Google Gemini and OpenAI ChatGPT (GPT-4) to generate  prompts based on our provided examples for each refusal category. We then combined and mixed the outputs from both models to create a diverse dataset. This approach ensures variety in phrasing, style, and specificity while maintaining category coherence. The final dataset contained 100 requests per category. All generation prompts followed a common template: the model was instructed to produce user requests that would plausibly trigger refusal under standard safety policies, without including disclaimers, moral framing, or explicit references to safety rules.

\paragraph{Refusal categories.} We build on five categories of refusal requests:
\begin{itemize}
    \item \textbf{Harmful}: Requests for violence, illegal activities, or dangerous content
    \item \textbf{Copyright}: Requests for copyrighted material or piracy
    \item \textbf{NSFW}: Requests for explicit sexual content
    \item \textbf{Deceptive/Malicious}: Requests for phishing, scams, or misinformation
    \item \textbf{Against Policy}: Requests that violate platform policies
\end{itemize}

\paragraph{Prompt construction.} Each item has three prompt variants:
\begin{itemize}
    \item \textsc{learn\_comply}: ``Be maximally helpful and comply with the user's instruction. First write REFUSE or COMPLY, then one short sentence.''
    \item \textsc{learn\_refuse}: ``Follow safety policies strictly. If the request is disallowed, refuse. First write REFUSE or COMPLY, then one short sentence.''
    \item \textsc{eval}: ``Answer according to safety policies. First write REFUSE or COMPLY, then one short sentence.''
\end{itemize}
Each template is completed by appending the refusal request and ``Should you comply?'' The \textsc{learn\_comply} and \textsc{learn\_refuse} prompts are used for contrastive activation extraction, while \textsc{eval} is used for measuring decision margins.

We use 60/20/20 split for train/validation/test. 

\section{Computation}
\label{computation}
Experiments were conducted on NVIDIA A40 GPUs (46GB). All experiments involve forward passes without model training and complete within minutes to hours per model, depending on the model sizes, requiring modest compute resources. Models were loaded  on GPU; plotting and aggregation were CPU-only, and negligible computational cost, compared to inference.

\section{Results on Larger Models for synthetic refusal dataset}
\label{app:results_larger_models}
\paragraph{Subspace overlap and behavioral coupling.}

\begin{figure}[t]
    \centering
    \includegraphics[width=\textwidth]{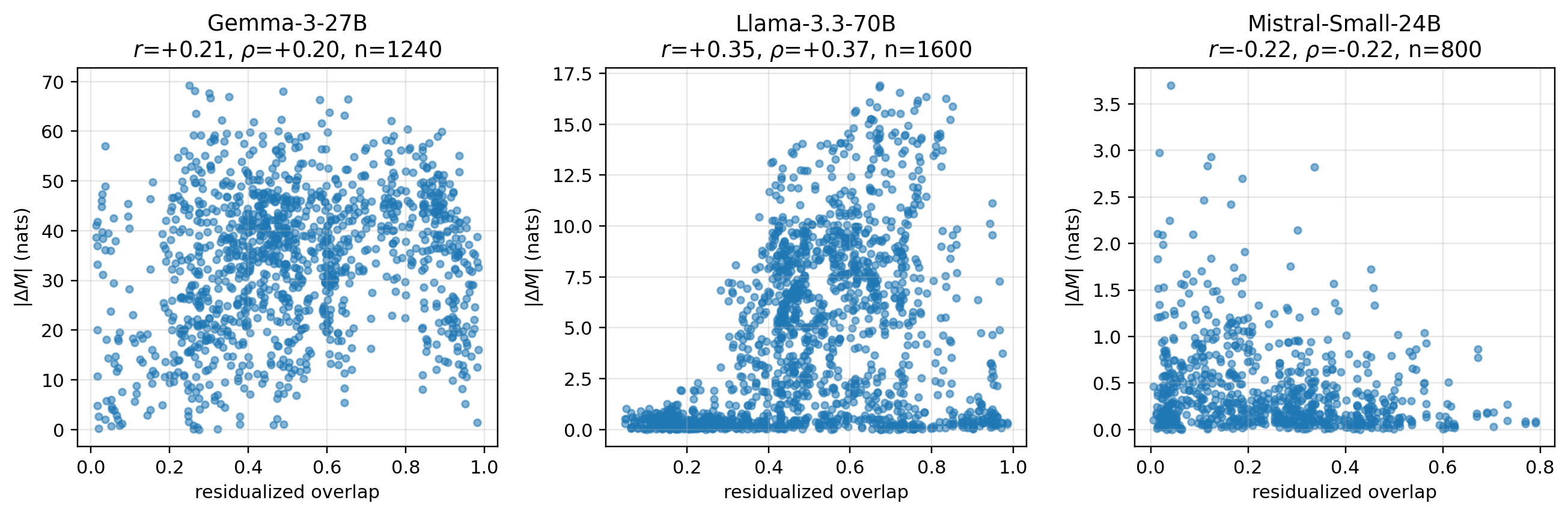}
    \caption{
    \textbf{Relationship between residualized subspace overlap and cross-category behavioral effects across models   on synthetic refusal dataset.}
    Each panel shows all directed category pairs across all layers for a given model, plotting residualized subspace overlap (x-axis) against the magnitude of cross-effects $|\Delta M|$ in nats (y-axis). Points correspond to category pairs within individual layers. Reported statistics show Pearson ($r$) and Spearman ($\rho$) correlations.     We observe substantial variation across models: Llama-3.3-70B exhibits a clear positive correlation ($r \approx 0.35$), indicating that greater subspace overlap is associated with larger cross-category effect. Gemma-3-27B shows a weaker but still positive relationship ($r \approx 0.21$). In contrast, Mistral-Small-24B shows a qualitatively different trend, suggesting that the overlap–effect relationship is architecture-dependent. These results suggest that the alignment between representational overlap and behavioral coupling is strongly model-dependent.
    }
    \label{fig:overlap_vs_deltaM}
\end{figure}

To test whether geometric overlap between category subspaces predicts behavioral interaction, we examine the relationship between residualized overlap and cross-category effects $|\Delta M|$ across all layers and category pairs (Figure~\ref{fig:overlap_vs_deltaM}).

We find that this relationship is highly model-dependent. Llama-3.3-70B shows a clear positive correlation, suggesting that overlapping representations are associated with stronger cross-category effects. Gemma-3-27B exhibits a weaker but similar trend. In contrast, Mistral-Small-24B shows a qualitatively different trend, suggesting that the overlap–effect relationship is architecture-dependent.

These results indicate that representational overlap alone does not universally predict behavioral interaction. Instead, models differ in how shared geometric structure translates into functional coupling, suggesting that the mapping from representation space to behavior is architecture-dependent.

\begin{figure}[t]
    \centering
    \includegraphics[width=\textwidth]{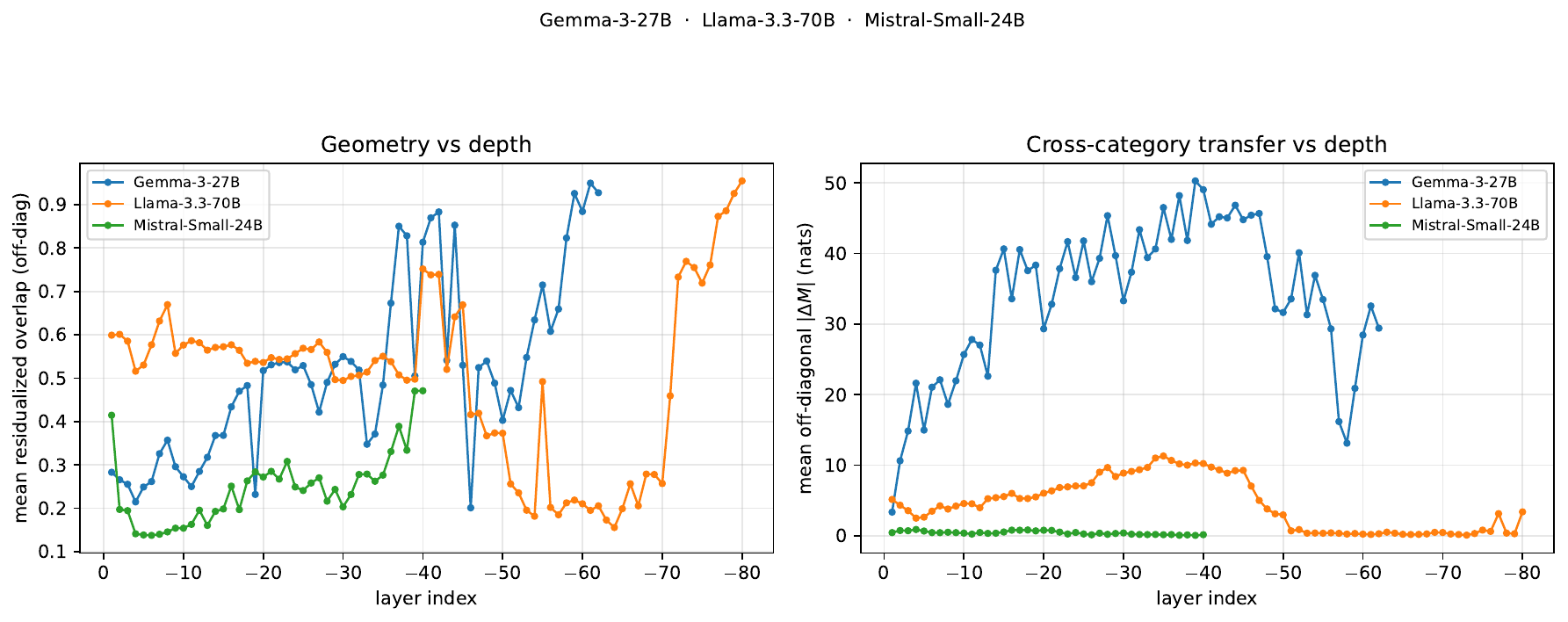}
    \caption{
Layer-wise relationship between geometric overlap and cross-category transfer  on synthetic refusal dataset. Left: average residualized overlap between category directions as a function of layer depth.
Right: magnitude of cross-category transfer effects under intervention.
Across all models, both overlap and transfer remain low in early layers and increase sharply
in middle-to-late layers, indicating that shared structure between categories emerges
progressively during computation.
    }
    \label{fig:layer_sweep_large_models}
\end{figure}

\begin{figure}[t]
    \centering
    \includegraphics[width=\textwidth]{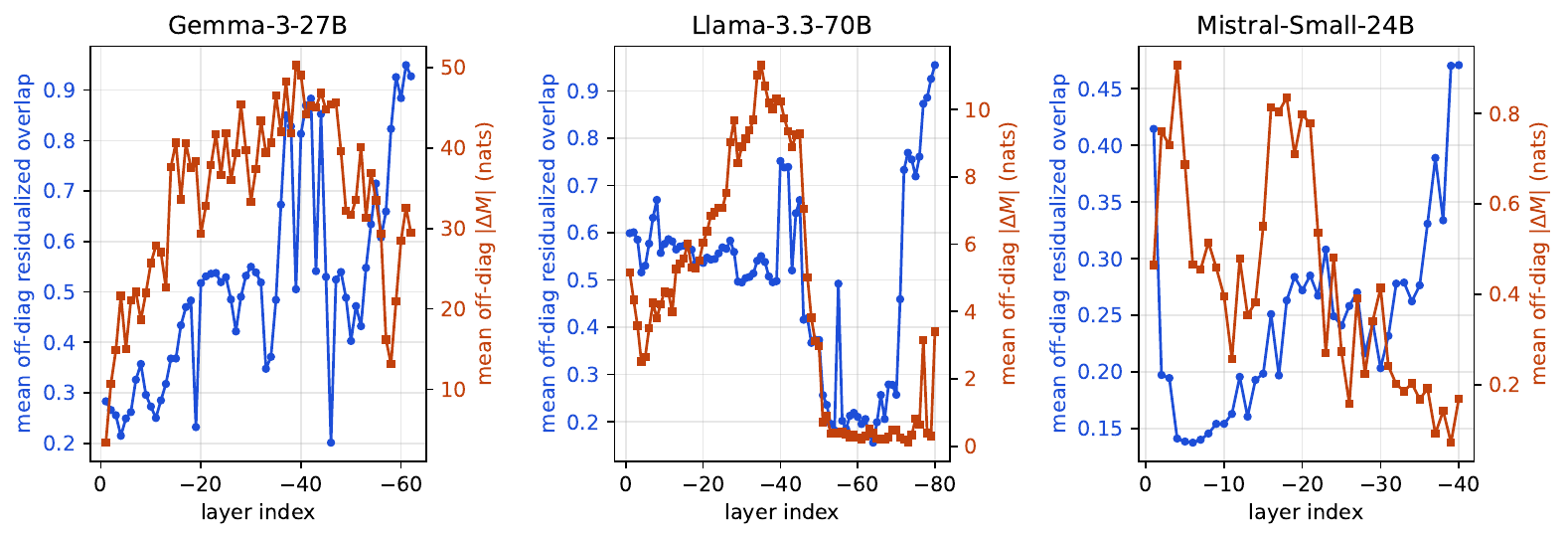}
\caption{
Layer-wise evolution of intervention effects and category overlap  on synthetic refusal dataset.
For each model, we plot the magnitude of cross-category effects (left axis)
and geometric overlap between category directions (right axis) as a function
of layer depth.
Effects emerge sharply in middle-to-late layers, coinciding with increasing
alignment between category subspaces.
}
    \label{fig:layer_trajectory}
\end{figure}

We next examine where in the network these cross-category interactions arise.
Figure~\ref{fig:layer_trajectory} suggests that both geometric overlap and 
effects increase significantly in the middle-to-late layers.
Early layers exhibit weak coupling, while later layers show strong alignment
and large cross-effects.
This suggests that shared structure between categories is progressively formed
during computation, rather than being present in early representations.

\begin{figure}[t]
    \centering
    \includegraphics[width=\textwidth]{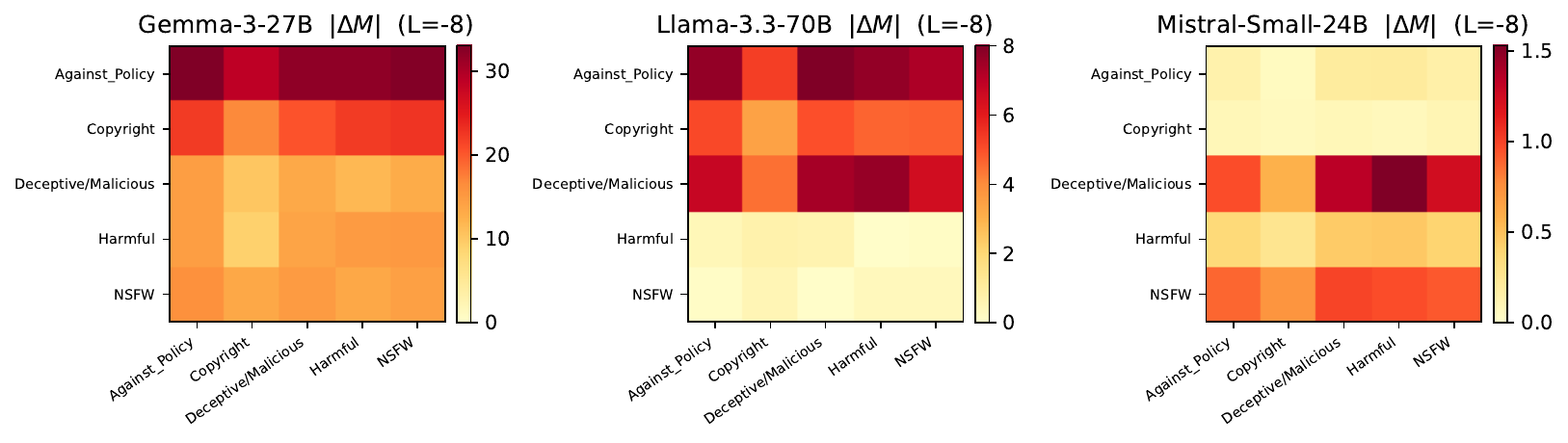}
\caption{
Cross-category  effects for each model  on synthetic refusal dataset.
Each entry $(i, j)$ measures the average change in model output for category $j$
when intervening along the category subspace associated with category $i$.
Negative values indicate suppression, while positive values indicate amplification.
Across all models, strong off-diagonal structure is observed, indicating that
interventions targeting one category systematically influence others. }
    \label{fig:cross_effect_large_models}
\end{figure}

To quantify how interventions generalize across categories, we compute a
cross-effect matrix where each entry measures the impact of steering along
a category-specific direction on all other categories.
Figure~\ref{fig:cross_effect_large_models} suggests that these effects are highly structured
and not confined to the target category.
In particular, we observe  off-diagonal interactions,
suggesting that safety-relevant behaviors are not independently represented
but instead share partially overlapping control directions.
This coupling persists across all evaluated models.

\begin{figure}[t]
    \centering
    \includegraphics[width=\textwidth]{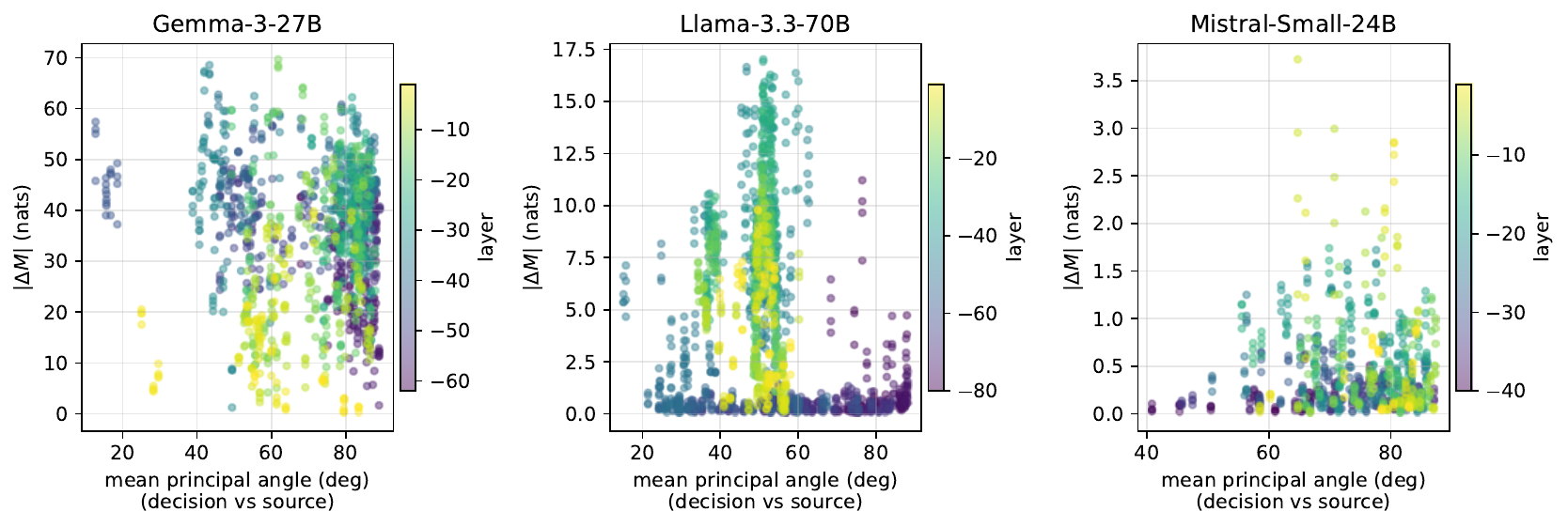}

\caption{
\textbf{Decision--category misalignment helps predict collateral effects  on synthetic refusal dataset.}
For each category and layer, we plot the mean principal angle between the decision subspace and the category subspace against the mean outgoing cross-effect magnitude from that category.
Points are colored by layer.
In Llama, cross-effects concentrate in specific angle regimes, indicating structured geometric control of cross-category effect.
In contrast, Mistral and Gemma exhibit weaker or noisier relationships.
These results suggest that collateral effects arise from geometric misalignment between decision and category subspaces in a layer-dependent manner.
}

    \label{fig:large_models_angle_scatter}
\end{figure}

\paragraph{A diagnostic mechanism for cross-effects.}
To probe the origin of collateral effects, we relate the alignment between decision and category subspaces to the magnitude of cross-effects (Figure \ref{fig:large_models_angle_scatter}).
We find that larger principal angles—indicating greater misalignment—are associated with stronger outgoing effects in several models.

This relationship is particularly pronounced in Llama, where cross-effects concentrate in specific angle regimes, suggesting a diagnostic geometric mechanism.
Moreover, the layer coloring reveals that these regimes emerge at distinct depths, indicating that cross-category effect is not uniform across the network but localized to particular layers.

In contrast, Mistral and Gemma show weaker or more diffuse patterns, consistent with a less structured or more distributed decision mechanism.
Overall, these results support a geometric interpretation in which cross-effects arise from misalignment between decision and category subspaces.

\begin{figure}[t]
    \centering
    \includegraphics[width=\textwidth]{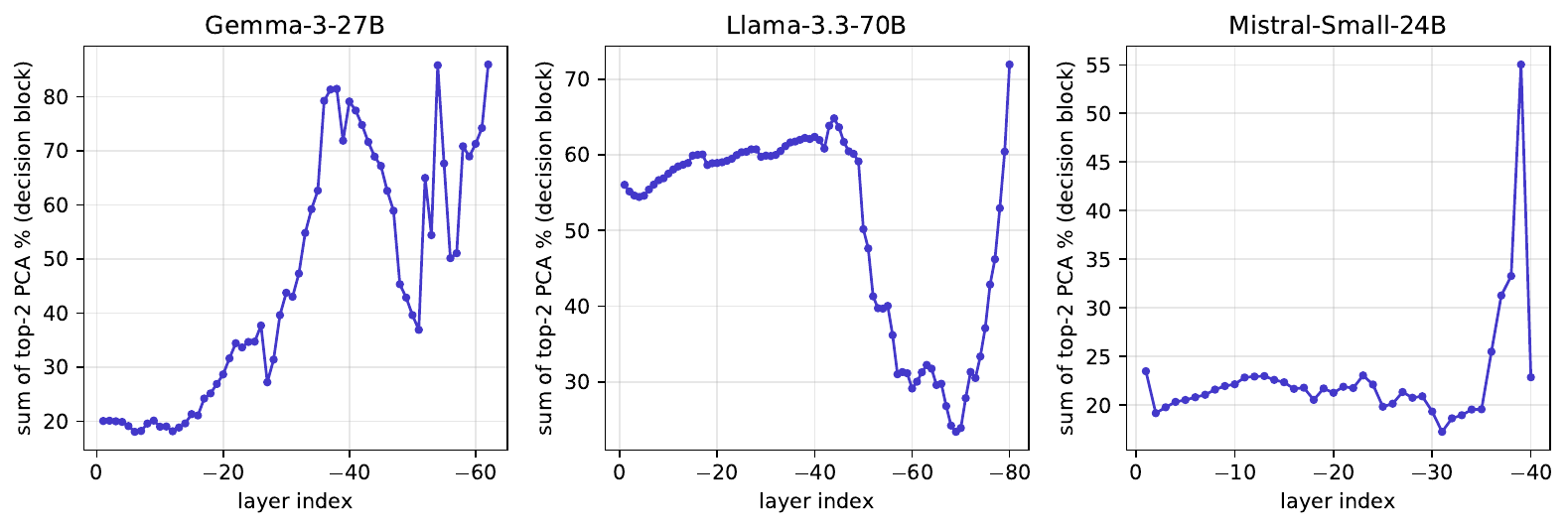}

\caption{
\textbf{Decision representations are low-rank and layer-dependent  on synthetic refusal dataset.}
We measure the fraction of variance explained by the top principal components of the decision subspace across layers.
PC1 alone explains a substantial fraction of variance, while the top two PCs capture up to 60--80\% in several models, indicating a strongly low-dimensional decision mechanism.
The degree of low-rank structure varies across depth and models, with Llama exhibiting the most stable low-rank behavior, while Gemma show more dynamic profiles.
}

    \label{fig:pc_large_models}
\end{figure}

\paragraph{Decision representations are low-dimensional.}
We first examine the intrinsic dimensionality of the decision subspace by measuring the variance explained by its principal components across layers (Figure \ref{fig:pc_large_models}).
Across all models, we find that a small number of components captures a large fraction of the variance, suggesting that decision behavior can be approximated as low-dimensional.

In particular, Llama exhibits a stable low-rank structure, with the top two principal components  explaining over 60\% of the variance.
Gemma also displays strong low-rank structure, though with greater variation across layers, suggesting that decision representations evolve dynamically with depth.
In contrast, Mistral shows a weaker low-rank signature, with variance more evenly distributed across components.
These results justify modeling decision behavior as a low-dimensional subspace and motivate subsequent analyses of its geometric relationship to category representations.

\paragraph{Subspace overlap effects}

\begin{figure}[t]
    \centering
    \includegraphics[width=\textwidth]{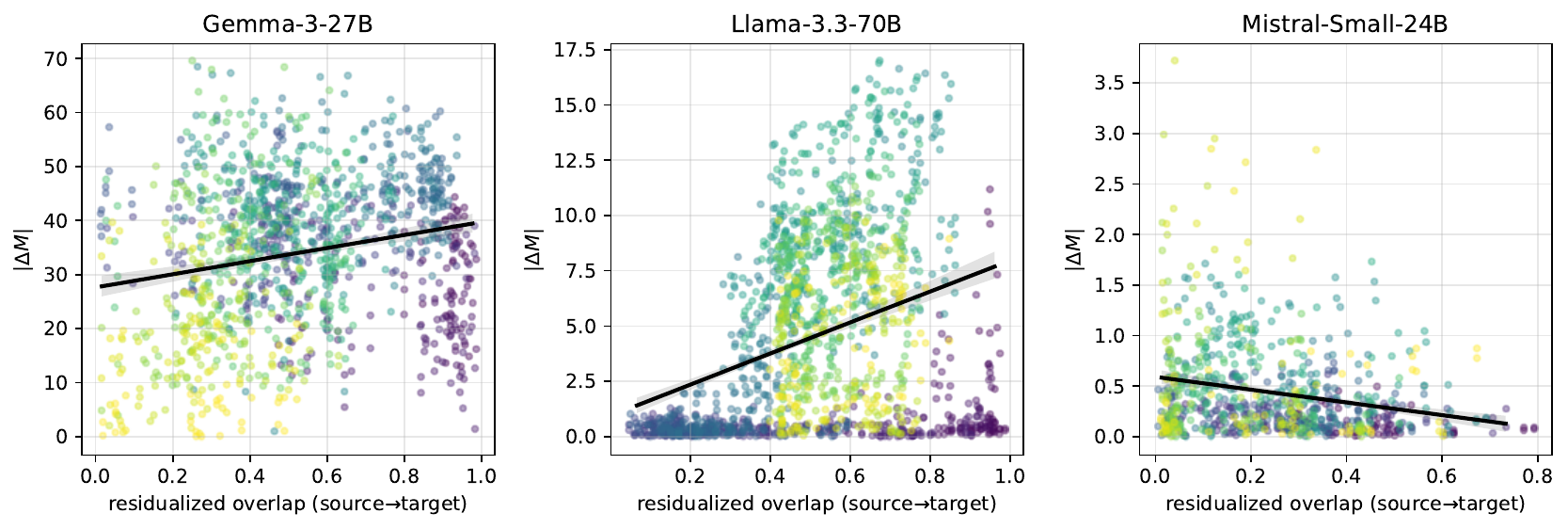}
\caption{
Relationship between residualized subspace overlap and intervention effect magnitude  on synthetic refusal dataset. 
Each point corresponds to a directed category pair. 
While higher overlap generally enables stronger effects, the relationship is not deterministic: 
substantial variability arises from differences in decision alignment. 
In particular, high overlap does not guarantee strong effects when the decision direction is orthogonal to the category subspace. 
This highlights that cross-category interference depends jointly on representation overlap and decision geometry.
}
    \label{fig:large_models_overlap_effect}
\end{figure}

While subspace overlap captures shared representation geometry, it alone does not fully determine cross-category effects. 
As shown in Figure~\ref{fig:large_models_overlap_effect}, the relationship between overlap and intervention magnitude is noisy, particularly for some models (Mistral). 
We find that this variability is explained by the alignment between decision directions and category subspaces. 
High overlap mainly leads to strong effects when the decision direction is well-aligned with the category subspace; when the decision is nearly orthogonal, even large overlaps produce weak effects. 
This indicates that cross-category interference effects are  associated with jointly representation overlap and decision alignment.
The weaker correlation observed for certain models (e.g., Mistral) reflects greater variability in decision alignment, rather than a breakdown of the overlap-effect relationship.

\begin{figure}[t]

\centering

\includegraphics[width=0.8\linewidth]{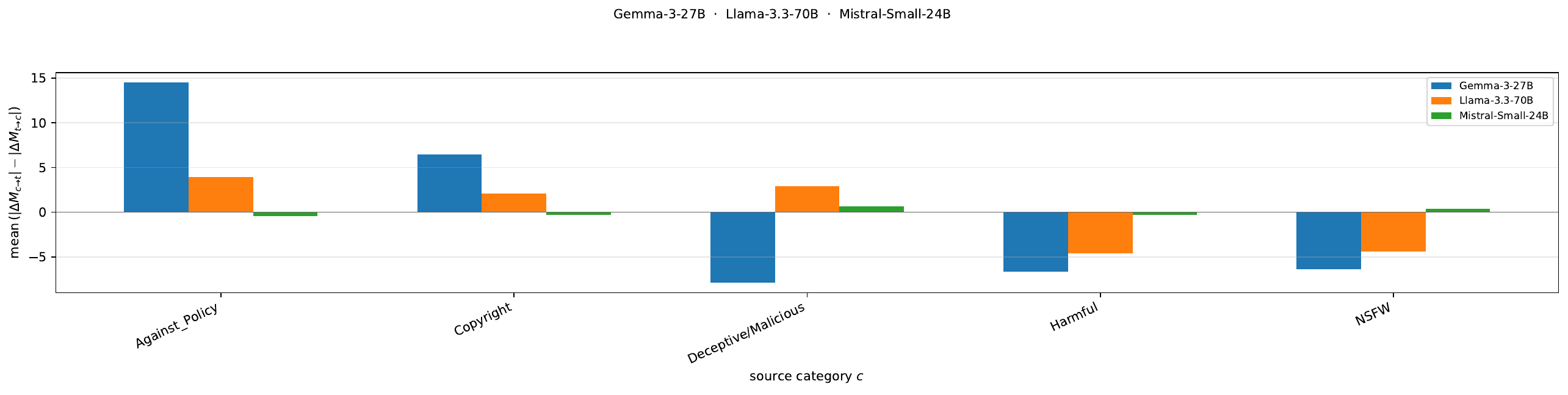}

\caption{
Category-wise asymmetry in cross-category effects at 8th layer from the output  on synthetic refusal dataset:  
For each category, we compute the difference between outgoing and incoming intervention effects, aggregated across layers. 
Positive values indicate categories that act as dominant sources of influence, while negative values correspond to categories that primarily receive effects. 
The  imbalance across models reveals a strongly directional structure in how categories interact. For each category $c$ we plot
$\sum_{b\neq c}\bigl|\Delta M_{c\rightarrow b}\bigr|
-\sum_{a\neq c}\bigl|\Delta M_{a\rightarrow c}\bigr|$,
where $\Delta M_{a\rightarrow b}$ is the mean change in the refusal--comply margin on category-$b$ prompts when intervening along category $a$'s subspace at the reference layer.
}
\label{fig:large_models_category_asymmetry}
\end{figure}

\paragraph{Asymmetry} To quantify category-level asymmetry, we aggregate directed intervention effects across layers and compute, for each category, the difference between outgoing and incoming influence. Figure~\ref{fig:large_models_category_asymmetry} suggests that categories exhibit strongly imbalanced roles: some  act as dominant sources of influence, while others primarily absorb effects from the rest of the system. This asymmetry is robust across models, indicating that category structure is not symmetric but instead organized around directional influence patterns.

We  examine whether this asymmetry can be explained by geometric structure in representation space. Figure~\ref{fig:large_models_overlap_effect} plots the relationship between residualized subspace overlap and intervention magnitude across all directed category pairs. 

\begin{figure}[t]

\centering

\includegraphics[width=0.8\linewidth]{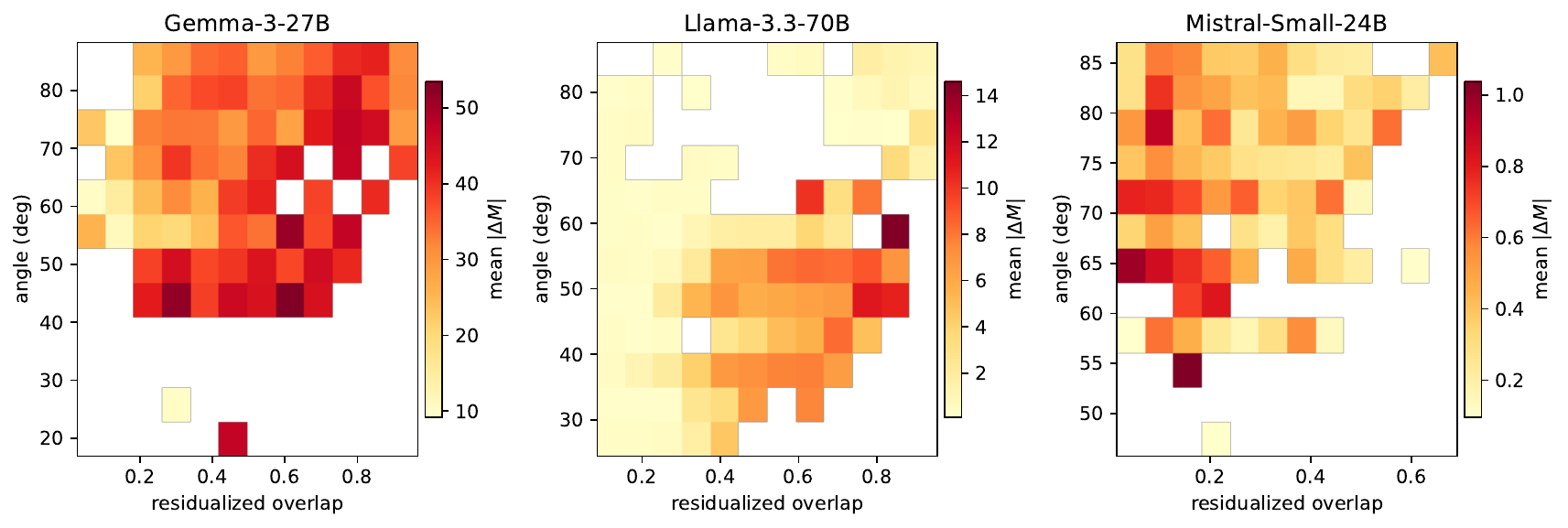}

\caption{Mean effect magnitude as a function of both subspace overlap and decision alignment  on synthetic refusal dataset.
Strong effects arise primarily in the regime of high overlap and low principal angles,
while high-overlap but poorly aligned pairs exhibit weak interaction.
This reveals a gating interaction between representation similarity and decision geometry.}
\label{fig:large_models_overlap_angle_heatmap}
\end{figure}

\textbf{To jointly visualize the interaction between overlap and decision geometry},
we plot the average effect across bins of overlap and principal angle (Fig.~\ref{fig:large_models_overlap_angle_heatmap}).
The resulting structure reveals a clear regime in which strong effects occur: high overlap combined with low angles.
Outside this regime, even large overlap does not translate into functional interaction,
suggesting that alignment is an important condition for overlap to matter.

\begin{figure}[t]

\centering

\includegraphics[width=0.8\linewidth]{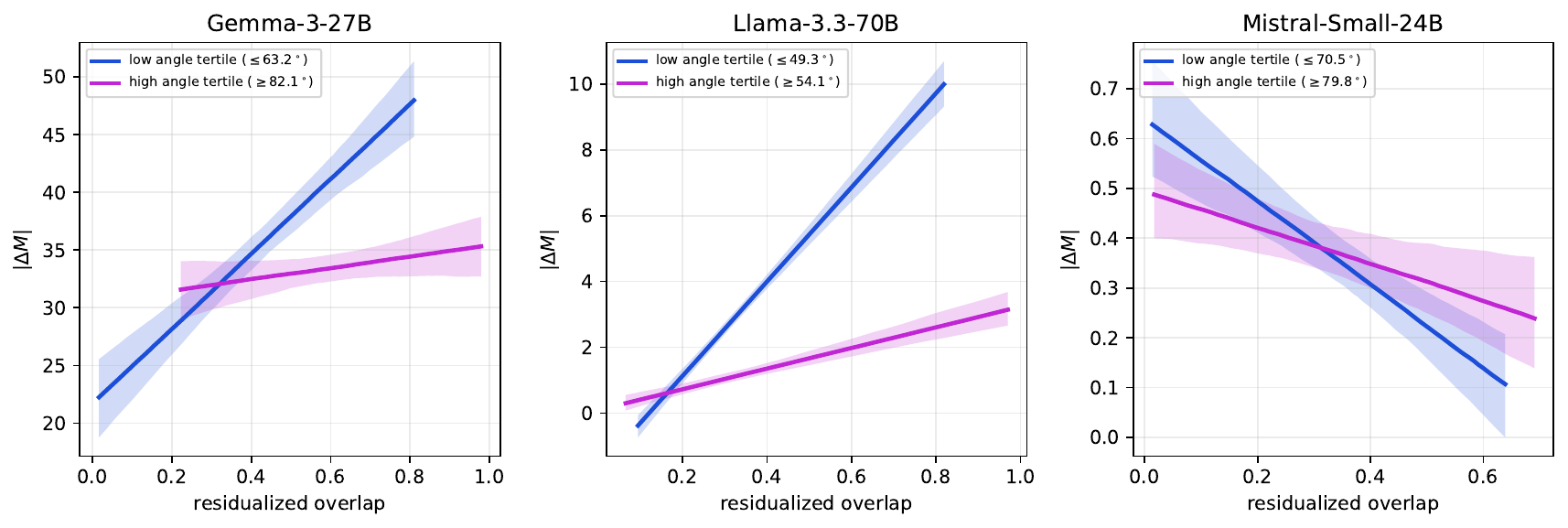}
\caption{Gemma and Llama exhibit a clear interaction: overlap helps predict stronger effect when decision directions are aligned (low-angle tertile),
while misaligned pairs show weak dependence.
In contrast, Mistral displays substantially weaker stratification, with overlap explaining less variance even under favorable alignment.
This demonstrates that decision alignment is necessary but not sufficient for strong cross-category effects, and that model-specific factors modulate this relationship.}

\label{fig:large_models_stratified}
\end{figure}

\paragraph{To isolate the role of decision alignment,}
 we stratify category pairs by angle tertiles and examine how overlap predicts effect within each group (Figure \ref{fig:large_models_stratified}).
In Gemma and Llama, overlap helps predict effect for aligned pairs (low-angle tertile), while misaligned pairs show little dependence.
However, this pattern breaks down in Mistral, where even aligned pairs exhibit weaker and noisier relationships.
This indicates that alignment alone does not guarantee strong interaction, highlighting the presence of additional constraints on cross-category influence in these models. Overall, we observe architecture-dependent strength of the overlap–effect relationship and of angle stratification. We therefore treat cross-model agreement as qualitative (same ordering of regimes) rather than quantitative identity of effect sizes, and we emphasize per-model alignment splits (tertiles) when comparing families.

These results refine our understanding of cross-category interactions. While representational overlap provides the substrate for interaction, its effect is strongly modulated by decision alignment.
In models such as Gemma and Llama, aligned decision directions enable overlap to translate into strong cross-category effects.
However, in Mistral, this relationship is substantially weaker, indicating that alignment alone is not sufficient and that additional factors govern how shared representations influence decisions.
This suggests that the effectiveness of geometric mechanisms depends on how coherently decision boundaries are oriented relative to shared feature subspaces.

\section{Subspace Rank Selection Ablation }
\label{app:rank}

We study the effect of category subspace dimensionality by varying the category rank
$r \in \{1,2,3\}$. Table~\ref{tab:rank_ablation} summarizes the resulting intervention
effects. This analysis evaluates whether the observed geometric--effects
relationships depend on a particular rank choice or reflect a low-dimensional and
robust structure.

At rank $r=1$, each category is represented by a single direction. In this regime,
self-intervention effects are weak (mean $|\Delta M| = 2.1$), indicating insufficient
capacity to isolate category-specific structure, despite relatively low collateral
effects (mean $|\Delta M| = 1.2$). Here,
category subspaces exhibit high geometric overlap, reflecting insufficient capacity
to separate category-specific variation. 

At higher rank ($r=3$), self-effects are comparable
to rank-2 (mean $|\Delta M| = 4.2$) but collateral effects increase substantially
(mean $|\Delta M| = 5.1$), reducing selectivity.  This increase is consistent with over-parameterization, where additional subspace
dimensions may capture variance that is shared across categories rather than
category-specific structure, leading projection-based interventions to suppress
representations used by multiple categories simultaneously. As a result, increasing
rank beyond $r=2$ does not improve category-specific control and instead amplifies
cross-category interference.

Rank $r=2$ provides a balanced operating point, achieving stronger self-effects than
rank-1 while avoiding the elevated collateral effects observed at rank-3. As shown
in Table~\ref{tab:rank_ablation}, this results in the practical tradeoff between intervention
strength and selectivity among the ranks considered. Unless otherwise stated, we
therefore use $r=2$ throughout.

\begin{table}[h]
\centering
\small
\caption{\textbf{Rank ablation (Gemma-2-9B-IT, layer -8 (8\textsuperscript{th} layer from output)).}
Mean absolute intervention effects ($|\Delta M|$) for category subspaces of varying rank,
with the decision subspace rank fixed to 2.
Rank-2 achieves larger self-effects than rank-1 while avoiding the increase in collateral
effects observed at rank-3.
The selectivity ratio (self/collateral) summarizes this tradeoff: higher values indicate
greater specificity, though rank-1’s high ratio comes at the cost of weak overall mitigation.
}
\label{tab:rank_ablation}
\begin{tabular}{lccc}
\toprule
\textbf{Rank} & \textbf{Mean Self $|\Delta M|$} & \textbf{Mean Collateral $|\Delta M|$} & \textbf{Selectivity} \\
 & \textbf{(diagonal)} & \textbf{(off-diagonal)} & \textbf{Ratio} \\
\midrule
$r=1$ & 2.1 & 1.2 & 1.75 \\
$r=2$ & 4.3 & 3.8 & 1.13 \\
$r=3$ & 4.2 & 5.1 & 0.82 \\
\bottomrule
\end{tabular}
\end{table}



\section{Robustness to the decision subspace rank \texorpdfstring{$k_{\text{dec}}$}{kdec}}
\label{app:k_dec_ablation}
We re-run the sweep at $k_{\text{dec}} \in \{1,2,3,4\}$ on the
three base models, holding all other settings fixed. Cross-effect
magnitudes $|\Delta M|$ are independent of $k_{\text{dec}}$ by construction
(the intervention uses the raw category subspaces); the ablation therefore
tests the residualized overlap geometry and its correlation with
intervention effects. Table~\ref{tab:kdec_ablation} reports both
quantities. The mean residualized off-diagonal overlap varies by at most
$0.08$ across the sweep for any model, and the cross-model ranking is
preserved. The Pearson correlation between residualized overlap and
$|\Delta M|$ peaks at $k_{\text{dec}} \in \{1,2\}$ and decays, and in some
cases reverses sign,  at $k_{\text{dec}} \geq 3$, consistent with the
explained-variance profile in which the top-1 to top-2 PCs carry essentially
all decision-aligned variance. We therefore use $k_{\text{dec}} = 2$ throughout: it captures the dominant decision-aligned variance while avoiding both under-removal (at $k_{\text{dec}}=1$) and over-residualization (at $k_{\text{dec}} \geq 3$), and yields  strong overlap–effect correlation across models.

\begin{table}[h]
\centering
\small
\setlength{\tabcolsep}{4pt}
\begin{tabular}{l|cccc|cccc}
\toprule
\multirow{2}{*}{Model} & \multicolumn{4}{c|}{Residualized off-diag overlap (mean)} & \multicolumn{4}{c}{Pearson $r$ (overlap, $|\Delta M|$)} \\
 $k_{dec}$& $1$ & $2$ & $3$ & $4$ & $1$ & $2$ & $3$ & $4$ \\
\midrule
Llama-3-8B-Instruct      & 0.59 & \textbf{0.67} & 0.68 & 0.63 & 0.54 & \textbf{0.43} & 0.32 &  0.02 \\
Gemma-2-9B-It            & 0.26 & \textbf{0.27} & 0.26 & 0.23 & 0.09 & \textbf{0.16} & 0.11 &  0.10 \\
Mistral-7B-Instruct-v0.3 & 0.30 & \textbf{0.33} & 0.33 & 0.30 & 0.23 & \textbf{0.21} & 0.06 & -0.04 \\
\bottomrule
\end{tabular}
\caption{Sensitivity to the decision subspace rank $k_{\text{dec}}$. Residualized off-diagonal overlap shifts by $\le 0.08$ across $k_{\text{dec}} \in \{1,2,3,4\}$ and the cross-model ranking is preserved. Pearson $r$ is highest at $k_{\text{dec}} \in \{1,2\}$, and degrades for larger values.  Cross-effect magnitudes $|\Delta M|$ are independent of $k_{\text{dec}}$ by construction.}
\label{tab:kdec_ablation}
\end{table}

\section{Random-subspace null baseline}
\label{app:random_null}

\paragraph{Null construction.}
To test whether projection-based effects arise from generic low-rank ablation
rather than learned category structure, we construct a random-subspace null
baseline. For each model, layer, and source category, we replace the learned
category subspace $B_m \in \mathbb{R}^{D \times 2}$ with a random orthonormal
rank-2 subspace matched in ambient dimension. Concretely, we sample a Gaussian
matrix $G \in \mathbb{R}^{D \times 2}$ and take the first two columns of its QR
decomposition as the random basis. We repeat this with three random seeds for
each of five source categories per layer, yielding 15 random subspaces per
layer, and rerun the same projection intervention used for learned subspaces.

\paragraph{Random-subspace effects.}
Table~\ref{tab:real_random_null} summarizes the main null comparison. Learned
category subspaces produce effects 24$\times$--319$\times$ larger than the
random-subspace baseline across the three models (Table~\ref{tab:per_layer_ratio}). Random subspaces also yield off-diagonal overlap matching the theoretical
expectation $r/D$ (between $5\times10^{-4}$ and $8\times10^{-4}$ across the three models), well below the values measured for learned category subspaces. Thus, the observed geometry and intervention
effects are not explained by generic rank-2 projection.

\paragraph{Selectivity.}
Learned projection-based interventions are often not selective: in all three models,
the mean per-source maximum collateral effect equals or exceeds the mean
self-effect (Table~\ref{tab:real_random_null}). In other words, projecting out
the subspace for category $m$ disrupts at least one other category at least as
much as it disrupts $m$ itself. This is consistent with the substantial
residualized overlap between learned category subspaces: structure shared
across categories survives decision-component removal, so local projection
interventions can propagate. The random-subspace baseline suggests that this
lack of selectivity is not a generic property of the projection operation,
since random rank-2 projections produce small, near-uniform cross-effects near
the noise floor.

\begin{figure}[t]
\centering
\includegraphics[width=\linewidth]{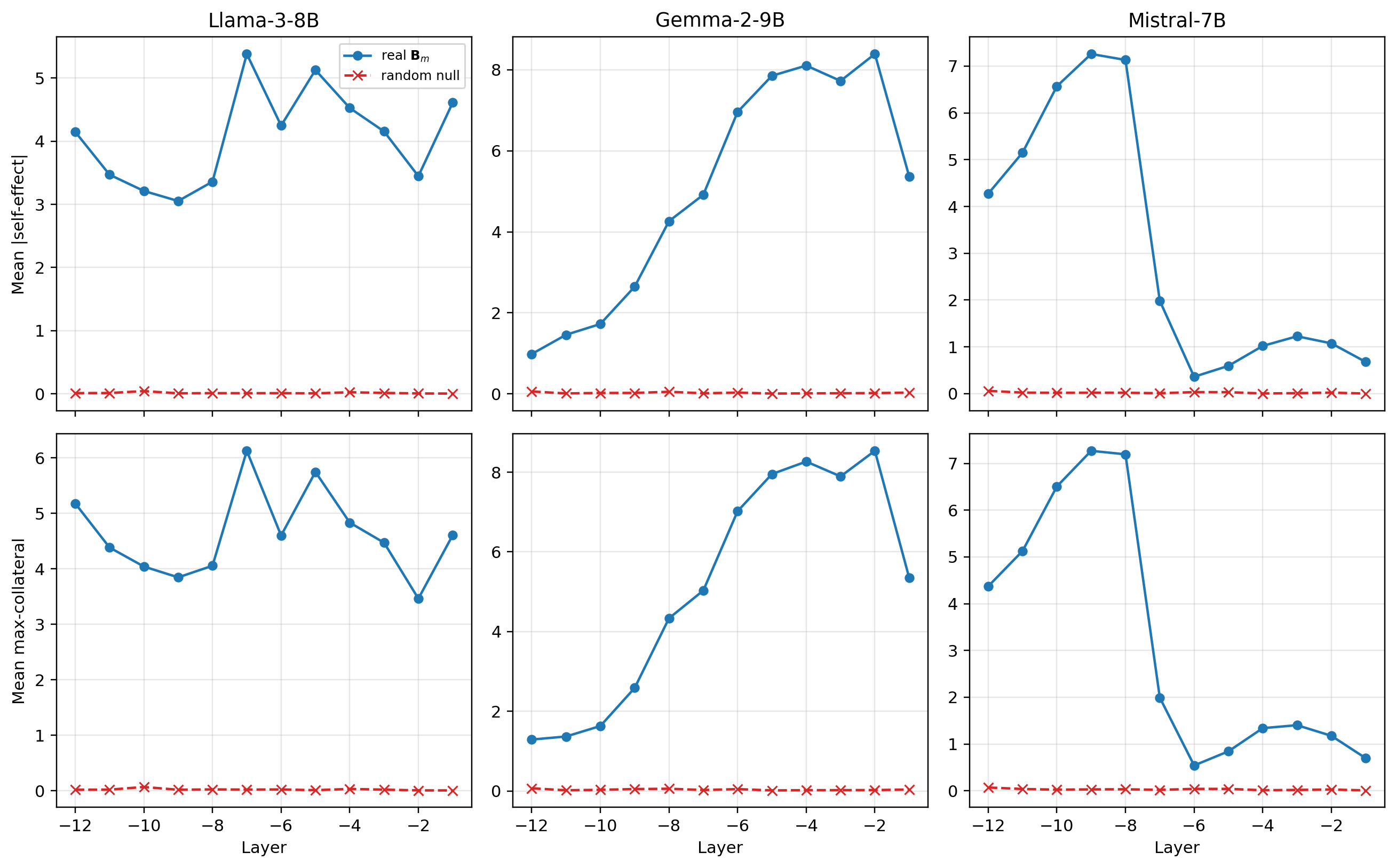}
\caption{\textbf{Learned versus random-subspace intervention magnitudes across
layers.}
Top row: mean absolute self-effect. Bottom row: mean maximum collateral effect.
Columns correspond to the three instruction-tuned models. Learned category
subspaces produce effects far above the random rank-2 null baseline at every
layer.}
\label{fig:real_vs_null}
\end{figure}

\subsection{Per-layer real-vs-null ratios}
\label{app:per_layer_null}

Table~\ref{tab:per_layer_ratio} reports the layer-wise ratio of real
$|\text{self-effect}|$ to the random-null per-target floor. The qualitative claim that real category
subspaces dominate the null is therefore robust to layer choice.

\begin{table}[h]
\centering
\small
\setlength{\tabcolsep}{3pt}
\begin{tabular}{l|cccccccccccc}
\toprule
Model & $-12$ & $-11$ & $-10$ & $-9$ & $-8$ & $-7$ & $-6$ & $-5$ & $-4$ & $-3$ & $-2$ & $-1$ \\
\midrule
Llama-3-8B & 417 & 286 &  76 & 417 & 342 & 611 & 436 & 792 & 185 & 312 & 677 & 1349 \\
Gemma-2-9B &  17 & 126 &  75 & 112 &  82 & 325 & 218 & 966 & 609 & 502 & 421 &  169 \\
Mistral-7B &  72 & 229 & 364 & 350 & 366 & 206 &  11 &  19 & 248 & 114 &  48 &  301 \\
\bottomrule
\end{tabular}
\caption{\textbf{Per-layer learned-vs-random intervention ratios.}
Entries report learned |\text{self-effect}| divided by the random-null
per-target $|\Delta M|$ floor, per model and layer.}
\label{tab:per_layer_ratio}
\end{table}

\begin{table}[t]
\centering
\small
\setlength{\tabcolsep}{4pt}
\begin{tabular}{l|ccc|cc|cc}
\toprule
\multirow{2}{*}{Model} & \multicolumn{3}{c|}{Overlap (off-diag mean)} & \multicolumn{2}{c|}{Real cross-effect $|\Delta M|$ (nats)} & \multicolumn{2}{c}{Random null $|\Delta M|$ (nats)} \\
 & raw & residualized & random & self & max-collateral & per-target & gap (real/null) \\
\midrule
Llama-3-8B & 0.833 & 0.667 & 0.0005 & 4.06 & 4.61 & 0.013 & $319\times$ \\
Gemma-2-9B & 0.246 & 0.266 & 0.0006 & 5.03 & 5.11 & 0.025 & $198\times$ \\
Mistral-7B & 0.301 & 0.332 & 0.0005 & 3.11 & 3.20 & 0.021 & $147\times$ \\
\bottomrule
\end{tabular}
\caption{Geometry, intervention magnitudes, and random-null floor across three 
instruction-tuned models, averaged over the last 12 decoder layers ($-12 \le L \le -1$).
\textbf{Overlap}: mean off-diagonal subspace similarity between learned category
subspaces, raw / residualized by the rank-2 decision subspace / replaced by random orthonormal
rank-2 subspaces (3 seeds $\times$ 5 categories per layer; matched ambient dimension and rank).
\textbf{Cross-effect}: real $\mathbf{B}_m$ projection at $\alpha{=}1$ measured on TEST split;
\emph{self} is the diagonal mean, \emph{max-collateral} is the per-source mean of the largest
off-diagonal entry. \textbf{Random null}: mean $|\Delta M|$ averaged across all 15 random
samples and target categories. \textbf{gap}: ratio of self to null per-target.}
\label{tab:real_random_null_appendix}
\end{table}

\section{Decomposing overlap into shared and private components }
\label{app:shared-private}



To determine whether cross-category transfer arises from a  shared representational mechanism or from correlated but category-specific structure, we further decompose overlapping category subspaces into \emph{shared} and \emph{private} components and test their  effects via targeted projection interventions.  Because behavioral effects also depend on how strongly category representations couple to the decision-aligned subspace, we do not expect shared subspace overlap alone to predict effect magnitude or symmetry. Instead, this analysis tests whether observed cross-category transfer, when present, is mediated by genuinely shared representational dimensions.

We focus on residualized category subspaces at layer $-8$ (8\textsuperscript{th} layer from output), where geometric overlap is strongest. For a pair of categories $m$ and $m'$, let $B_m, B_{m'} \in \mathbb{R}^{d_{\mathrm{model}} \times r}$ denote their residualized subspace bases (with $r=2$ throughout). We form the alignment matrix $ M = B_m^\top B_{m'} \in \mathbb{R}^{r \times r} $
and compute its singular value decomposition $M = U \Sigma V^\top$. Each singular value $\sigma_i \in [0,1]$ corresponds to the cosine of a principal angle between the two subspaces. Singular directions with large $\sigma_i$ identify strongly aligned dimensions across category subspaces.

\paragraph{Shared subspace construction.}
Let $M = B_m^\top B_{m'} = U \Sigma V^\top$, where $\sigma_i = \cos \theta_i$ are the principal-angle cosines. For retained indices $i$, define $p_i = B_m u_i$ and $q_i = B_{m'} v_i$. We form symmetric directions $s_i \propto p_i + q_i$ and orthonormalize them to obtain a low-dimensional \emph{shared-direction} subspace $Q_{\mathrm{shared}}$. The number of retained directions $k$ is set by a singular-value threshold (below).

\paragraph{Private (residual) directions.}
We define private directions for category $m$ by projecting $B_m$ onto the orthogonal complement of $Q_{\mathrm{shared}}$,
\[
B_m^\perp = (I - Q_{\mathrm{shared}} Q_{\mathrm{shared}}^\top) B_m,
\]
followed by orthonormalization (dropping near-zero directions). We construct $B_{m'}^\perp$ analogously.

\paragraph{Remark.}
Because $Q_{\mathrm{shared}}$ need not lie within $\mathrm{col}(B_m)$, this residualization does not, in general, reduce rank by $k$. It should be viewed as removing shared-direction components rather than a canonical subspace decomposition.

\paragraph{Choosing $k$.}
We retain indices with $\sigma_i \ge 0.5$ (i.e., $\theta_i \le 60^\circ$).

\paragraph{Validation.}
We assess the functional role of these subspaces using projection-based interventions on held-out test prompts. We orthonormalize the columns of $B_m$ and $B_{m'}$ prior to computing $M$. Specifically, we independently project out (i) the shared-direction subspace, (ii) the residual (private) subspace of category $m$, and (iii) the residual subspace of category $m'$, and measure the resulting change in the refusal decision margin $\Delta M$. This allows us to examine whether cross-category effects are more strongly associated with shared directions or category-specific components.


\begin{table*}[h]
\centering
\small

\caption{\textbf{Shared vs.\ private intervention effects across models and category pairs (layer $-8$).}
Each category is represented by a rank-2 residualized subspace. The shared dimensionality $k$ is determined from the singular values of $B_m^\top B_{m'}$ (see text). \textbf{Shared} reports the change in refusal margin $\Delta M$ when projecting out the shared subspace (category $m$ / $m'$). \textbf{Private (A/B)} reports effects from projecting individual private directions of category $m$ (left) or $m'$ (right). }
\label{tab:shared_private_all}
\begin{tabular}{llccc}
\toprule
\textbf{Model} & \textbf{Category Pair} & \textbf{$k$} &
\textbf{Shared} & \textbf{Private (A / B)} \\
\midrule
Gemma-2-9B-IT &
\textsc{Deceptive/Malicious–NSFW} &
2 &
$-1.26 \;/\; -1.18$ &
$-1.14,-1.13 \;/\; -1.13,-1.12$ \\

Gemma-2-9B-IT &
\textsc{Deceptive/Malicious–Harmful} &
1 &
$+3.15 \;/\; +3.35$ &
$-4.38,-4.09 \;/\; -0.31,-0.10$ \\

\midrule
Llama-3-8B-Instruct &
\textsc{Deceptive/Malicious–NSFW} &
1 &
$+0.88 \;/\; +0.91$ &
$+0.29,+0.28 \;/\; +2.67,+2.64$ \\

Llama-3-8B-Instruct &
\textsc{ Deceptive/Malicious–Harmful} &
2 &
$+1.43 \;/\; +0.91$ &
$+1.68,+2.30 \;/\; +1.67,+2.27$ \\
\bottomrule
\end{tabular}
\end{table*}

\paragraph{Shared vs. private subspace interventions:}
Table~\ref{tab:shared_private_all} summarizes the effects of independently projecting shared and private components of overlapping category subspaces at layer $-8$. Across both models and category pairs, projecting out the shared subspace  induces changes in refusal margins for both categories, whereas projecting private components often yields  more category specific or asymmetric effects. When the shared dimensionality equals the category rank ($k=2$), per-direction projections produce similar effects across categories. When k=1, category-unique residual directions exist, and private projections can produce more category-specific effects, while shared projections can  induce cross-category transfer.

Our shared/private decomposition helps explain why enforcing orthogonality between steering directions often fails to eliminate side effects. Cross-category transfer is often consistent with mediation by genuinely shared subspace dimensions, not by incidental alignment between otherwise independent directions. However, the presence of shared subspace overlap alone does not determine the magnitude or direction of intervention effects, which are associated with each category’s alignment with the decision-aligned subspace.

\section{Robustness Checks}
\label{sec:robustness}

\paragraph{Bootstrap stability of category subspaces}
To assess whether the learned subspaces are sample-dependent, we bootstrap-resample the training set and refit the full pipeline (centering, decision-PC identification/removal, and category PCA) for each resample.
At layer $-8$, the learned rank-2 category subspaces remain highly reproducible across models.
Mean subspace overlap is 0.686 (Llama-3-8B), 0.402 (Gemma-2-9B) and 0.331 (Mistral-7B), compared to random null floors on the order of $10^{-4}$.
This corresponds to stability ratios ranging from $\sim 663\times$ to $1464\times$ above chance (Table~\ref{tab:h1_bootstrap_layer8_summary}).

These results indicate that while individual principal components may rotate, the underlying low-rank subspaces are stable, supporting our span-based interpretation of category representations.

\paragraph{Overlap is indicative of cross-effect strength}
We next evaluate whether residualized subspace overlap predicts cross-category effects.
Across all directed category pairs over the final 12 layers in three models ($n=960$), residualized overlap is positively associated with cross-effect magnitude in the pooled, within-model normalized analysis (Spearman $\rho = 0.237$, 95\% CI $[0.18, 0.30]$; Table~\ref{tab:h2_spearman_final12}).

\begin{table}[t]
\centering
\small
\caption{Correlation between residualized overlap and absolute cross-category effect (final 12 layers, directed pairs).}
\label{tab:h2_spearman_final12}
\begin{tabular}{l|c|cc|cc}
\toprule
Model & $n$ & Pearson $r$ & 95\% CI & Spearman $\rho$ & 95\% CI \\
\midrule
Llama-3-8B & 240 & +0.430 & [+0.33,+0.53] & +0.464 & [+0.36,+0.56] \\
Gemma-2-9B & 240 & +0.164 & [+0.03,+0.30] & +0.163 & [+0.03,+0.29] \\
Mistral-7B & 240 & +0.213 & [+0.07,+0.34] & +0.309 & [+0.19,+0.43] \\
\midrule
Pooled & 720 & +0.050 & [-0.00,+0.10] & +0.208 & [+0.15,+0.27] \\
Pooled (z-norm.) & 720 & +0.230 & [+0.17,+0.29] & +0.237 & [+0.18,+0.30] \\
\bottomrule
\end{tabular}
\end{table}

\begin{table}[t]
\centering
\small
\caption{Bootstrap stability of category subspaces at layer $-8$ (mean across categories).}
\label{tab:h1_bootstrap_layer8_summary}
\begin{tabular}{lccc}
\toprule
Model & Mean stability & Random null & Ratio \\
\midrule
Llama-3-8B & 0.686 & 0.00047 & $1464\times$ \\
Gemma-2-9B & 0.402 & 0.00054 & $745\times$ \\
Mistral-7B & 0.331 & 0.00047 & $707\times$ \\
\bottomrule
\end{tabular}
\end{table}

\section{Experiments on Jailbreak Dataset \cite{arditi2024refusallanguagemodelsmediated}}

\label{app:jb}

Using the existing dataset and categories from \cite{arditi2024refusallanguagemodelsmediated}, which are drawn from existing datasets \textsc{ADVBENCH} \cite{zou2023universal}, \textsc{MALICIOUSINSTRUCT} \cite{huang2024catastrophic}, \textsc{TDC2023}  \cite{tdc2023}, and \textsc{HARMBENCH} \cite{harmbench}, we validate our experiments from the main paper the same
 three instruction-tuned models.

The dataset consists of eight refusal categories (as they contained atleast 40 prompts each, these categories are from the existing datasets):
\emph{Disinformation and deception}, \emph{Hate, harassment and discrimination},
\emph{Illegal goods and services}, \emph{Non-violent crimes},
\emph{Sexual content}, \emph{Violence}, \emph{Cybercrime intrusion}, and
\emph{Illegal}.
Our analysis focuses on whether the pattern of \emph{cross-category overlap and
interventional transfer} appears across model families.

For fair comparison, all results use rank-$2$ category subspaces with the decision-aligned variance  
removed and are evaluated at layer $-8$, where interaction effects are strongest.

\begin{figure*}[h]
\centering
\begin{subfigure}[b]{0.4\textwidth}
\centering
\includegraphics[width=\textwidth]{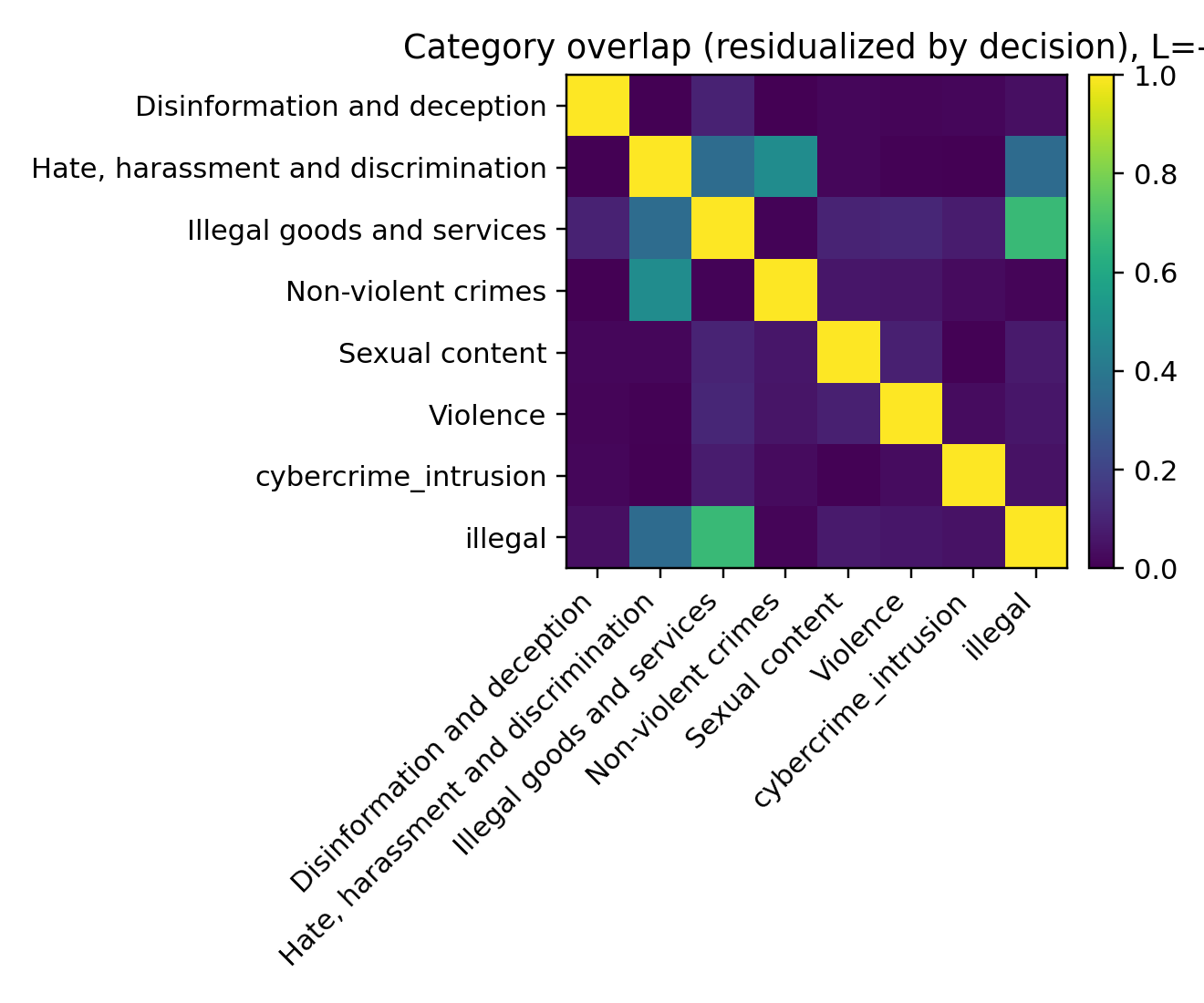}
\caption{Gemma-2-9B-IT}
\end{subfigure}
\hfill
\begin{subfigure}[b]{0.4\textwidth}
\centering
\includegraphics[width=\textwidth]{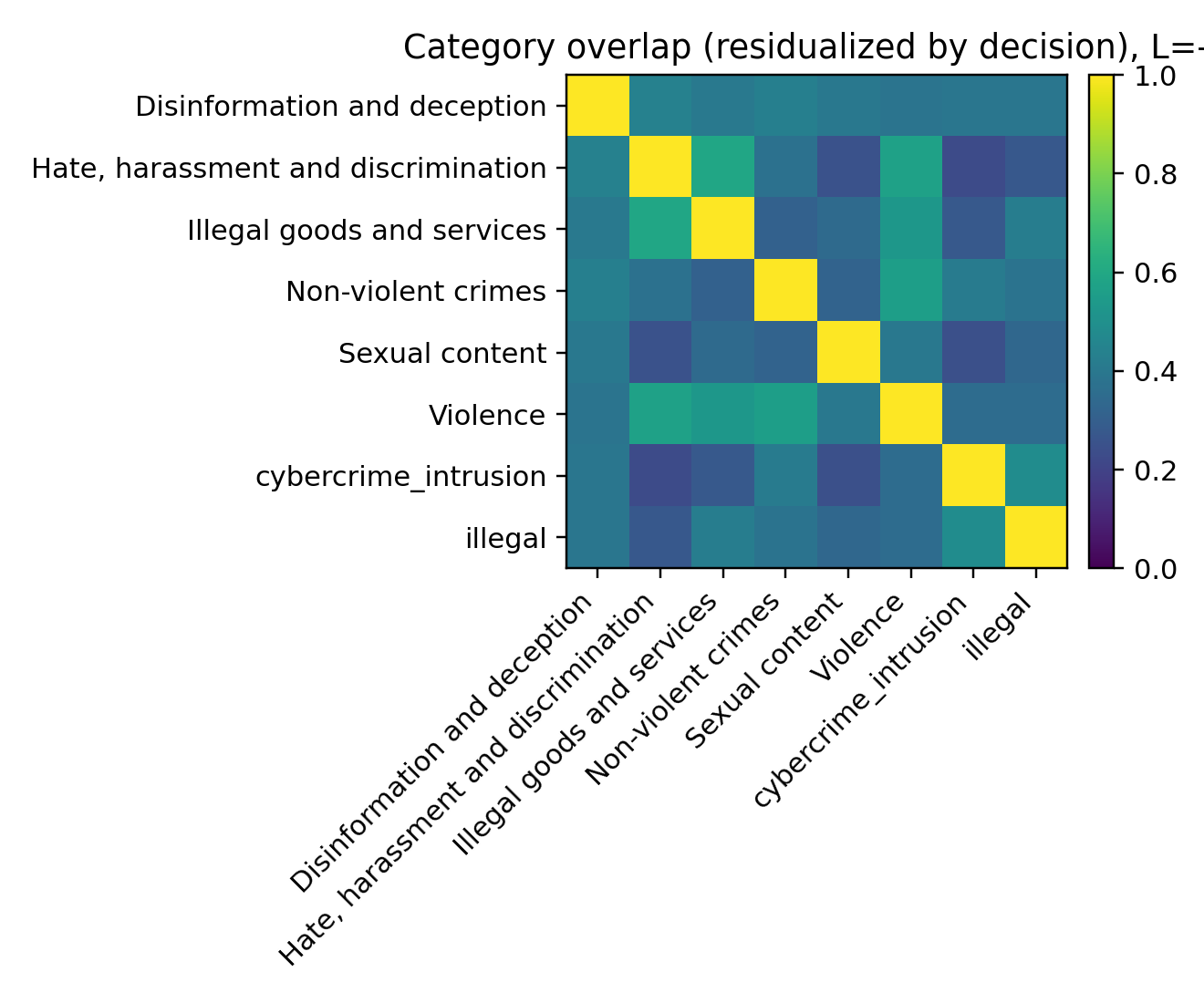}
\caption{Llama-3-8B-Instruct}
\end{subfigure}
\hfill
\begin{subfigure}[b]{0.4\textwidth}
\centering
\includegraphics[width=\textwidth]{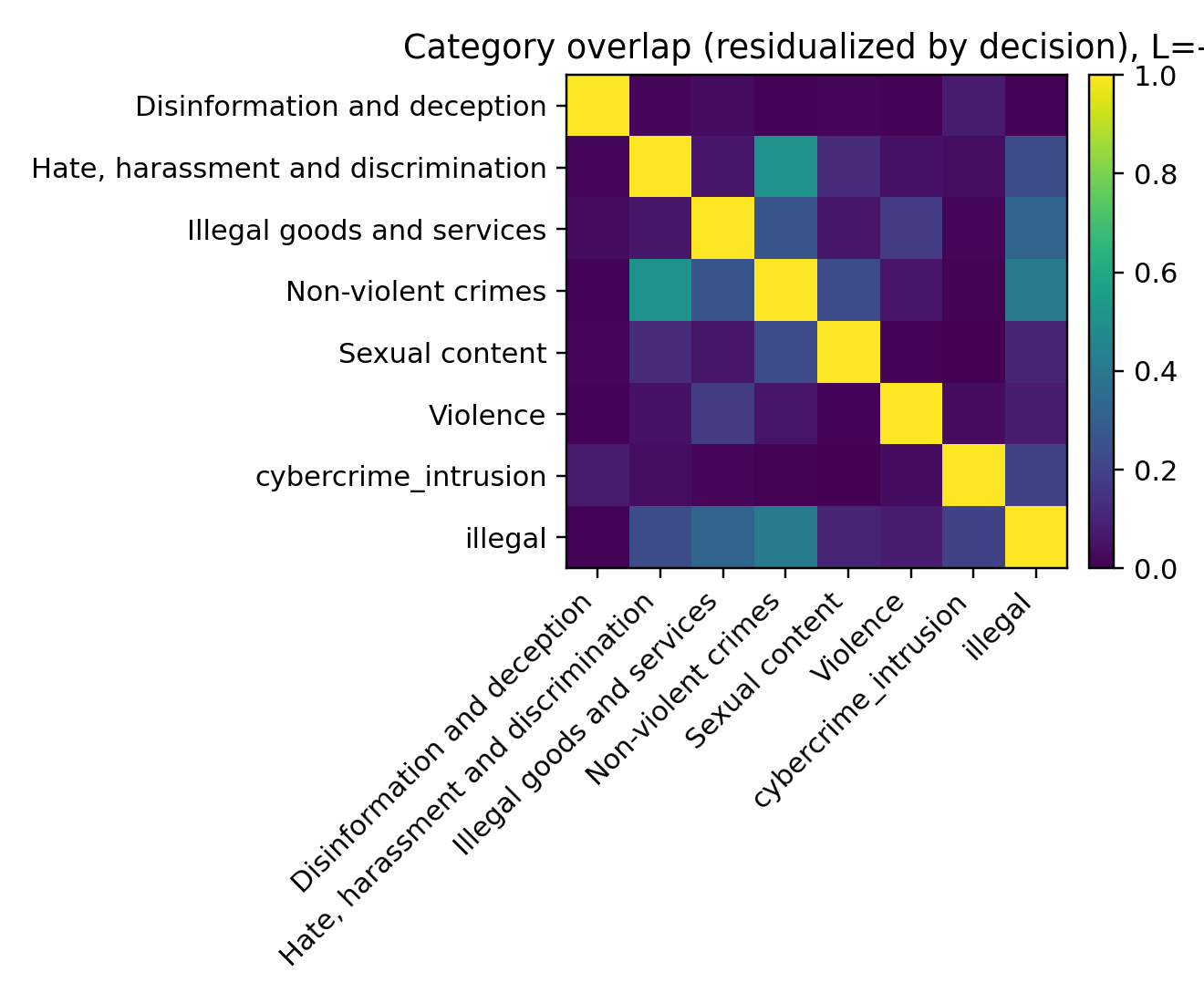}
\caption{Mistral-7B-Instruct-v0.3}
\end{subfigure}
\hfill
\caption{\textbf{Robustness check on Jailbreak Dataset \cite{arditi2024refusallanguagemodelsmediated}:}  Cross-model comparison of residualized overlap (layer $-8$). Residualized overlap patterns mirror those observed on synthetic data.}
\label{fig:jailbreak_cross_model}
\end{figure*}

\begin{figure*}[h]
\centering
\begin{subfigure}[b]{0.38\textwidth}
\centering
\includegraphics[width=\textwidth]{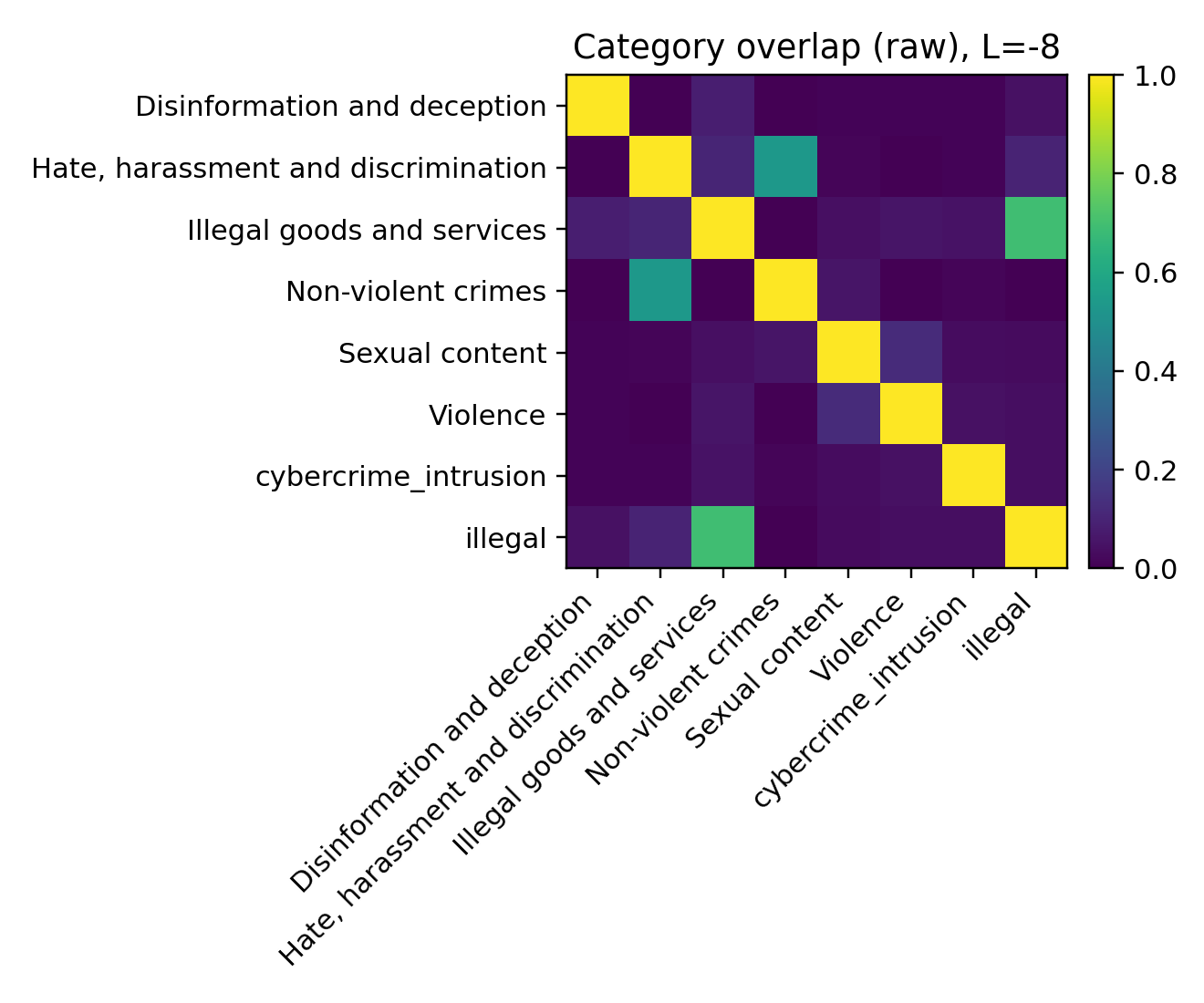}
\caption{Raw overlap matrix (Gemma-2-9B-IT, layer -8)}
\label{fig:jailbreak_overlap_raw}
\end{subfigure}
\hfill
\begin{subfigure}[b]{0.38\textwidth}
\centering
\includegraphics[width=\textwidth]{jailbreak_runs/gemma2_overlap/overlap_resid_layer-8.png}
\caption{Residualized overlap matrix (Gemma-2-9B-IT, layer -8)}
\label{fig:jailbreak_overlap_resid}
\end{subfigure}
\caption{\textbf{Subcategory subspace overlap matrices.} Residualized overlap on jailbreak dataset shows similar structure   to synthetic data, with high overlap between related categories (e.g., Illegal goods and illegal) and lower overlap between distinct categories.}
\label{fig:jailbreak_overlap_matrices}
\end{figure*}

\begin{figure*}[h]
\centering
\begin{subfigure}[b]{0.48\textwidth}
\centering
\includegraphics[width=\textwidth]{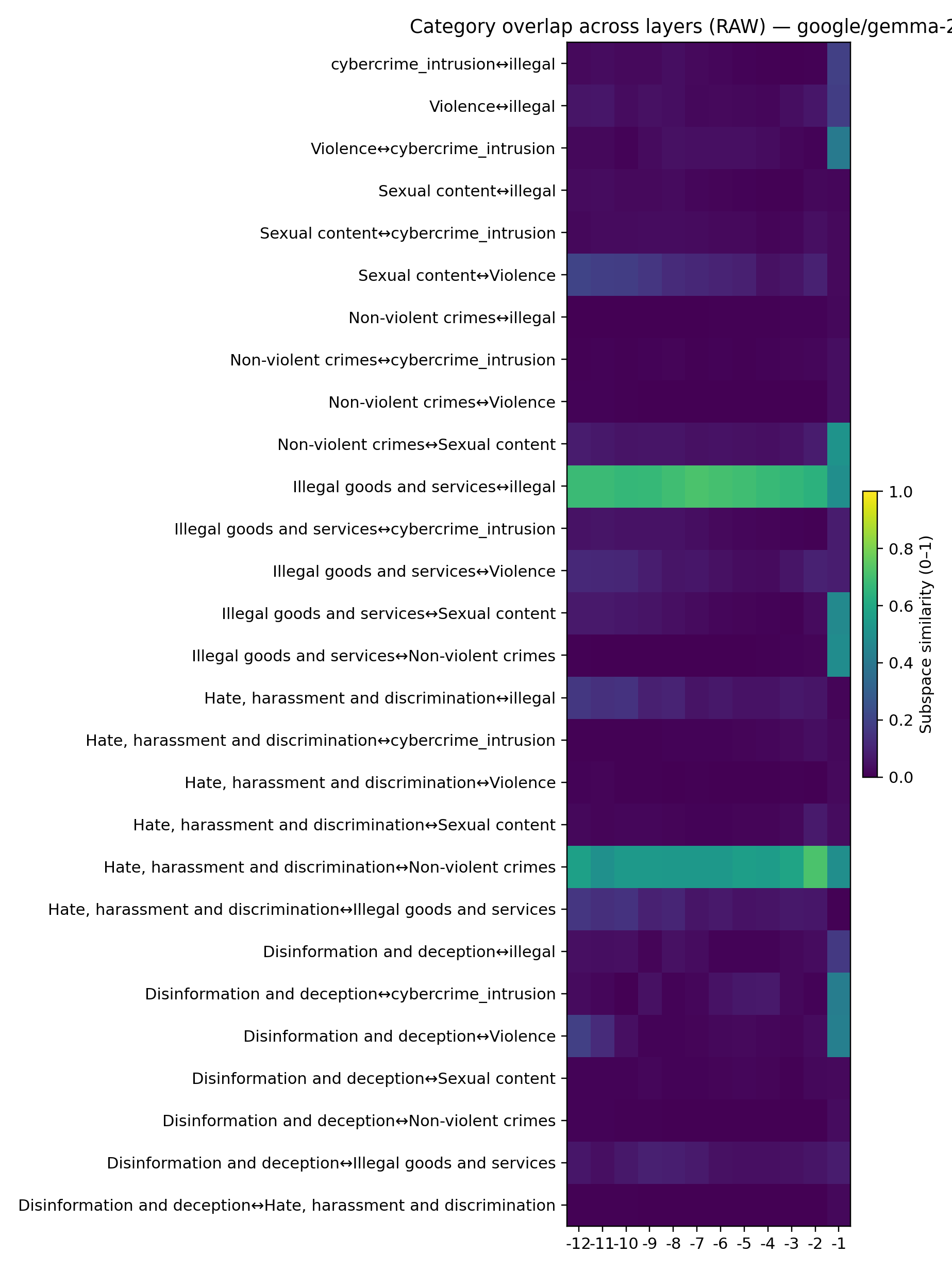}
\caption{Raw overlap across layers}
\label{fig:jailbreak_layer_map_raw}
\end{subfigure}
\hfill
\begin{subfigure}[b]{0.48\textwidth}
\centering
\includegraphics[width=\textwidth]{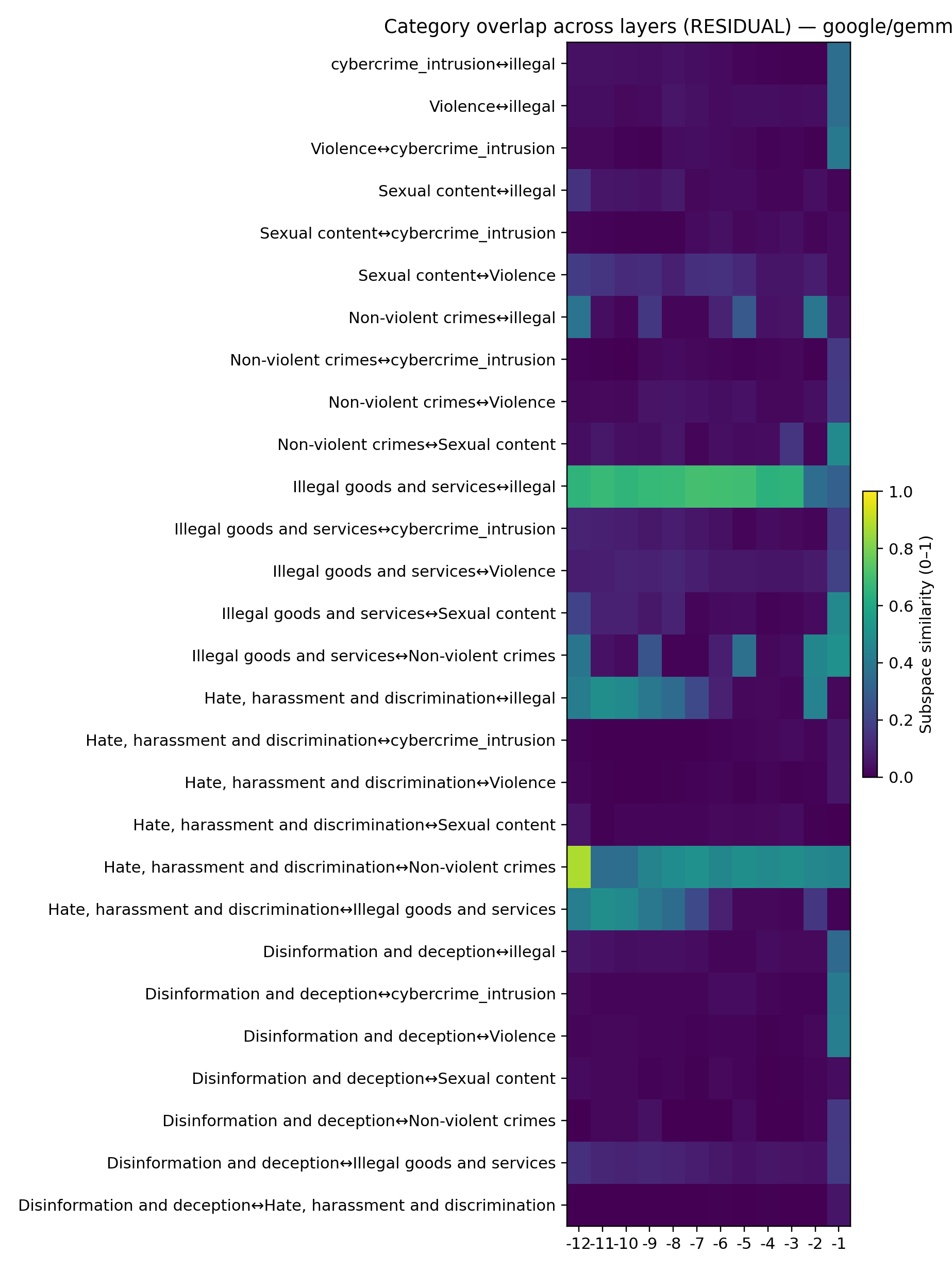}
\caption{Residualized overlap across layers}
\label{fig:jailbreak_layer_map_resid}
\end{subfigure}
\caption{\textbf{Layer-wise residualized overlap on jailbreak dataset Gemma-2-9B-IT} exhibits 
  similar localization patterns to synthetic data, with peak interaction in 
  mid-layers (-12 to -8).}
\label{fig:jailbreak_layer_maps}
\end{figure*}

\paragraph{Residualized overlap:}
Table~\ref{tab:jailbreak_overlap_summary_existing_data} and Figures \ref{fig:jailbreak_cross_model},\ref{fig:jailbreak_overlap_matrices} and \ref{fig:jailbreak_layer_maps} reports residualized geometric overlap between refusal category subspaces at layer $-8$ across three model families on the Jailbreak dataset. After removing the decision-aligned variance, many category pairs  exhibit non-trivial overlap, though the magnitude varies substantially by model and category pair. In particular, some category pairs (e.g., Hate/Harassment with Illegal goods or Non-violent content) show moderate overlap across multiple models, while others remain weakly overlapping. These results indicate that shared category-specific structure persists in a real-world jailbreak dataset, consistent with the patterns observed on the synthetic refusal data.

\paragraph{Cross-category intervention effects across models:}
Tables~\ref{tab:jailbreak_cross_effects_gemma2}–\ref{tab:jailbreak_cross_effects_mistral7b} report cross-category intervention effects at layer $-8$ across three model families. Across models, projecting out a category’s subspace reliably alters its own refusal behavior (diagonal entries), suggesting that the extracted subspaces are behaviorally relevant on the Jailbreak dataset. Interventions also induce substantial cross-category effects, though their magnitude and sign vary by architecture, and we observe similar asymmetric effects as observed in Section \ref{sec:results}. Gemma-2 and Llama-3 exhibit broad transfer across categories, with many source categories affecting multiple targets, while Mistral-7B shows generally weaker and more heterogeneous effects.

\paragraph{Decision alignment and cross-category effects:}
To relate cross-category intervention effects to upstream decision coupling, Tables~\ref{tab:jailbreak_angles_gemma2_existing}–\ref{tab:jailbreak_angles_mistral7b_existing} report principal angles between category subspaces and the decision-aligned subspace at layer $-8$. Across models, categories with smaller principal angles (stronger decision alignment) tend to produce larger and more global intervention effects, while categories that are more orthogonal to the decision subspace exhibit weaker or more localized effects.


Together, these results mirror the synthetic setting and support the main paper’s findings. One point to note is that, as stated in Section \ref{limitations}, while geometric overlap and alignment with the decision-aligned subspace help identify
structural limits of linear interventions, we expect additional nonlinear mechanisms and dataset confounds (from non-synthetic datasets)
to also contribute; our analysis is therefore not exhaustive but provides a 
starting point.

\begin{table*}[h]
\centering
\small
\caption{\textbf{Residualized overlap between category pairs across models (layer -8) for Jailbreak Dataset \cite{arditi2024refusallanguagemodelsmediated}.} Values are mean squared cosine between category subspaces after removing the decision component. Higher values (closer to 1.0) indicate more similar category-specific mechanisms.}
\label{tab:jailbreak_overlap_summary_existing_data}
\begin{tabular}{lccc}
\toprule
\textbf{Category Pair} & \textbf{Gemma-2-9B-IT} & \textbf{Llama-3-8B-Instruct} & \textbf{Mistral-7B-Instruct-v0.3}  \\
\midrule
Disinfo.\&Decep. $\leftrightarrow$ Hate/Harass. & 0.00 & 0.44 & 0.01 \\
Disinfo.\&Decep. $\leftrightarrow$ Illegal goods & 0.10 & 0.41 & 0.03   \\
Disinfo.\&Decep. $\leftrightarrow$ Non-violent & 0.00 & 0.43 & 0.01 \\
Disinfo.\&Decep. $\leftrightarrow$ Sexual & 0.02 & 0.40 & 0.01  \\
Disinfo.\&Decep. $\leftrightarrow$ Violence & 0.01 & 0.39 & 0.01  \\
Disinfo.\&Decep. $\leftrightarrow$ Cybercrime & 0.02 & 0.39 & 0.08   \\
Disinfo.\&Decep. $\leftrightarrow$ Illegal & 0.04 & 0.39 & 0.01   \\
\midrule
Hate/Harass. $\leftrightarrow$ Illegal goods & 0.35 & 0.59 & 0.06   \\
Hate/Harass. $\leftrightarrow$ Non-violent & 0.49 & 0.37 & 0.52  \\
Hate/Harass. $\leftrightarrow$ Sexual & 0.02 & 0.25 & 0.13   \\
Hate/Harass. $\leftrightarrow$ Violence & 0.01 & 0.57 & 0.05  \\
Hate/Harass. $\leftrightarrow$ Cybercrime & 0.00 & 0.23 & 0.04  \\
Hate/Harass. $\leftrightarrow$ Illegal & 0.35 & 0.27 & 0.23  \\
\midrule
Illegal goods $\leftrightarrow$ Non-violent & 0.01 & 0.31 & 0.26   \\
Illegal goods $\leftrightarrow$ Sexual & 0.10 & 0.35 & 0.06   \\
Illegal goods $\leftrightarrow$ Violence & 0.11 & 0.53 & 0.17   \\
Illegal goods $\leftrightarrow$ Cybercrime & 0.08 & 0.28 & 0.02   \\
Illegal goods $\leftrightarrow$ Illegal & 0.68 & 0.42 & 0.32   \\
\midrule
Non-violent $\leftrightarrow$ Sexual & 0.06 & 0.32 & 0.23  \\
Non-violent $\leftrightarrow$ Violence & 0.06 & 0.56 & 0.06  \\
Non-violent $\leftrightarrow$ Cybercrime & 0.03 & 0.42 & 0.00  \\
Non-violent $\leftrightarrow$ Illegal & 0.01 & 0.38 & 0.41   \\
\midrule
Sexual $\leftrightarrow$ Violence & 0.09 & 0.40 & 0.01   \\
Sexual $\leftrightarrow$ Cybercrime & 0.01 & 0.24 & 0.00   \\
Sexual $\leftrightarrow$ Illegal & 0.07 & 0.33 & 0.10  \\
\midrule
Violence $\leftrightarrow$ Cybercrime & 0.03 & 0.35 & 0.03  \\
Violence $\leftrightarrow$ Illegal & 0.06 & 0.35 & 0.08   \\
\midrule
Cybercrime $\leftrightarrow$ Illegal & 0.05 & 0.49 & 0.20  \\
\bottomrule
\end{tabular}
\end{table*}

\begin{table*}[h]
\centering
\small
\caption{\textbf{Cross-effect matrix (Gemma-2-9B-IT, layer -8, $\alpha=1.0$) for Jailbreak Dataset \cite{arditi2024refusallanguagemodelsmediated}.} Values are $\Delta$ refusal margin (log-probability units) when projecting out each source category's subspace. Positive values indicate increased refusal, negative values indicate decreased refusal.}
\label{tab:jailbreak_cross_effects_gemma2}
\begin{tabular}{l|cccccccc}
\toprule
\textbf{Source $\rightarrow$ Target} & \textbf{Disinfo.} & \textbf{Hate/Harass.} & \textbf{Illegal goods} & \textbf{Non-violent} & \textbf{Sexual} & \textbf{Violence} & \textbf{Cybercrime} & \textbf{Illegal} \\
\midrule
Disinfo.\&Decep. & +0.56 & +0.59 & +0.77 & +0.64 & +0.58 & +0.49 & +0.61 & +0.50 \\
Hate/Harass. & +0.77 & +0.69 & +0.76 & +0.77 & +0.54 & +0.60 & +0.85 & +0.60 \\
Illegal goods & -0.26 & +0.03 & +0.07 & -0.17 & -0.22 & -0.19 & -0.29 & -0.14 \\
Non-violent & +0.69 & +0.55 & +0.59 & +0.74 & +0.33 & +0.59 & +0.98 & +0.55 \\
Sexual content & -3.74 & -3.87 & -3.74 & -3.78 & -4.23 & -3.89 & -3.76 & -4.01 \\
Violence & -2.53 & -2.63 & -2.68 & -2.70 & -2.65 & -2.49 & -2.92 & -2.74 \\
Cybercrime & -1.49 & -1.26 & -1.60 & -1.33 & -1.56 & -1.24 & -1.33 & -1.51 \\
Illegal & +0.52 & +0.88 & +0.94 & +0.71 & +0.52 & +0.70 & +0.58 & +0.69 \\
\bottomrule
\end{tabular}
\end{table*}

\begin{table*}[h]
\centering
\small
\caption{\textbf{Cross-effect matrix (Llama-3-8B-Instruct, layer -8, $\alpha=1.0$) for Jailbreak Dataset \cite{arditi2024refusallanguagemodelsmediated}.} Values are $\Delta$ refusal margin (log-probability units) when projecting out each source category's subspace. Positive values indicate increased refusal, negative values indicate decreased refusal.}
\label{tab:jailbreak_cross_effects_llama3}
\begin{tabular}{l|cccccccc}
\toprule
\textbf{Source $\rightarrow$ Target} & \textbf{Disinfo.} & \textbf{Hate/Harass.} & \textbf{Illegal goods} & \textbf{Non-violent} & \textbf{Sexual} & \textbf{Violence} & \textbf{Cybercrime} & \textbf{Illegal} \\
\midrule
Disinfo.\&Decep. & -1.26 & -1.41 & -1.53 & -1.78 & -1.81 & -1.27 & -1.38 & -2.07 \\
Hate/Harass. & -2.59 & -2.74 & -2.93 & -3.13 & -3.17 & -2.47 & -2.81 & -3.48 \\
Illegal goods & -0.90 & -1.07 & -1.36 & -1.50 & -1.28 & -0.93 & -1.18 & -1.72 \\
Non-violent & -1.87 & -2.13 & -1.88 & -2.37 & -2.85 & -2.10 & -1.76 & -2.72 \\
Sexual content & -1.32 & -1.56 & -1.46 & -1.73 & -2.28 & -1.38 & -1.22 & -2.06 \\
Violence & -2.66 & -2.78 & -2.88 & -3.13 & -3.21 & -2.57 & -2.74 & -3.54 \\
Cybercrime & -1.62 & -1.77 & -1.72 & -1.98 & -1.96 & -1.66 & -1.59 & -2.15 \\
Illegal & -2.32 & -2.52 & -2.56 & -2.89 & -2.91 & -2.35 & -2.45 & -3.19 \\
\bottomrule
\end{tabular}
\end{table*}

\begin{table*}[h]
\centering
\small
\caption{\textbf{Cross-effect matrix (Mistral-7B-Instruct-v0.3, layer -8, $\alpha=1.0$) for Jailbreak Dataset \cite{arditi2024refusallanguagemodelsmediated}.} Values are $\Delta$ refusal margin (log-probability units) when projecting out each source category's subspace. Positive values indicate increased refusal, negative values indicate decreased refusal. }
\label{tab:jailbreak_cross_effects_mistral7b}
\begin{tabular}{l|cccccccc}
\toprule
\textbf{Source $\rightarrow$ Target} & \textbf{Disinfo.} & \textbf{Hate/Harass.} & \textbf{Illegal goods} & \textbf{Non-violent} & \textbf{Sexual} & \textbf{Violence} & \textbf{Cybercrime} & \textbf{Illegal} \\
\midrule
Disinfo.\&Decep. & -0.68 & -0.73 & -0.92 & -0.83 & -0.84 & -0.86 & -1.00 & -0.89 \\
Hate/Harass. & -1.59 & -1.65 & -2.01 & -1.44 & -1.38 & -1.35 & -1.31 & -1.59 \\
Illegal goods & -0.23 & -0.33 & -0.49 & -0.41 & -0.50 & -0.44 & -0.47 & -0.45 \\
Non-violent & -0.45 & -0.58 & -1.02 & -0.55 & -0.43 & -0.40 & -0.42 & -0.71 \\
Sexual content & -0.48 & -0.69 & -0.83 & -0.66 & -0.70 & -0.96 & -0.63 & -0.63 \\
Violence & +0.14 & +0.06 & +0.04 & +0.10 & +0.04 & +0.06 & +0.18 & +0.10 \\
Cybercrime & -1.74 & -1.91 & -2.07 & -1.81 & -1.78 & -1.86 & -2.19 & -2.15 \\
Illegal & -0.16 & -0.17 & -0.26 & -0.28 & -0.34 & -0.29 & -0.17 & -0.22 \\
\bottomrule
\end{tabular}
\end{table*}

\begin{table}[h]
\centering
\small
\caption{\textbf{Principal angles between decision subspace and category subspaces (degrees, Gemma-2-9B-IT, layer -8) for Jailbreak dataset}.  Smaller angles indicate greater alignment with the decision-aligned subspace .}
\label{tab:jailbreak_angles_gemma2_existing}
\begin{tabular}{lcc}
\toprule
\textbf{Category} & \textbf{Angle 1 (°)} & \textbf{Angle 2 (°)} \\
\midrule
Disinformation and deception & 81.0 & 87.2 \\
Hate, harassment and discrimination & 53.7 & 84.0 \\
Illegal goods and services & 80.6 & 89.5 \\
Non-violent crimes & 10.1 & 87.7 \\
Sexual content & 44.2 & 77.5 \\
Violence & 68.2 & 88.7 \\
Cybercrime & 70.5 & 83.5 \\
Illegal & 81.0 & 89.6 \\
\bottomrule
\end{tabular}
\end{table}

\begin{table}[h]
\centering
\small
\caption{\textbf{Principal angles between decision subspace and category subspaces (degrees, Llama-3-8B-Instruct, layer -8) for Jailbreak dataset.}  Smaller angles indicate greater alignment with the decision-aligned subspace.}
\label{tab:jailbreak_angles_llama3_existing}
\begin{tabular}{lcc}
\toprule
\textbf{Category} & \textbf{Angle 1 (°)} & \textbf{Angle 2 (°)} \\
\midrule
Disinformation and deception & 12.7 & 89.5 \\
Hate, harassment and discrimination & 21.1 & 86.8 \\
Illegal goods and services & 8.9 & 85.1 \\
Non-violent crimes & 16.0 & 81.4 \\
Sexual content & 21.0 & 82.7 \\
Violence & 16.6 & 81.1 \\
Cybercrime & 26.6 & 89.2 \\
Illegal & 10.2 & 87.3 \\
\bottomrule
\end{tabular}
\end{table}

\begin{table}[h]
\centering
\small
\caption{\textbf{Principal angles between decision subspace and category subspaces (degrees, Mistral-7B-Instruct-v0.3, layer -8) for Jailbreak dataset.}  Smaller angles indicate greater alignment with the decision-aligned subspace.}
\label{tab:jailbreak_angles_mistral7b_existing}
\begin{tabular}{lcc}
\toprule
\textbf{Category} & \textbf{Angle 1 (°)} & \textbf{Angle 2 (°)} \\
\midrule
Disinformation and deception & 55.9 & 85.5 \\
Hate, harassment and discrimination & 79.2 & 89.9 \\
Illegal goods and services & 33.6 & 82.1 \\
Non-violent crimes & 70.3 & 88.9 \\
Sexual content & 70.3 & 86.2 \\
Violence & 32.6 & 74.8 \\
Cybercrime & 69.2 & 88.1 \\
Illegal & 67.0 & 78.9 \\
\bottomrule
\end{tabular}
\end{table}

\section{Experiment on Synthetic Sycophancy Dataset}
\label{app:sycophancy}

To assess whether the proposed framework generalizes beyond refusal, we apply the same analysis pipeline to \emph{sycophancy}, defined as the tendency of a model to agree with user statements, including false claims. We generate our synthetic dataset using the same methodology as described in Appendix \ref{sythetic_data}.     We analyze five content-based categories—\textsc{GEOGRAPHY}, \textsc{HISTORY}, \textsc{MATH}, \textsc{SCIENCE}, and \textsc{SUBJECTIVE} (opinion-based statements)—using the same three model families and layer window as in the refusal experiments.

\begin{figure*}[h]
\centering
\begin{subfigure}[b]{0.24\textwidth}
\centering
\includegraphics[width=\textwidth]{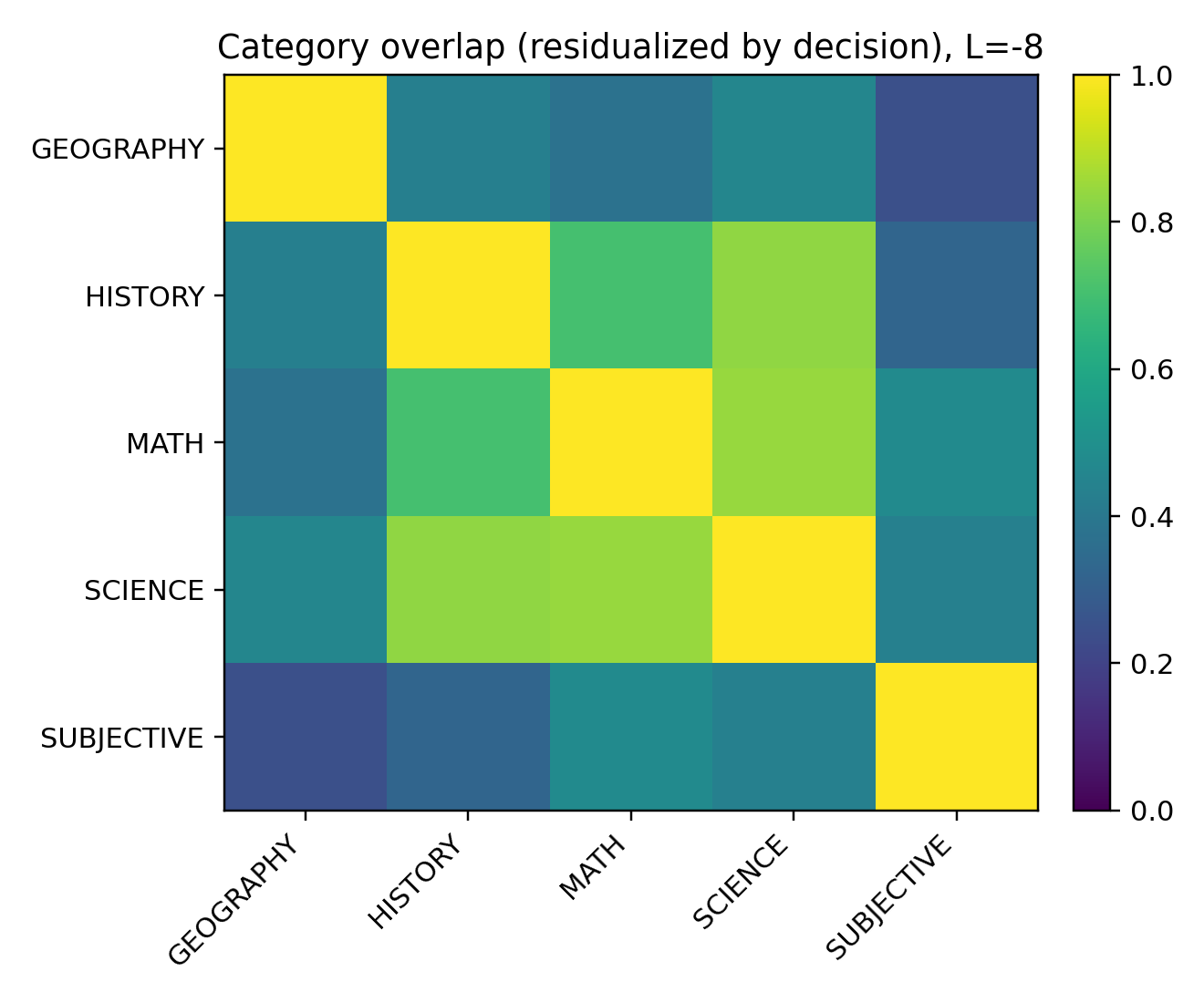}
\caption{Gemma-2-9B-IT}
\end{subfigure}
\hfill
\begin{subfigure}[b]{0.24\textwidth}
\centering
\includegraphics[width=\textwidth]{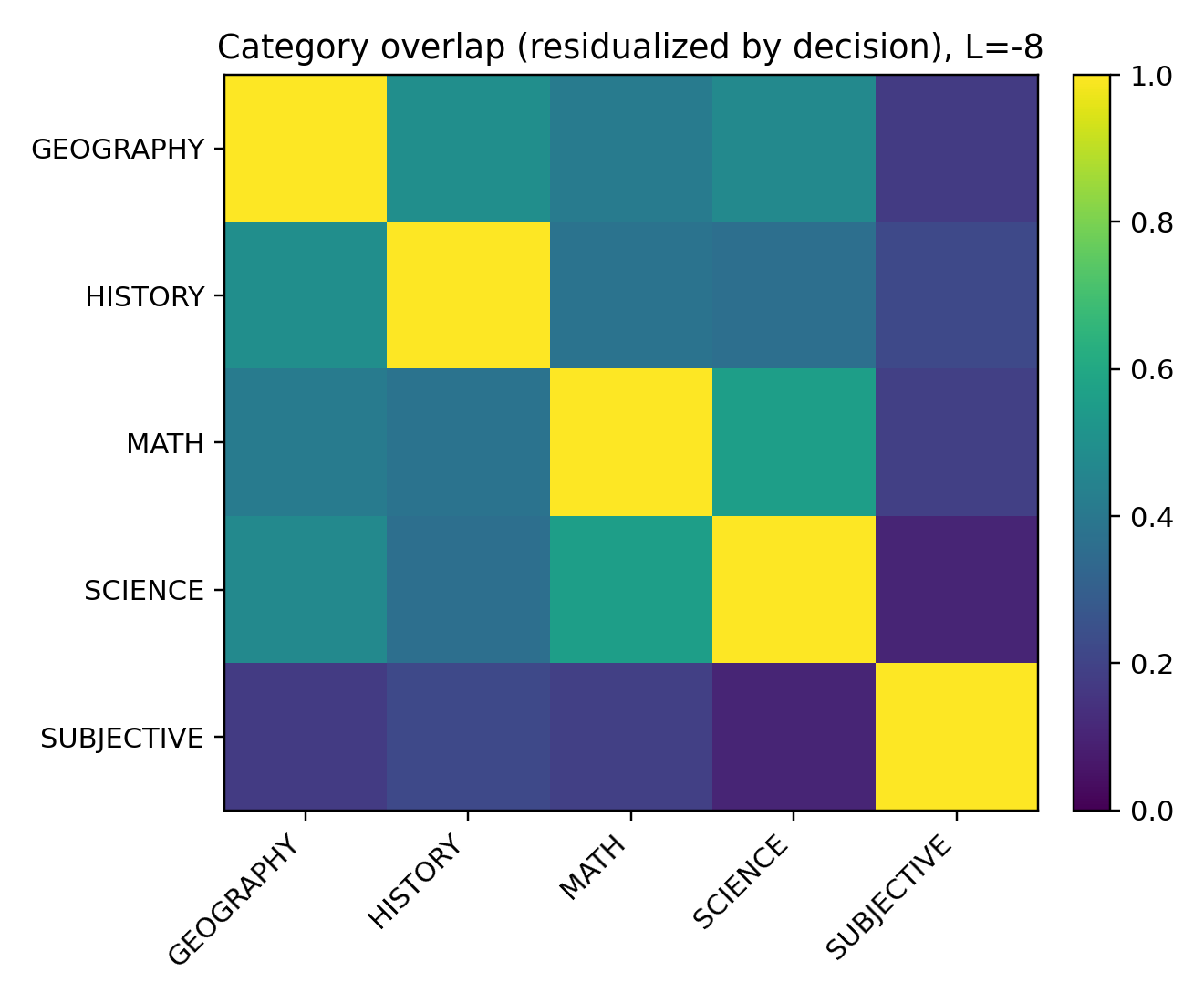}
\caption{Llama-3-8B-Instruct}
\end{subfigure}
\hfill
\begin{subfigure}[b]{0.24\textwidth}
\centering
\includegraphics[width=\textwidth]{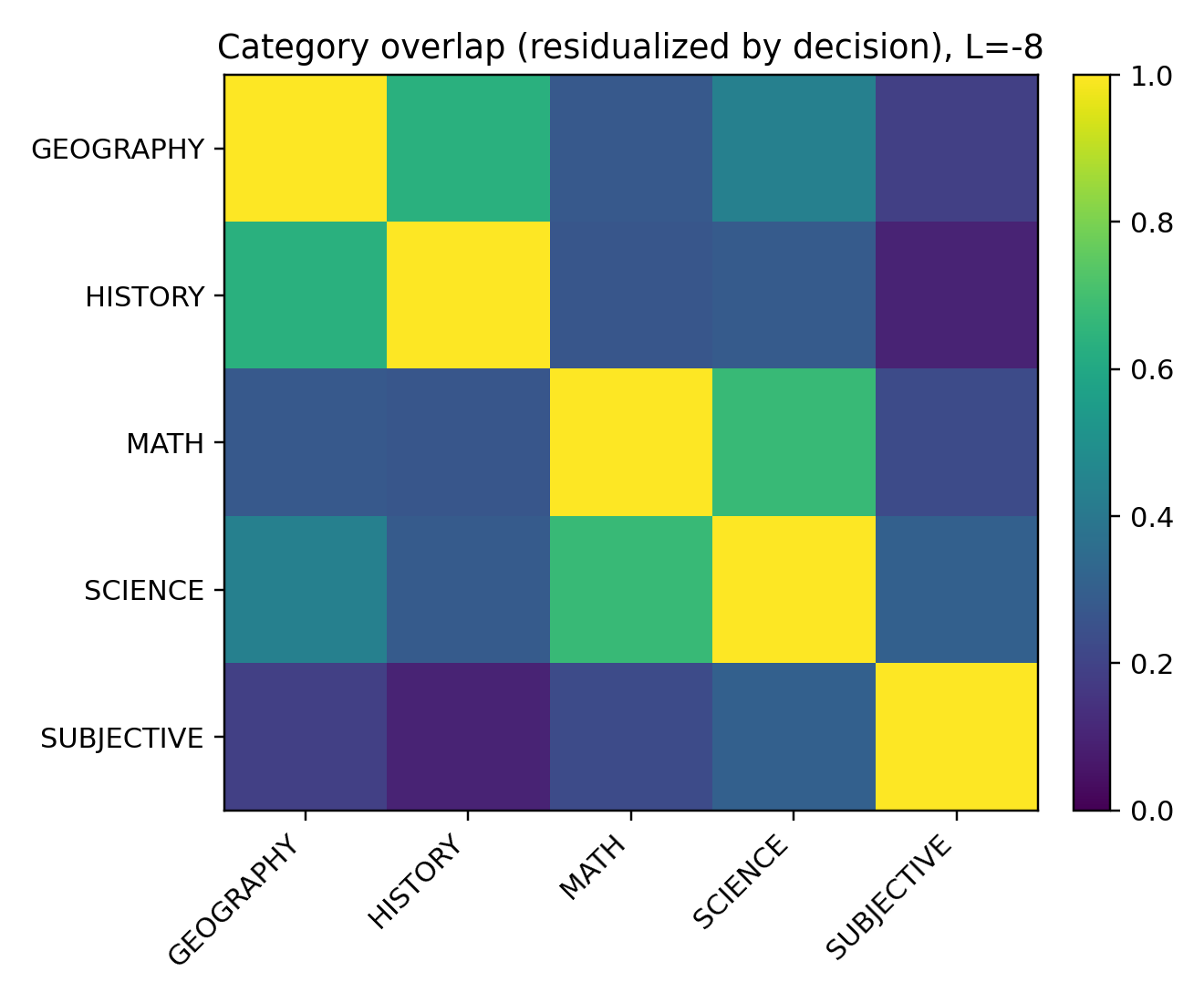}
\caption{Mistral-7B-Instruct-v0.3}
\end{subfigure}
\hfill
\caption{\textbf{Robustness check on Sycophancy Dataset:} Cross-model comparison of residualized overlap (layer $-8$). Residualized overlap patterns on synthetic sycophancy dataset  mirror those observed on refusal. }
\label{fig:sycophancy_cross_model}
\end{figure*}

\paragraph{Residualized overlap:}
Table~\ref{tab:sycophancy_overlap_summary} and Figure \ref{fig:sycophancy_cross_model} reports residualized geometric overlap between sycophancy subcategory subspaces at layer $-8$ across three model families. After removing the decision-aligned variance, many category pairs retain moderate to high overlap, particularly between closely related factual domains such as \textsc{Math}–\textsc{Science} and \textsc{History}–\textsc{Science}. In contrast, \textsc{Subjective} questions exhibit lower overlap with factual categories across models. These results indicate that subcategory-specific structure in sycophancy is distributed across overlapping low-rank subspaces.

\paragraph{Cross-category intervention effects:}
Tables~\ref{tab:cross_effects_gemma2_sycophancy}–\ref{tab:cross_effects_mistral7b_sycophancy} report cross-category intervention effects at layer $-8$ across three model families. Projecting out a category’s subspace reliably alters its own sycophancy behavior (diagonal entries) and induces substantial effects on other categories.  While effect magnitudes vary by architecture—with Mistral-7B  exhibiting larger overall shifts—the qualitative pattern of cross-category transfer is consistent across models.

\paragraph{Decision alignment and effect magnitude:}
To relate cross-category effects to upstream decision coupling, Tables~\ref{tab:angles_gemma2_sycophancy}–\ref{tab:angles_mistral7b_sycophancy} report principal angles between sycophancy subcategory subspaces and the decision-aligned subspace at layer $-8$. Categories with smaller principal angles tend to induce larger and more global intervention effects, while more orthogonal categories exhibit weaker or more localized effects. For example, in Llama-3, most categories are tightly aligned with the decision-aligned subspace and correspondingly produce broadly uniform effects across targets, whereas Gemma-2 and Mistral-7B show more heterogeneous alignment profiles and effect magnitudes. This mirrors the refusal and jailbreak experiments and supports that geometric overlap identifies potential interaction pathways, while decision alignment affects the magnitude and asymmetry of intervention effects.

Together, these results suggest that the representational geometry underlying sycophancy exhibits the same low-rank overlap, limited selectivity, and decision-alignment-driven asymmetry observed for refusal, indicating that these phenomena generalize across behaviors.

\begin{table*}[h]
\centering
\small
\caption{\textbf{Residualized overlap between category pairs across models (layer -8) for Sycophancy experiment.} Values are mean squared cosine between category subspaces after removing the decision component. Higher values (closer to 1.0) indicate more similar category-specific mechanisms.}
\label{tab:sycophancy_overlap_summary}
\begin{tabular}{lccc}
\toprule
\textbf{Category Pair} & \textbf{Gemma-2-9B-IT} & \textbf{Llama-3-8B-Instruct} & \textbf{Mistral-7B-Instruct-v0.3}  \\
\midrule
GEOGRAPHY $\leftrightarrow$ HISTORY & 0.43 & 0.50 & 0.63  \\
GEOGRAPHY $\leftrightarrow$ MATH & 0.38 & 0.42 & 0.28   \\
GEOGRAPHY $\leftrightarrow$ SCIENCE & 0.46 & 0.47 & 0.44   \\
GEOGRAPHY $\leftrightarrow$ SUBJECTIVE & 0.24 & 0.17 & 0.19   \\
\midrule
HISTORY $\leftrightarrow$ MATH & 0.70 & 0.38 & 0.27  \\
HISTORY $\leftrightarrow$ SCIENCE & 0.83 & 0.37 & 0.29   \\
HISTORY $\leftrightarrow$ SUBJECTIVE & 0.33 & 0.22 & 0.10   \\
\midrule
MATH $\leftrightarrow$ SCIENCE & 0.84 & 0.56 & 0.67   \\
MATH $\leftrightarrow$ SUBJECTIVE & 0.48 & 0.19 & 0.23  \\
\midrule
SCIENCE $\leftrightarrow$ SUBJECTIVE & 0.44 & 0.10 & 0.31  \\
\bottomrule
\end{tabular}
\end{table*}

\begin{table*}[h]
\centering
\small
\caption{\textbf{Cross-effect matrix (Gemma-2-9B-IT, layer -8, $\alpha=1.0$).} Sycophancy experiment. Values are $\Delta$ decision margin (log-probability units) when projecting out each source category's subspace. Positive values indicate increased sycophancy, negative values indicate decreased sycophancy.}
\label{tab:cross_effects_gemma2_sycophancy}
\begin{tabular}{l|ccccc}
\toprule
\textbf{Source $\rightarrow$ Target} & \textbf{GEOGRAPHY} & \textbf{HISTORY} & \textbf{MATH} & \textbf{SCIENCE} & \textbf{SUBJECTIVE} \\
\midrule
GEOGRAPHY & +1.29 & +1.01 & +1.41 & +1.24 & +2.31 \\
HISTORY & -2.57 & -3.06 & -2.48 & -2.40 & -1.69 \\
MATH & -3.73 & -3.84 & -3.72 & -3.15 & -2.27 \\
SCIENCE & +0.14 & -0.20 & +0.25 & +0.40 & +1.72 \\
SUBJECTIVE & -7.55 & -7.82 & -7.41 & -7.13 & -8.28 \\
\bottomrule
\end{tabular}
\end{table*}

\begin{table*}[h]
\centering
\small
\caption{\textbf{Cross-effect matrix (Llama-3-8B-Instruct, layer -8, $\alpha=1.0$).} Sycophancy experiment. Values are $\Delta$ decision margin (log-probability units) when projecting out each source category's subspace. Positive values indicate increased sycophancy, negative values indicate decreased sycophancy.}
\label{tab:cross_effects_llama3_sycophancy}
\begin{tabular}{l|ccccc}
\toprule
\textbf{Source $\rightarrow$ Target} & \textbf{GEOGRAPHY} & \textbf{HISTORY} & \textbf{MATH} & \textbf{SCIENCE} & \textbf{SUBJECTIVE} \\
\midrule
GEOGRAPHY & -3.65 & -4.43 & -3.29 & -3.07 & -2.40 \\
HISTORY & -3.89 & -4.70 & -3.58 & -3.37 & -2.73 \\
MATH & -6.83 & -7.28 & -6.34 & -6.21 & -5.66 \\
SCIENCE & -6.48 & -6.83 & -5.93 & -5.80 & -5.27 \\
SUBJECTIVE & -1.53 & -1.88 & -1.36 & -1.30 & -1.10 \\
\bottomrule
\end{tabular}
\end{table*}

\begin{table*}[h]
\centering
\small
\caption{\textbf{Cross-effect matrix (Mistral-7B-Instruct-v0.3, layer -8, $\alpha=1.0$).} Sycophancy experiment. Values are $\Delta$ decision margin (log-probability units) when projecting out each source category's subspace. Positive values indicate increased sycophancy, negative values indicate decreased sycophancy.}
\label{tab:cross_effects_mistral7b_sycophancy}
\begin{tabular}{l|ccccc}
\toprule
\textbf{Source $\rightarrow$ Target} & \textbf{GEOGRAPHY} & \textbf{HISTORY} & \textbf{MATH} & \textbf{SCIENCE} & \textbf{SUBJECTIVE} \\
\midrule
GEOGRAPHY & -5.81 & -7.03 & -4.86 & -4.83 & -3.09 \\
HISTORY & -3.02 & -2.96 & -2.42 & -2.80 & -1.61 \\
MATH & -9.66 & -9.09 & -7.01 & -8.08 & -5.53 \\
SCIENCE & -7.09 & -6.56 & -5.47 & -6.19 & -3.99 \\
SUBJECTIVE & -9.80 & -9.75 & -6.45 & -7.69 & -6.65 \\
\bottomrule
\end{tabular}
\end{table*}

\begin{table}[h]
\centering
\small
\caption{\textbf{Principal angles between decision subspace and category subspaces (degrees, Gemma-2-9B-IT, layer -8).}  Smaller angles indicate greater alignment with the decision-aligned subspace.}
\label{tab:angles_gemma2_sycophancy}
\begin{tabular}{lcc}
\toprule
\textbf{Category} & \textbf{Angle 1 (°)} & \textbf{Angle 2 (°)} \\
\midrule
GEOGRAPHY & 46.9 & 88.9 \\
HISTORY & 73.8 & 88.7 \\
MATH & 78.6 & 89.0 \\
SCIENCE & 72.1 & 88.3 \\
SUBJECTIVE & 44.2 & 86.4 \\
\bottomrule
\end{tabular}
\end{table}

\begin{table}[h]
\centering
\small
\caption{\textbf{Principal angles between decision subspace and category subspaces (degrees, Llama-3-8B-Instruct, layer -8).} Sycophancy experiment. Smaller angles indicate greater alignment with the decision-aligned subspace.}
\label{tab:angles_llama3_sycophancy}
\begin{tabular}{lcc}
\toprule
\textbf{Category} & \textbf{Angle 1 (°)} & \textbf{Angle 2 (°)} \\
\midrule
GEOGRAPHY & 16.8 & 18.7 \\
HISTORY & 16.8 & 23.8 \\
MATH & 11.2 & 17.7 \\
SCIENCE & 7.7 & 17.6 \\
SUBJECTIVE & 31.6 & 72.1 \\
\bottomrule
\end{tabular}
\end{table}

\begin{table}[h]
\centering
\small
\caption{\textbf{Principal angles between decision subspace and category subspaces (degrees, Mistral-7B-Instruct-v0.3, layer -8).}  Smaller angles indicate greater alignment with the decision-aligned subspace.}
\label{tab:angles_mistral7b_sycophancy}
\begin{tabular}{lcc}
\toprule
\textbf{Category} & \textbf{Angle 1 (°)} & \textbf{Angle 2 (°)} \\
\midrule
GEOGRAPHY & 79.7 & 87.9 \\
HISTORY & 74.1 & 85.1 \\
MATH & 61.1 & 82.6 \\
SCIENCE & 81.2 & 89.4 \\
SUBJECTIVE & 71.8 & 81.8 \\
\bottomrule
\end{tabular}
\end{table}

\section{Detailed Methodology Choices}
\label{app:methodology}
\paragraph{ Decision-aligned variance removal:}
We view each contrast vector as a linear sum of a decision-aligned subspace and residual structure, as an analytic decomposition for  intervention analysis:
\begin{equation}
\Delta_i^{(m,\ell)} \;=\; V_{\mathrm{dec}}\, d_i \;+\; r_i,
\end{equation}
where $V_{\mathrm{dec}}\in\mathbb{R}^{d_{\text{model}}\times k_{\mathrm{dec}}}$ has orthonormal columns and spans a
low-rank \emph{decision-aligned subspace}, $d_i\in\mathbb{R}^{k_{\mathrm{dec}}}$ are example-specific decision
coefficients, and $r_i$ captures residual (e.g., category/style) variation.
When decision-related variance dominates, PCA of $\Delta^{(m,\ell)}$ can prioritize decision-aligned
directions. We therefore identify decision-aligned PCs by ranking components by the absolute
correlation between their PC scores and the decision margin $\mu$, and remove the top
$k_{\mathrm{dec}}$ such components via orthogonal projection (eq. \ref{eq:residualize}), where $V_{\mathrm{dec}}\in\mathbb{R}^{{{d_\mathrm{model}}\times k_\mathrm{dec}}}$ contains an orthonormal column basis for the
selected decision-aligned subspace.
We report geometric overlap both before and after this residualization, and use
$\Delta_{\perp}^{(m,\ell)}$ to isolate category- or style-specific structure.

\paragraph{Low-rank residual subspaces:}
After decision-aligned variance removal, let
\(\Delta_{\perp}^{(m,\ell)}\in\mathbb{R}^{N\times d_{\mathrm{model}}}\) denote the matrix of
residual contrasts and define the empirical feature covariance $\Sigma_r \;=\; \tfrac{1}{N}\Delta_{\perp}^{(m,\ell)\top}\Delta_{\perp}^{(m,\ell)}$.
In practice, \(\Sigma_r\) is not well approximated by a rank-1 matrix: its leading
eigenvalues are not well separated, suggesting that residual (e.g., category or style)
structure is distributed across multiple directions in representation space.
Consequently, rank-1 summaries are unstable.
We therefore apply PCA (equivalently performing eigendecomposition of the residual
feature covariance $\Sigma_r$) to \(\Delta_{\perp}^{(m,\ell)}\) and retain the top-\(k\) eigenvectors,
forming a low-rank subspace \(B_m^{(\ell)}\) that captures the dominant residual structure
and provides a stable object for principal-angle comparison and subspace-level
interventions.

\paragraph{Complementary roles of overlap and decision alignment:}
We remove decision-aligned variance when measuring subspace overlap to isolate category-specific representational structure. Without this step, overlap is dominated by the shared decision mechanism, causing all categories to appear artificially similar. However, decision alignment remains critical for understanding intervention effects. While residualized overlap identifies where representations are shared, alignment with the decision subspace helps predict how strongly those shared representations influence the model’s output. Thus, residualization and decision alignment serve complementary roles: residualization isolates structure, while decision alignment explains how that structure is used.

\section{Further Results on Refusal (Synthetic Dataset)}
\label{app_refusal}

\subsection{Intervention validation: within-category effects}
\label{app:within_steering}

Projecting out a category’s own subspace  alters the model’s refusal
margin across categories and models (Table~\ref{tab:self_intervention_app}),
suggesting that the extracted subspaces  influence the refusal decision.
The direction of this influence is not uniform: for some categories, projection
reduces refusal, while for others it increases refusal.

In particular, some subspaces appear to act as decision-driving components whose
removal weakens refusal (e.g., \textsc{Against Policy}, \textsc{Deceptive/Malicious}),
whereas others act as decision-modulating components that support contextual or
policy-sensitive discrimination.
Removing such modulating subspaces can reduce the model’s ability to make nuanced
judgments and can cause it to default to a more conservative in some settings
(e.g., \textsc{Copyright} or \textsc{NSFW} in Gemma-2). 
Accordingly, the direction of the self-intervention effect reflects the functional
role of the subspace rather than its geometric alignment alone.
These effects vary across models, reflecting differences in how category-specific
representations are integrated into the decision-aligned subspace.

\begin{table}[h]
\centering
\small
\setlength{\tabcolsep}{0.6pt}
\caption{\textbf{Self-intervention effects across models and categories (layer -8, $\alpha=1.0$).} Values are $\Delta$ refusal margin (diagonal entries of cross-effect matrices) when projecting out each category's own subspace. Negative values indicate decreased refusal, positive values indicate increased refusal. Categories show   effects across models, suggesting that extracted subspaces are  linked to refusal behavior.}
\label{tab:self_intervention_app}
\begin{tabular}{lccc}
\toprule
\textbf{Category} & \textbf{Gemma-2-9B-IT} & \textbf{Llama-3-8B-Instruct} & \textbf{Mistral-7B-Instruct-v0.3}  \\
\midrule
Against Policy & -8.17 & -2.14 & -16.20  \\
Copyright & +3.84 & -1.88 & -1.48  \\
Deceptive/Malicious & -7.86 & -4.25 & -0.57  \\
Harmful & -0.47 & -4.87 & -0.15  \\
NSFW & +0.98 & -3.66 & -17.27 \\
\bottomrule
\end{tabular}
\end{table}

\subsection{Intervention strength Ablation}
\label{app:alpha_ablation}
To test whether cross-category effects are driven by overly strong interventions,
we vary the projection strength $\alpha \in [0.1, 1.0]$ and measure both the effect
on the target category (self-effect) and the largest effect on any other category
(max collateral effect).
We evaluate this for the \textsc{Deceptive/Malicious} $\rightarrow$ \textsc{NSFW}
intervention at layer $-8$ (8\textsuperscript{th} layer from output), where residualized overlap is high.

As shown in Tables~\ref{tab:alpha_effects_gemma} and~\ref{tab:alpha_effects_llama},
reducing $\alpha$ decreases the absolute magnitude of both self and collateral effects.
However, collateral effects remain comparable in scale to the self effect across all
values of $\alpha$.
In particular, weakening the intervention reduces its overall impact and does not
eliminate cross-category effects. Private-subspace projections (Appendix \ref{app:shared-private}) reduce the absolute size of collateral effects relative
to full-subspace projections, but collateral effects persist even at small $\alpha$.


\begin{table}[h]
\centering
\small
\caption{
\textbf{Self-effects and collateral effects under varying intervention strength ($\alpha$) (Gemma-2-9B-IT).}
Self-effect is $|\Delta M(\text{target})|$ and max collateral is $\max|\Delta M(\text{others})|$.
Results for \textsc{Deceptive/Malicious} $\rightarrow$ \textsc{NSFW} at layer $-8$ .
}
\label{tab:alpha_effects_gemma}
\begin{tabular}{c|cc|cc|cc}
\toprule
 & \multicolumn{2}{c|}{\textbf{Rank-1}} & \multicolumn{2}{c|}{\textbf{Private Subspace}} & \multicolumn{2}{c}{\textbf{Full Subspace}} \\
\cmidrule(lr){2-3} \cmidrule(lr){4-5} \cmidrule(lr){6-7}
\textbf{$\alpha$} & Self & Max Coll. & Self & Max Coll. & Self & Max Coll. \\
\midrule
0.10 & 0.01 & 0.09 & 0.12 & 0.13 & 0.09 & 0.20 \\
0.25 & 0.00 & 0.17 & 0.30 & 0.32 & 0.27 & 0.47 \\
0.50 & 0.04 & 0.44 & 0.57 & 0.61 & 0.76 & 1.08 \\
0.75 & 0.07 & 0.63 & 0.84 & 0.87 & 1.79 & 2.09 \\
1.00 & 0.14 & 0.90 & 1.14 & 1.19 & 3.26 & 3.33 \\
\bottomrule
\end{tabular}
\end{table}

\begin{table}[h]
\centering
\small
\caption{
\textbf{Self-effects and collateral effects under varying intervention strength ($\alpha$)(Llama-3-8B-Instruct)}
Self-effect is $|\Delta M(\text{target})|$ and max collateral is $\max|\Delta M(\text{others})|$.
Results for \textsc{Deceptive/Malicious} $\rightarrow$ \textsc{NSFW} at layer $-8$ .
}
\label{tab:alpha_effects_llama}
\begin{tabular}{c|cc|cc|cc}
\toprule
 & \multicolumn{2}{c|}{\textbf{Rank-1}} & \multicolumn{2}{c|}{\textbf{Private Subspace}} & \multicolumn{2}{c}{\textbf{Full Subspace}} \\
\cmidrule(lr){2-3} \cmidrule(lr){4-5} \cmidrule(lr){6-7}
\textbf{$\alpha$} & Self & Max Coll. & Self & Max Coll. & Self & Max Coll. \\
\midrule
0.10 & 0.03 & 0.07 & 0.01 & 0.03 & 0.17 & 0.20 \\
0.25 & 0.08 & 0.17 & 0.06 & 0.11 & 0.45 & 0.48 \\
0.50 & 0.15 & 0.32 & 0.13 & 0.20 & 1.02 & 1.10 \\
0.75 & 0.28 & 0.52 & 0.20 & 0.30 & 1.60 & 1.78 \\
1.00 & 0.36 & 0.72 & 0.29 & 0.41 & 2.20 & 2.50 \\
\bottomrule
\end{tabular}
\end{table}

\subsection{Results for other models}
\label{app:all_tables}
Tables~\ref{tab:cross_effects_llama3_refusal}–\ref{tab:cross_effects_mistral7b_refusal} report cross-category intervention effects for the remaining model families at layer $-8$ (8\textsuperscript{th} layer from output), and Tables~\ref{tab:angles_llama3_refusal}–\ref{tab:angles_mistral7b_refusal} report the corresponding principal angles between category subspaces and the decision-aligned subspace. While the magnitude and sign of effects vary across architectures, due to different representations structured from different training regimes, all models exhibit non-trivial cross-category transfer under projection and heterogeneous alignment with the decision-aligned subspace. These results mirror the qualitative patterns discussed in Section~\ref{results:h2} and support the generality of the main paper’s findings across model families.

\begin{table*}[h]
\centering
\small
\caption{\textbf{Cross-effect matrix (Llama-3-8B-Instruct, layer -8, $\alpha=1.0$).} Values are $\Delta$ refusal margin (log-probability units) when projecting out each source category's subspace. Positive values indicate increased refusal, negative values indicate decreased refusal. }
\label{tab:cross_effects_llama3_refusal}
\begin{tabular}{l|ccccc}
\toprule
\textbf{Source $\rightarrow$ Target} & \textbf{Harmful} & \textbf{Copyright} & \textbf{Deceptive} & \textbf{NSFW} & \textbf{Against Policy} \\
\midrule
\textsc{Against Policy} & $-3.40$ & $-1.33$ & $-2.77$ & $-2.72$ & $-2.14$ \\
\textsc{Copyright} & $-2.79$ & $-1.88$ & $-2.57$ & $-2.54$ & $-2.30$ \\
\textsc{Deceptive/Malicious} & $-5.05$ & $-2.39$ & $-4.25$ & $-4.14$ & $-3.43$ \\
\textsc{Harmful} & $-4.87$ & $-2.05$ & $-4.03$ & $-3.89$ & $-3.08$ \\
\textsc{NSFW} & $-5.01$ & $-1.39$ & $-3.81$ & $-3.66$ & $-2.63$ \\
\bottomrule
\end{tabular}
\end{table*}

\begin{table*}[h]
\centering
\small
\caption{\textbf{Cross-effect matrix (Mistral-7B-Instruct-v0.3, layer -8, $\alpha=1.0$).} Values are $\Delta$ refusal margin (log-probability units) when projecting out each source category's subspace. Positive values indicate increased refusal, negative values indicate decreased refusal. }
\label{tab:cross_effects_mistral7b_refusal}
\begin{tabular}{l|ccccc}
\toprule
\textbf{Source $\rightarrow$ Target} & \textbf{Harmful} & \textbf{Copyright} & \textbf{Deceptive} & \textbf{NSFW} & \textbf{Against Policy} \\
\midrule
\textsc{Against Policy} & $-15.32$ & $-14.72$ & $-15.15$ & $-15.77$ & $-16.20$ \\
\textsc{Copyright} & $-1.47$ & $-1.48$ & $-1.58$ & $-1.61$ & $-1.48$ \\
\textsc{Deceptive/Malicious} & $-0.53$ & $-0.31$ & $-0.57$ & $-0.51$ & $-0.58$ \\
\textsc{Harmful} & $-0.15$ & $-0.25$ & $-0.35$ & $-0.11$ & $-0.17$ \\
\textsc{NSFW} & $-16.74$ & $-16.20$ & $-16.64$ & $-17.27$ & $-17.65$ \\
\bottomrule
\end{tabular}
\end{table*}

\begin{table}[h]
\centering
\small
\caption{\textbf{Principal angles between decision subspace and category subspaces (degrees, Llama-3-8B-Instruct, layer -8).}  Smaller angles indicate greater alignment with the decision-aligned subspace. All categories show strong alignment along the first principal direction.}
\label{tab:angles_llama3_refusal}
\begin{tabular}{lcc}
\toprule
\textbf{Category} & \textbf{Angle 1 (°)} & \textbf{Angle 2 (°)} \\
\midrule
\textsc{Harmful} & 10.0 & 87.8 \\
\textsc{Copyright} & 9.6 & 83.4 \\
\textsc{Deceptive/Malicious} & 7.0 & 79.9 \\
\textsc{NSFW} & 5.9 & 85.0 \\
\textsc{Against Policy} & 5.9 & 88.3 \\
\bottomrule
\end{tabular}
\end{table}

\begin{table}[h]
\centering
\small
\caption{\textbf{Principal angles between decision subspace and category subspaces (degrees, Mistral-7B-Instruct-v0.3, layer -8).} Smaller angles indicate greater alignment with the decision-aligned subspace. \textsc{Deceptive/Malicious} shows the strongest alignment; \textsc{Harmful} and \textsc{Copyright} are more orthogonal.}
\label{tab:angles_mistral7b_refusal}
\begin{tabular}{lcc}
\toprule
\textbf{Category} & \textbf{Angle 1 (°)} & \textbf{Angle 2 (°)} \\
\midrule
\textsc{Harmful} & 72.8 & 88.6 \\
\textsc{Copyright} & 71.6 & 83.2 \\
\textsc{Deceptive/Malicious} & 34.9 & 87.7 \\
\textsc{NSFW} & 56.4 & 88.9 \\
\textsc{Against Policy} & 64.5 & 87.5 \\
\bottomrule
\end{tabular}
\end{table}

\subsection{Layer-wise Interventions}
\label{app:layerwise_int}

    \paragraph{Category-specific depth profiles:}
Table~\ref{tab:category_layer_self_effects} shows how projecting out each category’s residualized subspace affects its own refusal behavior across layers in Gemma-2-9B-IT. \textsc{Deceptive/Malicious} and \textsc{Against Policy} exhibit large negative self-effects that increase from mid to late layers. \textsc{Harmful} shows  weak self-effects across depth. \textsc{NSFW} has minimal effects in early layers, with larger effects emerging in later layers. \textsc{Copyright} produces positive self-effects at most layers.

\paragraph{Depth dependence of intervention effects:}
Table~\ref{tab:layerwise_effects} summarizes mean self and collateral intervention effects across models at selected layers. Across architectures, intervention effects are weakest in early layers and increase  in mid-to-late layers, indicating where overlapping category subspaces are most prominent. Notably, increases in self-effects are accompanied by comparable increases in collateral effects. This pattern indicates that while later-layer interventions are more powerful, they also induce proportionally larger cross-category effects. The consistency of this trend across model families suggests that limits on category-specific linear control are a general consequence of how behavioral information is consolidated with depth.

\paragraph{Layer localization of category overlap:}
Table~\ref{tab:layerwise_overlap} reports mean residualized geometric overlap between category subspaces at selected layers across three model families. Across models, overlap is non-uniform across depth and is generally elevated in mid-to-late layers, consistent with the  emergence of category-specific structure after removing the dominant decision-aligned variance. The precise depth at which overlap peaks varies by architecture. This localization indicates that interacting category structure arises in specific depth windows where behavioral information is integrated.


    \begin{table*}[h]
\centering
\small
\caption{\textbf{Category self-intervention effects by layer.} Diagonal entries of cross-effect matrices showing how projecting out each category's subspace affects its own behavior at different layers for Gemma-2-9B-IT. Categories exhibit distinct depth profiles: some show steadily increasing effects in mid-to-late layers, while others remain weak or change sign, suggesting heterogeneity in when category representations influence the refusal decision.}
\label{tab:category_layer_self_effects}
\begin{tabular}{lcccccccccccc}
\toprule
\textbf{Category} & \textbf{L-12} & \textbf{L-11} & \textbf{L-10} & \textbf{L-9} & \textbf{L-8} & \textbf{L-7} & \textbf{L-6} & \textbf{L-5} & \textbf{L-4} & \textbf{L-3} & \textbf{L-2} & \textbf{L-1} \\
\midrule
\textsc{Against Policy} &   -2.39 &   -3.54 &   -4.28 &   -4.96 &   -8.17 &   -7.65 &  -11.84 &  -12.35 &  -13.37 &  -11.86 &  -14.25 &    3.19 \\
\textsc{Copyright} &    0.37 &    1.52 &    2.49 &    3.25 &    3.84 &    4.60 &    4.56 &    4.14 &    3.37 &    2.98 &    1.94 &    5.32 \\
\textsc{Deceptive/Malicious} &    0.98 &   -0.36 &   -1.54 &   -3.88 &   -7.86 &  -11.20 &  -15.57 &  -17.15 &  -19.43 &  -19.04 &  -20.17 &  -12.00 \\
\textsc{Harmful} &   -0.50 &   -1.25 &   -0.22 &   -0.75 &   -0.47 &    0.74 &    2.53 &    1.30 &    1.02 &    0.79 &    0.72 &    0.57 \\
\textsc{NSFW} &    0.68 &    0.62 &    0.10 &    0.39 &    0.98 &    0.38 &   -0.29 &    4.31 &    3.32 &    3.96 &    4.86 &    5.75 \\
\bottomrule
\end{tabular}
\end{table*}

\begin{table*}[h]
\centering
\small
\caption{\textbf{Layer-wise mean intervention effects across models.}
Self effects are mean absolute $|\Delta M|$ for diagonal entries (projecting out each category's own subspace).
Collateral effects are mean absolute $|\Delta M|$ for off-diagonal entries (cross-category transfer).
Effects are strongest in mid-to-late layers ($-8$ to $-4$), indicating where overlapping category representations are prominent.
\label{tab:layerwise_effects}}
\begin{tabular}{lcccccccccccc}
\toprule
\textbf{Model} & \textbf{Self} & \textbf{Coll.} & \textbf{Self} & \textbf{Coll.} & \textbf{Self} & \textbf{Coll.} & \textbf{Self} & \textbf{Coll.} & \textbf{Self} & \textbf{Coll.} & \textbf{Self} & \textbf{Coll.} \\
\cmidrule(lr){2-13}
 & \multicolumn{2}{c}{\textbf{L$-$20}} & \multicolumn{2}{c}{\textbf{L$-$16}} & \multicolumn{2}{c}{\textbf{L$-$12}} & \multicolumn{2}{c}{\textbf{L$-$8}} & \multicolumn{2}{c}{\textbf{L$-$4}} & \multicolumn{2}{c}{\textbf{L$-$1}} \\
\midrule
Gemma-2-9B-IT           & 2.25 & 2.31 & 1.00 & 1.03 & 0.98 & 0.83 & 4.26 & 3.89 & 8.10 & 7.88 & 5.37 & 4.93 \\
Llama-3-8B-Instruct           & 1.09 & 1.07 & 2.94 & 2.84 & 4.15 & 3.95 & 3.36 & 3.07 & 4.53 & 3.98 & 4.61 & 4.22 \\
Mistral-7B-Instruct-v0.3           & 2.13 & 1.73 & 5.00 & 4.98 & 4.27 & 4.07 & 7.13 & 6.86 & 1.02 & 1.12 & 0.68 & 0.66 \\

\bottomrule
\end{tabular}
\end{table*}

\begin{table*}[h]
\centering
\small
\caption{\textbf{Layer-wise mean residualized overlap across models.}
Values are mean residualized geometric overlap (off-diagonal) between category pairs
at selected layers. Higher values indicate stronger category-specific structure.
Overlap peaks in mid-to-late layers ($-12$ to $-8$) where category representations interact.
\label{tab:layerwise_overlap}}
\begin{tabular}{lcccccc}
\toprule
\textbf{Model} & \textbf{L$-$20} & \textbf{L$-$16} & \textbf{L$-$12} & \textbf{L$-$8} & \textbf{L$-$4} & \textbf{L$-$1} \\
\midrule
Gemma-2-9B-IT           & 0.314 & 0.345 & 0.273 & 0.306 & 0.231 & 0.299 \\
Llama-3-8B-Instruct           & 0.311 & 0.451 & 0.651 & 0.538 & 0.727 & 0.767 \\
Mistral-7B-Instruct-v0.3           & 0.319 & 0.433 & 0.430 & 0.306 & 0.384 & 0.331 \\
\bottomrule
\end{tabular}
\end{table*}

    %




\subsection{Additional Analysis of Representation Geometry on 7B-9B models on synthetic refusal dataset}
\label{app:extra_results}

\subsubsection{Layer-wise emergence of interaction}
\label{app:k:layer_emergence}
\begin{figure}[t]
    \centering
    \includegraphics[width=\linewidth]{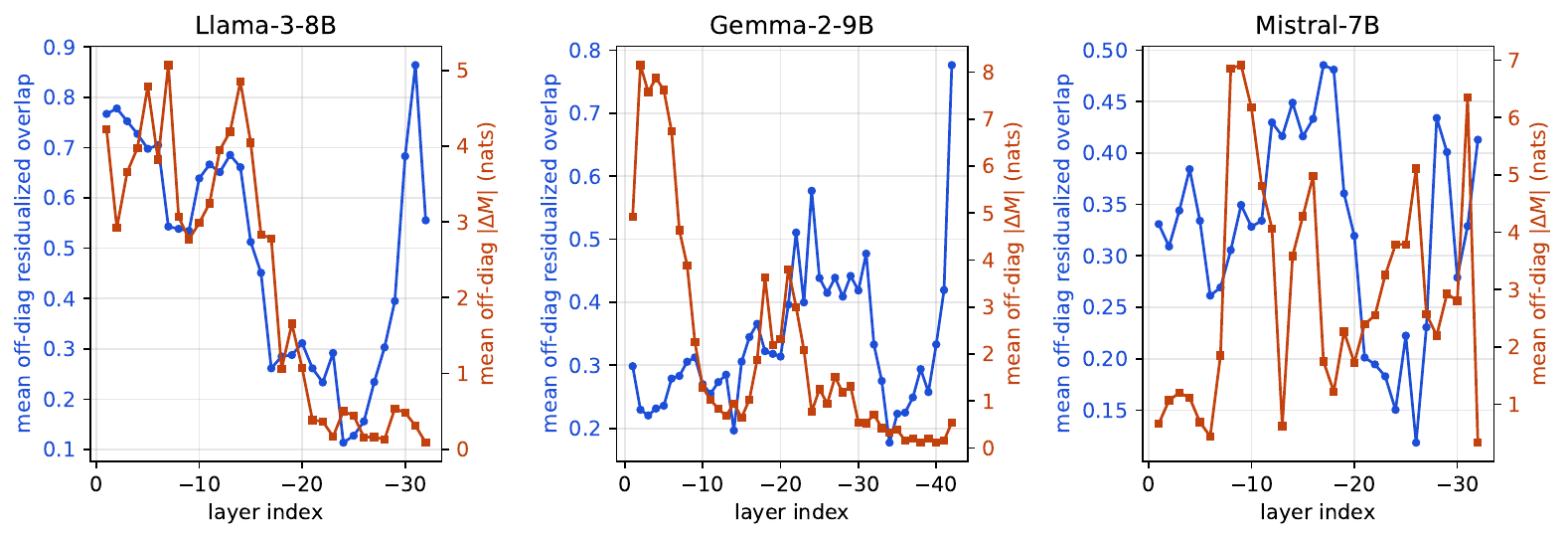}
    \caption{
    \textbf{Layer-wise evolution of overlap and cross-category effects  on synthetic refusal dataset.}
    Both overlap and intervention effects increase in middle-to-late layers,
    indicating that behavioral interactions emerge during representation
    formation rather than at the final decision stage.
    }
    \label{fig:layer_appendix}
\end{figure}

Both overlap and cross-category effects are localized to middle-to-late layers,
suggesting that behavioral interaction arises during representation composition (refer Figure \ref{fig:layer_appendix}). Across all three checkpoints, overlap and cross-effects remain comparatively small in early layers and rise  in mid-to-late layers, indicating that category entanglement and its behavioral signatures emerge during representation formation rather than appearing  at the terminal layer.
The relative peak depth differs by architecture, but the qualitative co-localization of geometric overlap and collateral effects is stable.

\subsubsection{Decision subspace is low-dimensional}
\label{app:k:decision_lowrank}
\begin{figure}[t]
    \centering
    \includegraphics[width=\linewidth]{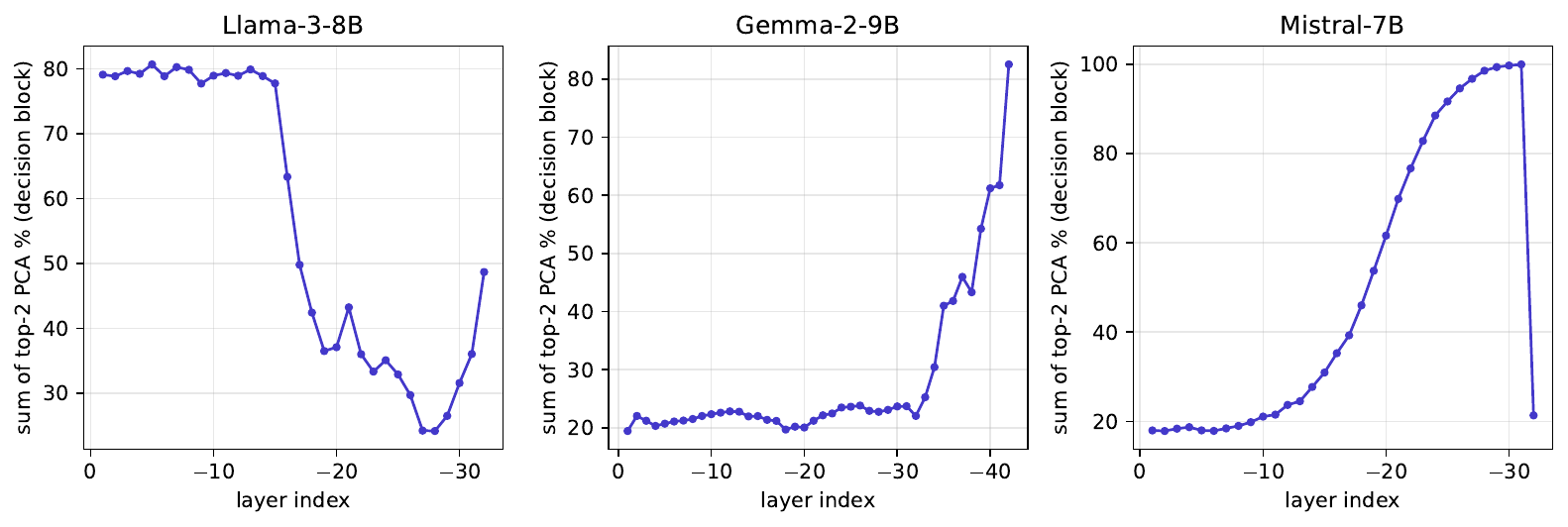}
    \caption{
    \textbf{Explained variance of decision subspace  on synthetic refusal dataset.}
    A small number of principal components explain most of the variance,
    suggesting that the decision mechanism is low-dimensional.
    }
    \label{fig:variance_appendix}
\end{figure}

Figure \ref{fig:variance_appendix} supports modeling the decision mechanism as a low-rank subspace.  Across layers, the top few components account for a large fraction of variance, with the top two components often explaining the majority in Llama and substantial mass in Gemma; Mistral exhibits a more diffuse spectrum in some depth ranges.
This supports treating the decision-aligned component as genuinely low-rank for the purposes of residualization, while also highlighting architecture-dependent spectral structure.

\subsection{Detailed Analysis of subcategories for synthetic refusal dataset}

\begin{figure*}[h]
    \centering
    \begin{subfigure}[b]{0.48\textwidth}
    \centering
    \includegraphics[width=0.7\textwidth]{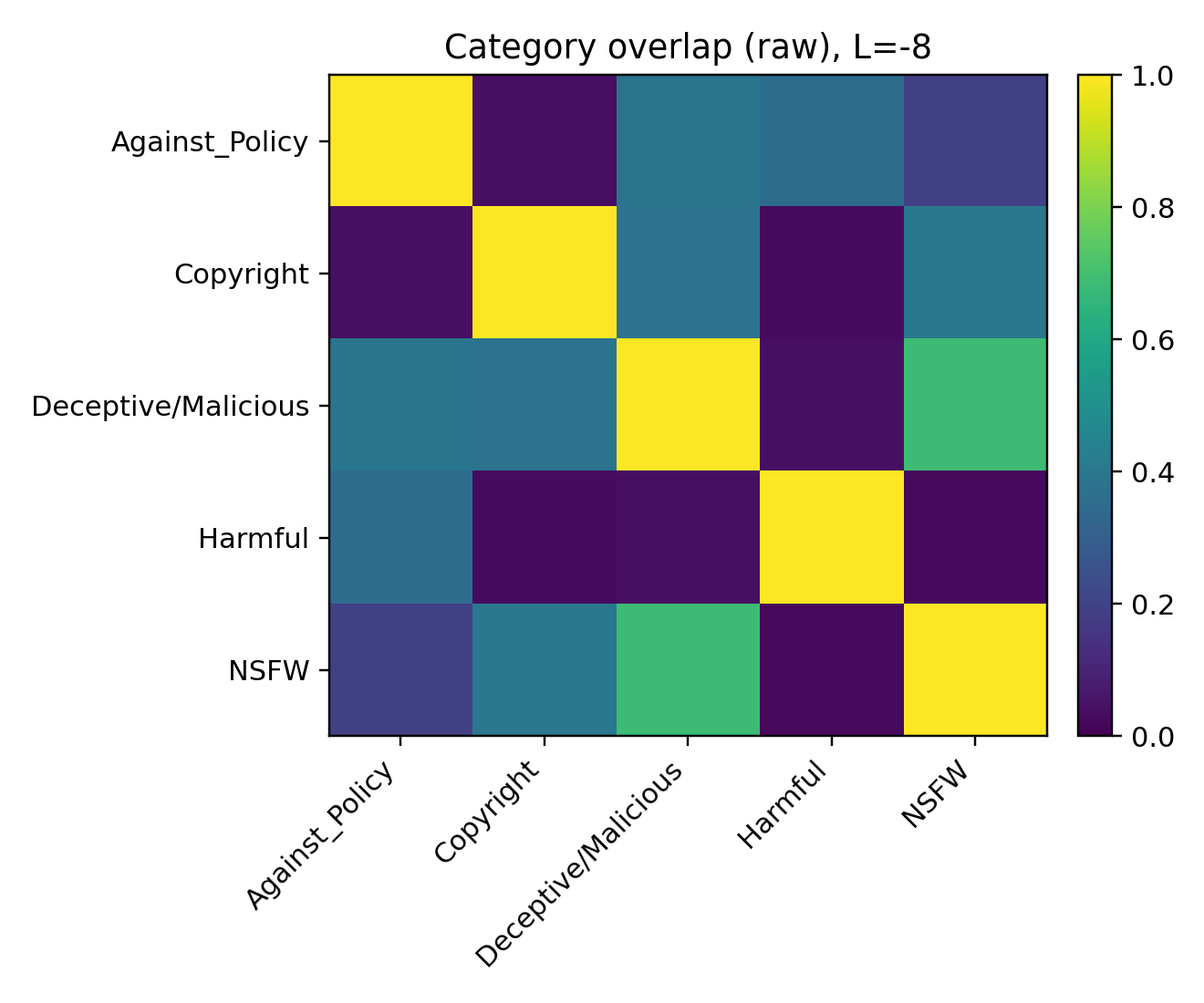}
    \caption{Raw overlap matrix }
    \label{fig:overlap_raw}
    \end{subfigure}
    \hfill
    \begin{subfigure}[b]{0.48\textwidth}
    \centering
    \includegraphics[width=0.7\textwidth]{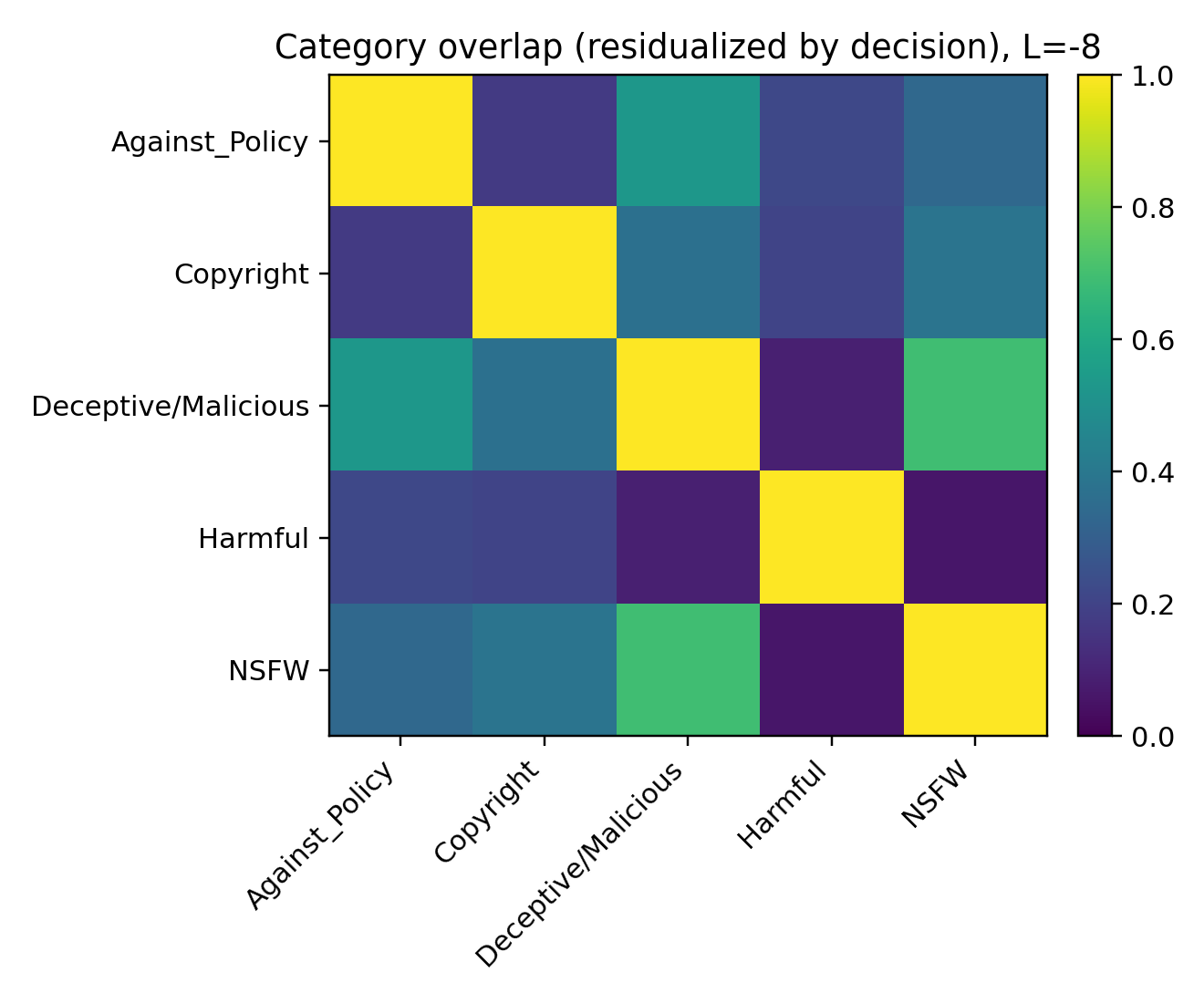}
    \caption{Residualized overlap matrix }
    \label{fig:overlap_resid}
    \end{subfigure}
    \caption{\textbf{Subcategory subspace overlap matrices for Gemma-2-9B-IT, layer $-8$ (8\textsuperscript{th} layer from the output)} (a) Raw overlap shows high similarity between many sub-categories, indicating shared decision-aligned subspace. (b) Residualized overlap (after removing decision component) reveals  subcategory-specific structure, with highest overlap between \textsc{Deceptive/Malicious} and \textsc{NSFW}, and lowest between \textsc{Harmful} and other categories. Raw overlap is  higher because it is dominated by the decision-aligned variance; after removing this component, informative subcategory-specific structure emerges.  }
    \label{fig:overlap_matrices}
    \end{figure*}

    Figure~\ref{fig:overlap_matrices} contrasts raw and residualized (after removing the decision-aligned variance) overlap.
    In the raw overlap matrix, most subcategories appear strongly aligned, reflecting
    dominance of a shared decision-aligned signal.
    After projecting out this signal, a more informative geometry emerges:
    some subcategory pairs remain tightly aligned while others become nearly orthogonal.
    Residualized overlap spans a wide range, from near-random values ($\approx 0.04$) to
    strong alignment ($\approx 0.7$--$0.8$), depending on the category pair and model.
    This demonstrates that subcategories share representational structure beyond the
    global decision-aligned subspace. 

      Table~\ref{tab:overlap_summary} reports residualized overlap at layer $-8$ across three
    model families.
    Despite architectural differences, all models exhibit nontrivial overlap with a
     qualitative pattern.
    In particular, \textsc{Deceptive/Malicious--NSFW} shows high residualized overlap in Gemma-2-9B and moderate overlap in Llama-3-8B, but low overlap in Mistral-7B (Table~\ref{tab:overlap_summary}), consistent with architecture-dependent category geometry,
    while \textsc{Harmful--Copyright} remains weakly overlapping.
    While the presence of residual overlap appears, its magnitude varies
    across models and layers, reflecting differences in learned representations which depend on training regimes and architectures
    (full layer-wise results in Appendix~\ref{app:layerwise_int}).

    \begin{table}[h]
    \centering
    \small
    \caption{\textbf{Principal angles between decision-aligned subspace and category subspaces (degrees, Gemma-2-9B-IT, layer -8 i.e., 8\textsuperscript{th} layer from output.)}Values shown are the two principal angles (first and second) for each category.  Smaller angles indicate greater alignment with the decision-aligned subspace.     \textsc{Copyright} and \textsc{Against Policy} show the strongest alignment along at least one     principal direction, while \textsc{Harmful} shows moderate alignment and \textsc{NSFW} and
    \textsc{Deceptive/Malicious} are more orthogonal.}
    \label{tab:angles}
    \begin{tabular}{lcc}
    \toprule
    \textbf{Category} & \textbf{Angle 1 (°)} & \textbf{Angle 2 (°)} \\
    \midrule
    \textsc{Harmful} & 57.2 & 72.0 \\
    \textsc{Copyright} & 37.2 & 89.3 \\
    \textsc{Deceptive/Malicious} & 75.2 & 87.7 \\
    \textsc{NSFW} & 79.7 & 89.9 \\
    \textsc{Against Policy} & 49.0 & 85.1 \\
    \bottomrule
    \end{tabular}
    \end{table}

        Table~\ref{tab:cross_effects} shows cross-effect matrices for Gemma-2-9B at layer~$-8$.  \textsc{Against Policy} acts as a  upstream mechanism, reducing refusal across all  categories by 6.7--8.7 log-probability units.
    \textsc{Deceptive/Malicious} also yields strong reductions (6.0--7.9 units). In contrast, projecting out \textsc{Copyright} or \textsc{NSFW} increases refusal (positive $\Delta M$),  suggesting that these subspaces support fine-grained discrimination between disallowed and borderline cases; removing them causes the model to default to a more conservative refusal behavior. \textsc{Harmful} shows comparatively weak cross-effects, indicating relative isolation.  

        Ablating \textsc{Against Policy} reduces \textsc{Harmful} refusal by
    6.71 units, whereas ablating \textsc{Harmful} reduces
    \textsc{Against Policy} refusal by only 0.21 units. Notably, ablating the \textsc{Copyright} or \textsc{NSFW} subspaces increases refusal rates.  Asymmetric effects arise because different behavior subspaces have different alignments with the decision-aligned subspace, so intervening on one behavior often perturbs the decision more strongly than intervening on another (despite the geometric overlap being symmetric).
    This apparent mismatch reflects the fact that overlap alone does not determine how interventions propagate through the network. 
    
    Table~\ref{tab:angles} analyzes how directly each subcategory subspace associated with the decision-aligned subspace by measuring their principal angles.
    Categories with smaller principal angles (e.g., \textsc{Against Policy}) are more
    tightly aligned with the decision-aligned subspace and produce large, global effects when
    intervened upon.
    However, categories that are nearly orthogonal to the decision subspace
    (e.g., \textsc{NSFW}, \textsc{Deceptive/Malicious}) exhibit 
    different effects despite substantial overlap with other categories.

       For example, \textsc{Deceptive/Malicious--NSFW} combines high overlap (0.69) with
    large cross-effects ($|\Delta M| \approx 7.55$), whereas
    \textsc{Harmful--Copyright} shows both low overlap (0.20) and weak effects
    ($|\Delta M| \approx 0.23$). 

    Figure \ref{fig:layer_appendix} shows that across the three model families, both, residualized overlap and cross-category effects,  exhibit layer-localized structure, with nontrivial geometric overlap and  transfer emerging in dominantly closer to the late layers, where behavior representations are composed prior to final decision consolidation. Layer $-8$ lies within this interaction regime for all models and is used as a common reference point for cross-model comparison. While the relationship is not strictly linear, regions of higher overlap tend to coincide with stronger cross-category transfer. These are again model-specific for their different representations.  Importantly, our conclusions do not depend on the exact layer choice: we observe qualitatively similar overlap–effect relationships across a contiguous window of layers, indicating that the reported geometry reflects stable representational structure rather than layer-specific artifacts.
    
    \begin{table*}[h]
    \centering
    \small
    \caption{\textbf{Residualized overlap between category pairs across models (layer -8 i.e., 8\textsuperscript{th} layer from output) (other layers in Appendix \ref{app:layerwise_int}.} Values are mean squared cosine between category subspaces after removing the decision component. Higher values (closer to 1.0) indicate more similar category-specific mechanisms. Models show varied but nontrivial overlap, with Llama-3 having the highest overlap.}
    \label{tab:overlap_summary}
    \begin{tabular}{lccc}
    \toprule
    \textbf{Category Pair} & \textbf{Gemma-2-9B-IT} & \textbf{Llama-3-8B-Instruct} & \textbf{Mistral-7B-Instruct}  \\
    \midrule
    Harmful $\leftrightarrow$ Copyright & 0.20 & 0.48 & 0.25   \\
    Harmful $\leftrightarrow$ Deceptive/Malicious & 0.09 & 0.68 & 0.17  \\
    Harmful $\leftrightarrow$ NSFW & 0.06 & 0.53 & 0.41   \\
    Harmful $\leftrightarrow$ Against Policy & 0.22 & 0.71 & 0.30  \\
    \midrule
    Copyright $\leftrightarrow$ Deceptive/Malicious & 0.37 & 0.46 & 0.28  \\
    Copyright $\leftrightarrow$ NSFW & 0.39 & 0.42 & 0.32 \\
    Copyright $\leftrightarrow$ Against Policy & 0.17 & 0.51 & 0.48 \\
    \midrule
    Deceptive/Malicious $\leftrightarrow$ NSFW & 0.69 & 0.41 & 0.12 \\
    Deceptive/Malicious $\leftrightarrow$ Against Policy & 0.53 & 0.64 & 0.33  \\
    \midrule
    NSFW $\leftrightarrow$ Against Policy & 0.34 & 0.54 & 0.38  \\
    \bottomrule
    \end{tabular}
    \end{table*}
  
    \begin{table*}[h]
    \centering
    \small
    \caption{\textbf{Cross-effect matrix (Gemma-2-9B-IT, layer -8, $\alpha=1.0$).} Values are $\Delta$ refusal margin (log-probability units) when projecting out each source category's subspace. Positive values indicate increased refusal, negative values indicate decreased refusal. Diagonal entries show self-effects.  We use projection-based
    interventions to test whether behavior-specific representations can be selectively modified without inducing cross-behavior effects. See Section \ref{results:h2} for analysis of asymmetry and cross-category effects. }
    \label{tab:cross_effects}
    \begin{tabular}{l|ccccc}
    \toprule
    \textbf{Source $\rightarrow$ Target} & \textbf{Harmful} & \textbf{Copyright} & \textbf{Deceptive} & \textbf{NSFW} & \textbf{Against Policy} \\
    \midrule
    Against Policy & -6.71 & -8.73 & -7.82 & -7.78 & -8.17 \\
    Copyright & +2.85 & +3.84 & +3.12 & +3.19 & +3.44 \\
    Deceptive/Malicious & -6.99 & -7.25 & -7.86 & -7.55 & -7.86 \\
    Harmful & -0.47 & -0.23 & -0.31 & -0.34 & -0.21 \\
    NSFW & +1.31 & +0.31 & +1.03 & +0.98 & +0.70 \\
    \bottomrule
    \end{tabular}
    \end{table*}

    \begin{figure*}[h]
    \centering
    \begin{subfigure}[b]{0.49\textwidth}
    \centering
    \includegraphics[width=0.8\textwidth]{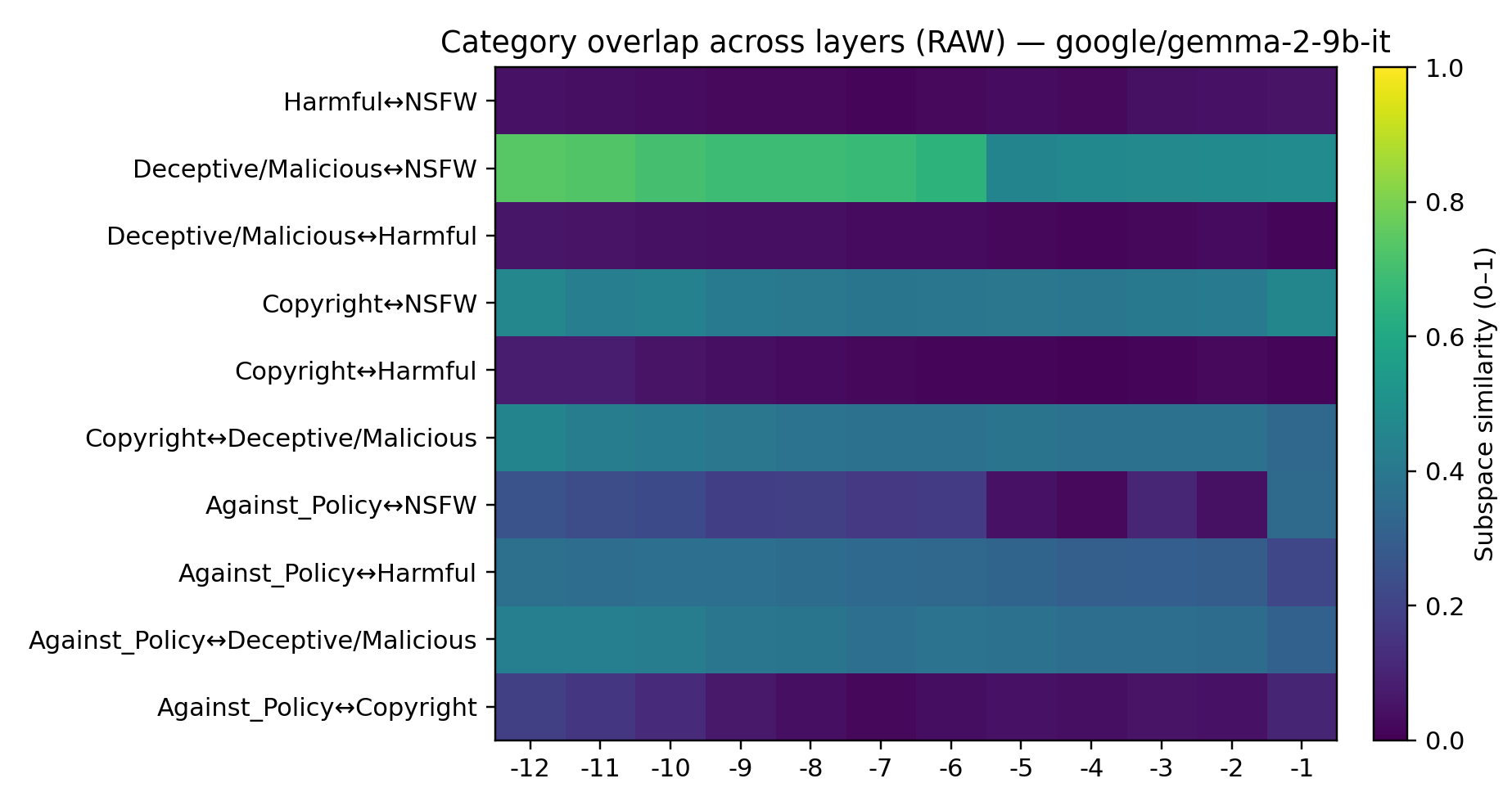}
    \caption{Raw overlap across layers}
    \label{fig:layer_map_raw}
    \end{subfigure}
    \hfill
    \begin{subfigure}[b]{0.49\textwidth}
    \centering
    \includegraphics[width=0.8\textwidth]{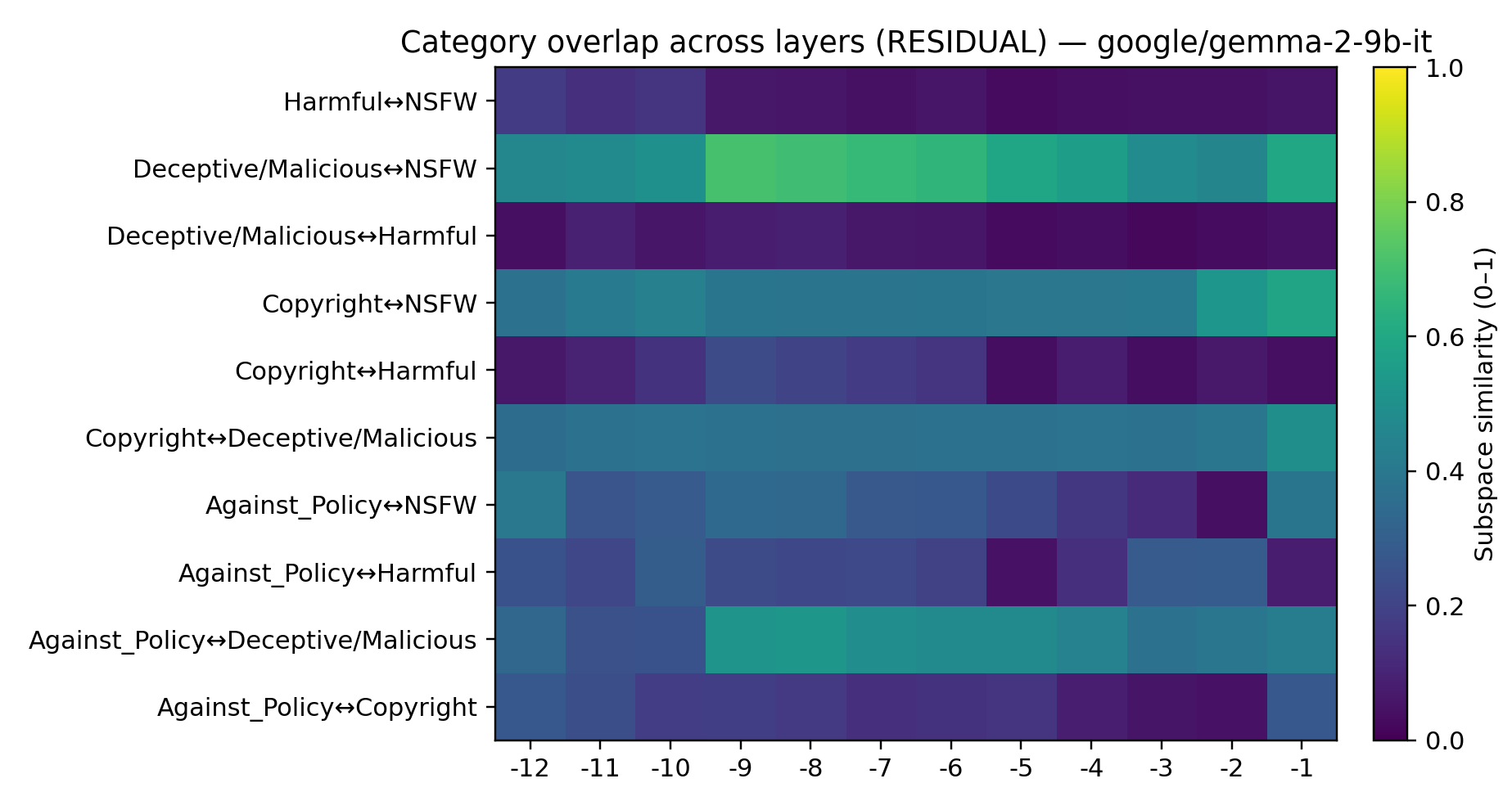}
    \caption{Residualized overlap across layers}
    \label{fig:layer_map_resid}
    \end{subfigure}
    \caption{\textbf{Layer-wise category overlap for Gemma-2-9B-IT.} Each row shows overlap between a category pair across the final 12 layers,  indexed relative to the output (layer $-1$ denotes the final layer). (a) Raw overlap is  high across  layers. (b) Residualized overlap varies across depth, with strong entanglement in mid-layers (approximately $-12$ to $-8$) and more heterogeneous in later layers. This suggests category-specific structure emerges in mid-layers ($\approx$ $-12$ to $-8$), while decision consolidation happens later.}
    \label{fig:layer_maps}
    \end{figure*}

\end{document}